\documentclass[11pt]{style/prism-princeton}

\makeatletter
\newcommand{\longdash}[1][2em]{%
  \makebox[#1]{$\m@th\smash-\mkern-7mu\cleaders\hbox{$\mkern-2mu\smash-\mkern-2mu$}\hfill\mkern-7mu\smash-$}}
\makeatother
\newcommand{\omitskip}{\kern-\arraycolsep}

\author[1,2]{Tanush Yadav}
\author[1,2]{Mohammadreza Salehi}
\author[1,2]{Jae Sung Park}
\author[1]{Vivek Ramanujan}
\author[1,2]{Hannaneh Hajishirzi}
\author[3]{Yejin Choi}
\author[1,2]{Ali Farhadi}
\author[2 \dagger]{Rohun Tripathi}
\author[1,2 \dagger]{Ranjay Krishna}

\affiliation[1]{University of Washington}
\affiliation[2]{Allen Institute for AI}
\affiliation[3]{Stanford University}

\contribution[\dagger]{Equal advising}

\usepackage{wrapfig}
\usepackage[most]{tcolorbox}
\usepackage{xcolor}
\usepackage{arydshln, subcaption}
\usepackage{epigraph}

\definecolor{polaris-bg-elevated}{HTML}{f5f5f5}
\definecolor{polaris-border-subtle}{HTML}{e5e5e5}

\definecolor{filter-bg}{HTML}{ebebeb}
\definecolor{filter-border}{HTML}{c8c8c8}

\newcommand{\model}{Molmo2-4B}

\newcommand{\filter}[1]{%
  \tcbox[
    on line,
    boxsep=0pt,
    left=3pt, right=3pt, top=1pt, bottom=1pt,
    boxrule=0.4pt,
    arc=2pt,
    colback=filter-bg,
    colframe=filter-border,
  ]{\textsc{#1}}%
}

\newcommand{\titleoneclip}{%
  \filter{SingleAction}%
}

\newcommand{\transcript}{%
  \filter{TranscriptLocalized}%
}

\newcommand{\strict}{%
  \filter{TranscriptLocalizedTitleMatch}%
}

\begin{document}
\newcommand{\dataset}{VideoNet}

\abstract{
Videos are unique in their ability to capture \textbf{actions} which transcend multiple frames. Accordingly, for many years action recognition was \textit{the} quintessential task for video understanding. Unfortunately, due to a lack of sufficiently diverse and challenging data, modern vision-language models (VLMs) are no longer evaluated on their action recognition capabilities. To revitalize action recognition in the era of VLMs, we advocate for a returned focus on \textbf{domain-specific} actions. To this end, we introduce \dataset, a domain-specific action recognition benchmark covering 1,000 distinct actions from 37 domains. We begin with a multiple-choice evaluation setting, where the difference between closed and open models is stark: Gemini 3.1 Pro attains 69.9\% accuracy while Qwen3-VL-8B gets a mere 45.0\%. To understand why VLMs struggle on \dataset, we relax the questions into a binary setting, where random chance is 50\%. Still, Qwen achieves only 59.2\% accuracy. Further relaxing the evaluation setup, we provide $k\in\{1,2,3\}$ in-context examples of the action. Some models excel in the few-shot setting, while others falter; Qwen improves $+7.0\%$, while Gemini declines $-4.8\%$. Notably, these gains fall short of the $+13.6\%$ improvement in non-expert humans when given few-shot examples. Finding that VLMs struggle to fully exploit in-context examples, we shift from test-time improvements to the training side. We collect the first large-scale training dataset for domain-specific actions, totaling nearly 500k video question-answer pairs. Fine-tuning a Molmo2-4B model on our data, we surpass all open-weight 8B models on the \dataset~benchmark.
}


\title{\dataset: A Large-Scale Dataset for Domain-Specific Action Recognition}

\website{
https://tanu.sh/research/videonet  
}
{
tanu.sh/videonet   
}



\maketitle

\section{Introduction}
\label{sec:intro}
\vspace{-5pt}
\renewcommand{\epigraphflush}{flushleft}
\setlength{\epigraphrule}{0pt}
\epigraph{``\textit{Ignorato motu, ignoratur natura.}\\Who knows not motion, knows not nature.''}
{Aquinas}


\begin{figure*}[t!]
    \centering
    \includegraphics[width=\linewidth]{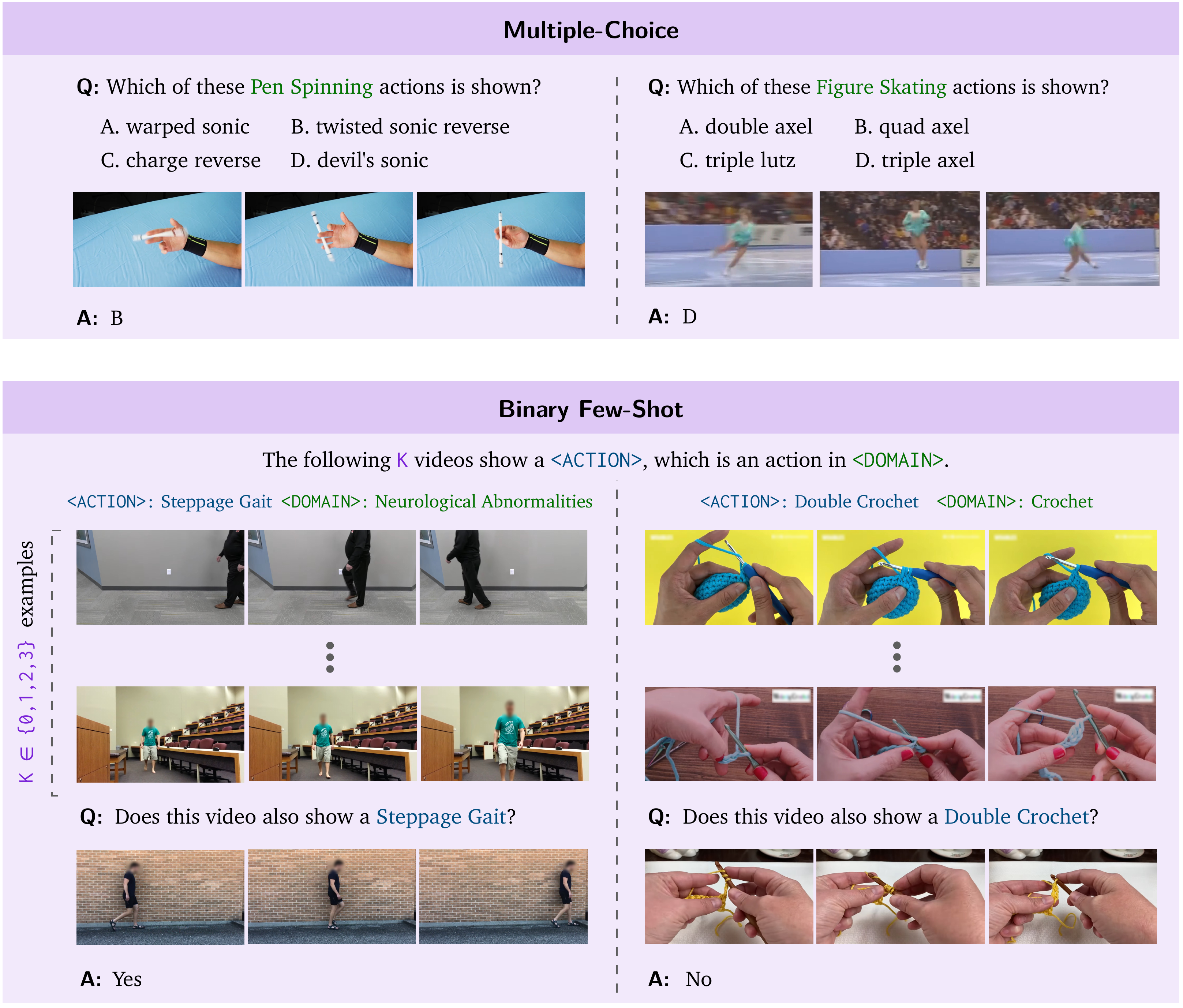}
    \caption{\textbf{Q\&A examples from \dataset.} We provide two evaluation settings: multiple-choice and few-shot binary. The former focuses on the core task of \textit{domain-specific action recognition}; the latter focuses on a model's ability to learn from \textit{in-context videos}. (The prompts above have been simplified for succinctness.)
    }
    \label{fig:teaser}
\end{figure*}

Action recognition has proven to be an evergreen goal of the computer vision community. Since as early as 1992, highly-influential works have highlighted the difficulty of recognizing \textit{domain-specific} actions in particular (e.g., \cite{cvpr1992_tennis_actions} focused on categorizing six distinct tennis strokes). Yet domain-specific data is \textit{notoriously difficult} to collect, so little work has been done on gathering domain-specific data across a wide variety of domains. In the era of large vision-language models (VLMs), where testing generalizability is a key concern of many researchers, this lack of diverse domain-specific data has prevented VLMs from being evaluated on this ``forgotten'' task. Instead, the VLM community has focused on fine-grained actions that are \textit{not} domain-specific, such as whether a ball rotates clockwise or counter-clockwise \cite{tomato}. While such benchmarks are valuable, they fail to capture the real-world applicability of inquiring about domain-specific actions. Furthermore, they only test perception skills, whereas recognizing actions like a ``triple flip jump'' in figure skating requires models to excel not only at  perception but also at compositional reasoning (i.e., are all elements of the action present and in the correct order?). In fact, fine-grained movements often underlie domain-specific actions (e.g., the use of a toepick differentiates a ``flip jump'' from a ``Salchow jump''), so testing domain-specific action understanding inevitably tests fine-grained action understanding.\footnote{As another example, consider the ``thumbaround'' and ``thumbaround reverse'' in pen spinning, which differ only in the direction of rotation. They both differ from a ``fingerless thumbaround'' and ``fingerless thumbaround reverse'' only on the basis of whether the middle finger remains stationary.} 

In this paper, we introduce the data necessary to make domain-specific action recognition relevant in the VLM era. To this end, we present a benchmark covering 1,000 actions across 37 domains. We confirm the validity of our test set labels with expert verification, signaling a near 97\% accuracy rate.

VLMs struggle on our benchmark. In the multiple-choice setting, the best open-weight 8B VLM attains 45.0\% accuracy, while the best proprietary VLM achieves 69.9\%. In the relaxed binary setting, where random chance is 50\%, the best open-weight 8B VLM reaches a mere 59.2\% accuracy, while non-expert humans achieve 69.1\%. We ablate our visual and textual inputs to the VLMs to understand why models perform poorly on this task. We hypothesize that a lack of domain-specific action data in these models' training mixtures is partially responsible for poor performance.

\begin{figure*}
    \centering
    \includegraphics[width=\linewidth]{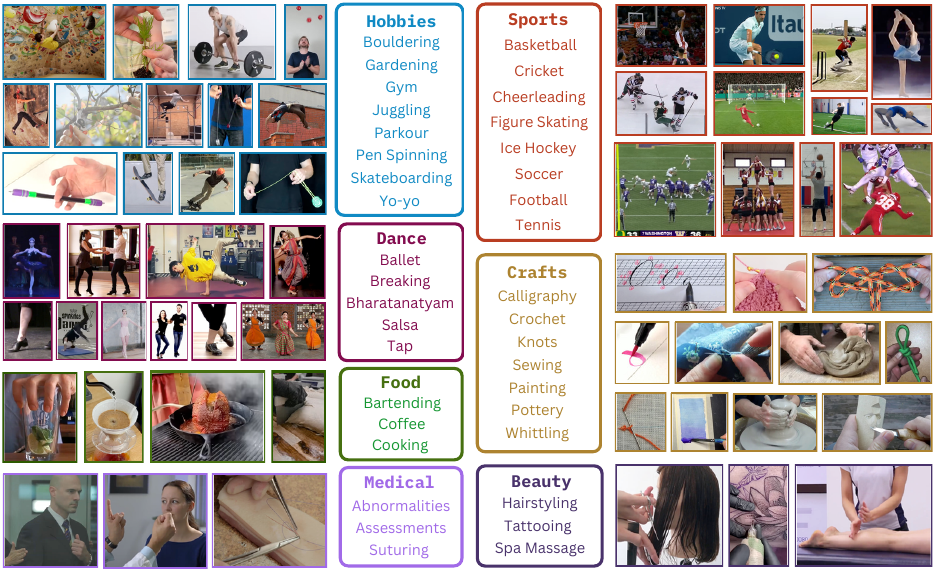}
    \caption{\textbf{Video samples} from all 7 categories and 37 domains in \dataset. An \href{https://tanu.sh/videonet/data}{interactive demo} of the benchmark's videos is available on the project website.}
    \label{fig:domains}
\end{figure*}

Inspired by few-shot learning \cite{brown2020languagemodelsfewshotlearners, min-etal-2022-metaicl}, we investigate whether this lack of domain-specific training data can be overcome with few-shot examples of actions at test time. Indeed, non-expert human performance improves by 13.6 percentage points when given three few-shot examples. Yet VLMs improve, on average, by $2.9$ percentage points, suggesting that they are poor few-shot learners and implying that domain-specific action understanding deficiencies cannot currently be fixed at test time.

Finally, we explore post-training on domain-specific action data. We collect a training set containing 160,000 clips. Fine-tuning a 4B VLM on our data yields an 11.5 percentage point improvement on \dataset. Notably, this nears the performance improvement observed in humans when given three in-context examples. Our 4B model surpasses the current generation of open-weight 8B models.

\noindent Our contributions include:

\begin{itemize}
    \item A \textbf{domain-specific action recognition benchmark} covering 1,000 actions across 37 domains. 
    \item A \textbf{domain-specific action training dataset} with 160,000 clips that \textit{enables 4B models to surpass Qwen3-VL-8B}. 
    \item Two innovative \textbf{data pipelines}, for human annotation and synthetic  labeling, that break from traditional literature by \textit{circumventing the need for domain experts}.
    \item Few-shot evaluation of VLMs, highlighting their \textbf{deficiencies with in-context learning}. 
\end{itemize}

We are particularly excited about how our data unlocks future research into modeling decisions for perception, visual reasoning, and real-world action understanding.\footnote{Action understanding is a prerequisite to action quality analysis. Imagine if a VLM could help a new gym goer learn proper squat technique or critique a novice figure skater's lutz jumps.}

\section{Related Work}
\label{sec:related_work}

Action recognition has been extensively explored. Existing efforts fall broadly into three categories. The first~\cite{KTH,ucf101,kuehne2011hmdb,activitynet,ava,moments,kinetics700} predominantly contain \textit{coarse-grained} labels (e.g., \cite{activitynet} has a single class for "rock climbing", whereas \dataset~contains 22 distinct bouldering actions). Unsurprisingly, foundation models excel at recognizing such coarse-grained labels, with InternVideo2 \cite{internvideo2} attaining 92.1\% on Kinetics-400 and 95.9\% on ActivityNet. The second set~\cite{breakfast,mpi-cooking,finediving,diving48,actionatlas,multisports,finesports,finegrained_novel_basketball} focus on a limited set of sports, rendering them unable to test the generalization promise of foundation models. The third set~\cite{temporalbench,tomato,motionbench,burgess2025videoactiondifferencing} fixate on fine-grained temporal attributes, such as the direction and trajectory of moving objects. While these works pose interesting perception questions, they focus on details (e.g., does an object move from left to right) that an end user is unlikely to consult a large model for, raising concerns about their real-world utility. \dataset, on the other hand, incorporates these fine-grained movements--which are innate to domain-specific actions--into a more realistic setting. Thus, unlike these three groups, \dataset~contains fine-grained labels with real-world applicability across a sufficiently large set of domains.

There are three notable works that collect domain-specific action data across a variety of domains. The first, Ego-Exo4D~\cite{egoexo4d}, covers only 8 domains, compared to \dataset's 37. Our benchmark rivals the size of Ego-Exo4D's entire dataset, while our training data contains 30x more videos. Perhaps most critically, Ego-Exo4D lacks visual diversity; its 728 bouldering videos, for instance, were filmed at 2 climbing gyms. In contrast, \dataset~sources videos from the web, enabling a great range of visual composition. The second, Ego4D~\cite{ego4d}, collects fine-grained actions in videos. However, it is restricted to egocentric videos. The third, ActionAtlas \cite{actionatlas}, collects 934 videos across 56 sports. It is similar to \dataset~in style, but less generalizable due to its exclusive focus on sports. ActionAtlas notably forgoes the question of training data, and even its benchmark is 5 times smaller than \dataset's.

\section{Benchmark Construction}
\label{sec:benchmark}

\subsection{Preparing actions}
\label{subsec:benchmark_actions}

We employ a top-down approach to generate our taxonomy of actions. First, we formulate a list of categories designed to cover actions that are applicable to daily life (e.g., food), require expert-level knowledge (e.g., medical), or demand a high frame sampling rate for recognizing rapid motions (e.g., sports). Within each category, we find domains that have sufficient videos and trusted expert content online. We then compile actions for each domain from expert-written sources (e.g., skateboarding actions from a respected skateboarding blog) and augment these lists using LLMs (following \cite{actionatlas}, see Appendix \ref{appendix:action_list_llm} for details). Finally, we remove actions with an insufficient amount of videos online.

\label{subsubsec:definitions}
Action definitions are used throughout our project to help humans and models classify videos without specialized domain expertise. To maximize their usefulness, the definitions are written to focus on visual cues and defining characteristics of an action, as well as key differentiators from similar actions. We initially used LLMs to generate definitions, following~\cite{actionatlas}. However, specialized domains pose challenges, as LLMs occasionally encode incorrect or outdated domain knowledge~\cite{Tonmoy2024ACS}. To mitigate this issue, we enable LLMs to perform targeted web searches \cite{claude_web_search}, retrieving expert-curated information from reputable online knowledge bases and domain-specific communities. The LLMs use this information to cross-check definitions and correct inaccuracies, providing a final set of definitions aligned with established domain expertise.

\subsection{Collecting well-trimmed clips}
\label{subsec:benchmark_collection}

\begin{figure*}
    \centering
    \includegraphics[width=0.98\linewidth]{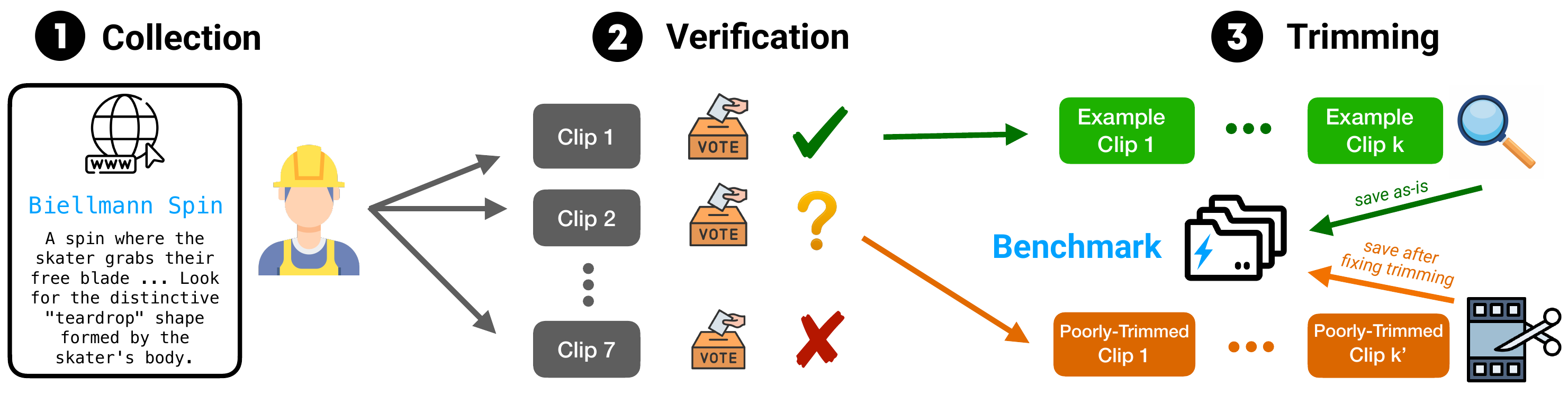}
    \caption{\textbf{Benchmark data collection pipeline}, as described in Section \ref{subsec:benchmark_collection}. Given an action name and definition, humans (1) find clips on the web, (2) remove outliers among these clips, and (3) fix the clip trimmings. This pipeline yields five well-trimmed clips per action.}
    \label{fig:collection}
\end{figure*}

After preparing our action lists, we launch our three-stage human-annotation pipeline, as visualized in \Cref{fig:collection}. Our pipeline design is guided by rigorously validated HCI practices \cite{crowdsourcing_error_bounds, crowdsourcing_quality_control_mechanisms, crowdsourcing_quality_control_survey}. Across the entire pipeline, five distinct annotators review each clip before it is finalized.

\textbf{Video collection.} We provide a human annotator, sourced from Prolific, with an action's name, domain, and definition (\S~\ref{subsubsec:definitions}). They are told to search for the action online and find seven clips where the action occurs. We require the clips to be sourced from distinct videos to increase generalizability. 

\textbf{Clip verification.} We provide an annotator an action's name, definition, and its seven candidate clips from the previous stage. We ask them to rate each clip as (1) containing the action and being well-trimmed, (2) containing the action but being poorly-trimmed, or (3) not containing the action. Determining if a clip is well-trimmed or not is trivial for humans; however, determining the presence of an action can be tricky, especially since these annotators are \textit{not} domain experts. We solve this dilemma by reducing the problem from $k$-way classification to 2-way classification. Where \cite{finesports} and \cite{multisports} showed a domain expert a random clip and asked them to classify it as one of $k$ actions, we ask non-expert annotators to classify each clip as containing or not containing the desired action. Empirically, five to six of the seven clips typically contain the desired action, further simplifying this task to an outlier detection problem. For increased confidence, we take the majority vote from three annotators on this stage \cite{crowdsourcing_error_bounds}.

\textbf{Clip trimming.} We reach this stage with nearly all actions having five or more clips that were deemed to contain the desired action. At least one of these clips was always well-trimmed; in four-fifths of cases, there were at least three well-trimmed clips. 
To preserve clips that contain the desired action but are poorly trimmed, we have an additional stage of trimming to refine their temporal boundaries. Here, we show a Prolific annotator an action's name, definition, and these well-trimmed examples, thereby training them to be an ``expert" on the action. We then ask them to fix the trimmings on the poorly-trimmed clips. This leaves us with at least five accurately trimmed clips for the desired action.

This process yields 5,000 clips, with average and typical durations of 12.2 and 5.0 seconds respectively. The clips are \textit{well-trimmed} in that they contain the entirety of an action and minimal fluff around that action. Certain domains, like suturing and crochet, contain actions that take longer to demonstrate, causing a noticeable tail in the distribution of video lengths (see Appendix \ref{appendix:benchmark_statistics}). To alleviate context length issues, especially in the 3-shot setting, we impose a maximum duration of 5 minutes.

For good measure, one of the authors manually inspected and adjusted the labels and trimmings of all 5,000 clips produced by the aforementioned pipeline.

\subsection{Verifying clip labels}
\label{subsec:expert_verification}

\begin{wraptable}{r}{0.55\linewidth}
    \centering
    \vspace{-1.5cm}
    \begin{tcolorbox}[
    enhanced,
    width=\linewidth,
    colback=polaris-bg-elevated,
    colframe=polaris-border-subtle,
    arc=3mm,
    boxrule=0.4pt,
    left=4pt,
    right=4pt,
    top=6pt,
    bottom=6pt,
]
    \caption{\textbf{Expert Verification.} These results confirm that our benchmark data pipeline is robust to annotator error.}
    \label{tab:expert_verification}
    \centering
    \resizebox{\linewidth}{!}{
    \begin{tabular}{@{}ccccc@{}}
        \toprule[1.5pt] 
         \textbf{Category} & \textbf{Domain} & \textbf{Correct Clips} & \textbf{Total Clips} & \textbf{Percent Correct} \\ \midrule
         Sports & Tennis & 92 & 95 & 96.8\% \\
         Food & Coffee & 75 & 80 & 93.8\% \\
         Crafts & Painting & 39 & 40 & 97.5\% \\
         Medical & Neuro. Exams & 75 & 75 & 100.0\% \\
         Dance & Break Dance & 162 & 165 & 98.2\% \\
         Hobbies & Gym & 107 & 110 & 97.3\% \\
         Beauty & Spa Massage & 55 & 55 & 100\% \\
         [0.8ex] \hdashline \\[-1.8ex]
         All & All & 605 & 620 & 97.6\% \\
        \bottomrule[1.1pt]
    \end{tabular}
    }
\end{tcolorbox}
    \vspace{-0.5cm}
\end{wraptable}

To measure the correctness of our benchmark, we conduct expert verification. We choose one domain from each of our 7 categories for verification, hypothesizing that accuracies for domains within each category should be similar. In total, experts verify 620 clips; generalizing human performance from this scale is in line with prior works \cite{mmmu, ilsvrc}. When possible, we find experts in our local communities and ask them to verify the data labels, akin to \cite{ego4d,mmlu_pro,tomato}. For domains where we are unable to locate experts, we train someone on a large sample of the domain's data, before asking them to verify labels, following \cite{ilsvrc}. As shown in \Cref{tab:expert_verification}, we see 97\% accuracy in our data, exceeding MMLU-Pro's \cite{mmlu_pro} expert accuracy of 85.4\% and ImageNet's \cite{ilsvrc} estimates of top-5 error at 5.1\% and 12.0\%. This confirms the validity of our pipeline as a replacement for hiring domain experts during the domain-specific data collection process. It also enables researchers developing future models to confidently use \dataset~as a test bed for domain-specific capabilities.

\subsection{Generating (hard) negatives}
\label{subsec:hard_negatives}
With the verified \textit{positive} clips in-hand, we gather suitable \textit{negative} examples to be used in our benchmark.\footnote{Section \ref{subsec:hard_negatives} describes how we generate, for a given action, its negative \textit{actions}. Since these negative actions exist in \dataset, it is easy to find negative \textit{clips} once we have the negative actions; that process is described in Section \ref{subsec:forming_qa_sets}.}
One approach is to gather ``random negatives'' by randomly sampling different actions within the same domain. This approach has a fatal flaw: different actions often have distinct contexts, backgrounds, or static visual cues. Without careful control, models may achieve high performance by exploiting scene-level details alone (e.g. \textit{alley-oop dunk} vs. \textit{free throw} in basketball), rather than closely watching the entire clip. Instead, we create challenging ``hard negatives'' by selecting actions that closely resemble the positive clip, only differing in subtle visual or motion-related aspects. We first generate these hard negatives with an LLM, akin to \cite{actionatlas, bansal2023videoconrobustvideolanguagealignment}. Unlike prior methods, we then refine this candidate set using a reasoning model to filter out candidates that could realistically co-occur with the positive action or are otherwise ambiguous. This ensures that our hard negatives are valid and challenging (e.g. \textit{alley-oop dunk} vs. \textit{put-back dunk}). The prompts used to generate hard negatives are provided in Appendix \ref{appendix:action_hard_negatives_llm}, along with additional refinement details.

\subsection{Forming the Q\&A sets}
\label{subsec:forming_qa_sets}

Once we have 5 video clips and 3 hard-negative text labels for each of our 1,000 actions, we form the multiple-choice and binary versions of our carefully-curated evaluation set. The differences between these two evaluation settings are summarized in \Cref{tab:mcq_vs_binary_summary}.

In the multiple choice setting, for each clip we use its 1 positive (ground-truth) label and 3 hard negative labels to form a question with 4 text options. This yields 5,000 questions, 1,000 of which are set aside as a validation set.

In the binary setting, we use an action's first 3 clips as its in-context examples. The remaining 2 clips become positive test clips. We then select 2 of the hard negative text labels for that action, and from those we source 2 negative test clips. This yields 3 in-context examples, 2 verified positive test clips, and 2 hard negative test clips for each action. This forms a 4,000 question test set. We do not provide a validation set for the binary setting.

\begin{tcolorbox}[
    enhanced,
    width=\linewidth,
    colback=polaris-bg-elevated,
    colframe=polaris-border-subtle,
    arc=3mm,
    boxrule=0.4pt,
    left=8pt,
    right=8pt,
    top=8pt,
    bottom=8pt,
]
\vspace{-0.5cm}
\begin{table}[H]
    \centering
    \caption{\textbf{Multiple-Choice vs Binary.} We anticipate that future models benchmarking their action recognition capabilities via \dataset~will primarily report results on the multiple-choice evaluation setting. We include the few-shot setting as a resource for researchers who wish to explore video in-context learning.}
    \resizebox{0.9\linewidth}{!}{
    \begin{tabular}{cccc}
        \toprule
        \textbf{Evaluation Setting}  & \textbf{Test Set Size} & \textbf{Val Set Size} & \textbf{Targeted Model Capability} \\
        \midrule
        multiple-choice & 4,000 & 1,000 & domain-specific action recognition \\
        binary few-shot & 4,000 & - & video in-context learning \\
    \bottomrule
    \end{tabular}
    \label{tab:mcq_vs_binary_summary}
    }
\end{table}
\end{tcolorbox}
\section{Model Training}
\label{sec:model_training}

We create a large-scale training dataset of domain-specific actions using a fully automated pipeline. Fine-tuning an open VLM on this dataset, we demonstrate a significant improvement in the base model's performance on both the binary and multiple-choice settings of \dataset.

\subsection{Training Data}
\label{subsec:training_data}

While the data collection pipeline described in Section \ref{subsec:benchmark_collection} leads to high-quality clips, its reliance on human annotators renders it \textit{prohibitively expensive} for collecting training-scale data. A common solution in such scenarios is to rely on synthetic labels generated by foundation models \cite{gpt3_data_annotator, openthoughts}. As shown in Section \ref{subsec:zero_shot}, VLMs struggle to recognize domain-specific actions, so distilling directly from even the best-performing VLM is unideal. Instead, we choose to rely on signals surrounding the video, specifically the video's title and transcript.

We build up our training data one domain at a time for each of the 37 domains. For a given domain, we begin by crawling relevant videos from the web. To do so, we construct queries from our action list, e.g. from ``laser flip'' we construct queries like ``skateboarding laser flip'' and ``how to laser flip''. Once we have a pool of relevant videos for a domain, we extract clips of that domain's actions using Gemini 2.5 Flash as a localizer. For instance, we ask Gemini to provide start and end timestamps for each clip in a video where a skateboarding action occurs. Critically, even though Gemini struggles to \textit{label} the actions in these clips, it excels at \textit{localizing} them. Once we have a set of domain-specific action clips extracted from our pool of domain-specific videos, we must filter and label these clips. A video's audio can be helpful for labeling clips, so we extract word-level timestamps using WhisperX~\cite{bain2022whisperx}. With a video's title and transcript in-hand, we experiment with three strategies to filter and label the Gemini-localized clips:

\begin{enumerate}
    \item \transcript. If an action name appears in the video's transcript within $T=1$ seconds of a localized clip, the clip is labeled with that action.
    \item \strict. Refining on top of \transcript, we further require that the action also appear in the video's title. 
    \item \titleoneclip. If an action appears in the video's title, and the localizer identifies \textit{only one clip} in the entire video, that clip is labeled with the action from the title.
\end{enumerate}

In total, we crawl 8 million videos before localizing 1.5 million videos. This yields 6 million clips, which we filter into training sets ranging in size from roughly 160,000 clips to 500,000 clips. We generate 3 video question-answer (VQA) pairs from each clip, as described in Appendix \ref{appendix:dataset_construction}. Training results for the different data filtering strategies are provided in Section \ref{subsec:training_results}. 

\subsection{Training Details}
\label{subsec:post_training}
We fine-tune \model ~\cite{clark2025molmo2}, an instruction-tuned VLM, on our filtered training datasets. The model's architecture consists of a vision transformer (ViT)~\cite{dosovitskiy2021vit} connected to a LLM via a MLP connector module. 
During training, the frames are sampled at $S=4$ frames per second, up to a max frames of $F=64$. If a video's duration is greater than $F / S$ seconds, $F$ frames are uniformly sampled from the video instead. To preserve temporal information, before each sampled frame we encode the frame's timestamp in seconds as text input to the LLM. In all of our experiments, we train the model for 8,000 steps with a batch size of $128$. Additional training details are reported in Appendix \ref{appendix:training_setup}.
\section{Experiments}
\label{sec:experiments}

We evaluate state-of-the-art VLMs on \dataset, beginning with a brief coverage of the multiple-choice setting followed by in-depth analysis of the binary few-shot setting. We also discuss the results of fine-tuning Molmo2 on our dataset.

For open models, we use Qwen3-VL-8B-Instruct~\cite{Qwen3-VL}, InternVL3.5-8B~\cite{internvl3_5}, and Molmo2-8B~\cite{clark2025molmo2}. For proprietary models, we use Gemini 3.1 Pro, Gemini 3 Flash, GPT-5.4, and GPT-5.\footnote{Snapshots/versions of proprietary models listed in Appendix \ref{appendix:eval}.} When passing videos as inputs, we use a model's recommended sampling strategy: uniform sampling for InternVL3.5-8B (max 48 frames); fps sampling for Qwen3-VL (2fps), GPT (1fps), and Gemini (1 fps). For our model, we use 4 fps sampling up to a max of 64 frames. A discussion of these models' context lengths--which may impact few-shot performance--is available in Appendix \ref{appendix:eval}. Results for CLIP models \cite{XCLIP,internvid,longclip,videoclipxl} and optical flow models \cite{quovadis_kinetics} can be found in Appendix \ref{appendix:traditional_models}. 

\begin{table}[h!]
    \begin{tcolorbox}[
    enhanced,
    width=1.0\linewidth,
    colback=polaris-bg-elevated,
    colframe=polaris-border-subtle,
    arc=3mm,
    boxrule=0.4pt,
    left=8pt,
    right=8pt,
    top=8pt,
    bottom=8pt,
    ]
    \caption{\textbf{Multiple-choice evaluation results.} Open models hover at 45\%, while closed models fall just short of 70\%. This sizeable gap suggests that \dataset~is challenging for existing VLMs, with high performance requiring larger models. A category-wise random baseline is the best accuracy attainable by guessing 1 letter for all questions in that category; the overall random baseline is the best accuracy attainable by guessing 1 letter for the entire benchmark. Highest accuracy for each column is in \textbf{bold}; highest accuracy by a non-proprietary model is \underline{underlined}. Our fine-tuned model is in \textcolor{princeton_orange}{purple}.}
    \vspace{5pt}
    \centering
    \resizebox{\linewidth}{!}{
    \begin{tabular}{@{}llcccccccc@{}}
        \toprule[1.5pt] 
         & \textbf{Model Name} & \textbf{Beauty} & \textbf{Crafts} & \textbf{Dance} & \textbf{Food} & \textbf{Hobbies} & \textbf{Medical} & \textbf{Sports} & \textbf{Overall} \\ \midrule
         & Random & 33.6 & 26.3 & 26.5 & 26.1 & 27.7 & 28.7 & 25.8 & 25.5 \\
         [0.8ex] \hdashline \\[-1.2ex]

         \multirow{4}{*}{\centering \textit{Closed}} 
         & Gemini 3.1 Pro & 78.5 & \textbf{83.9} & \textbf{58.6} & \textbf{94.1} & 61.9 & 80.3 & 65.7 & \textbf{69.9} \\
         & Gemini 3 Flash & 75.7 & 80.2 & 57.2 & 92.5 & \textbf{63.1} & \textbf{81.3} & 63.4 & 68.7 \\
         & GPT-5.4 & \textbf{81.9} & 74.1 & 54.7 & 92.5 & 60.3 & 77.1 & \textbf{67.6} & 68.0 \\
         & GPT-5 & 79.3 & 76.1 & 53.3 & 93.6 & 60.2 & 76.6 & 65.9 & 67.6 \\
        [0.8ex] \hdashline \\[-1.2ex]
         
        \multirow{2}{*}{\centering \textit{Open Weight}}
         & Qwen3-VL-8B & 42.2 & 44.8 & 32.7 & 74.5 & 40.2 & 60.6 & 43.7 & 45.0 \\
         & InternVL3.5-8B & 44.8 & 40.4 & 33.4 & 74.9 & 40.8 & 56.9 & 41.9 & 44.1 \\
         [0.8ex] \hdashline \\[-1.2ex]
         
        \multirow{3}{*}{\centering \textit{Fully Open}}
        & Molmo2-8B & 53.5 & 40.9 & 33.9 & \underline{78.2} & 41.3 & 59.0 & 41.4 & 44.9 \\
         & Molmo2-4B (base) & 44.8 & 35.7 & 30.3 & 74.7 & 40.2 & 59.0 & 38.7 & 42.0 \\
         & \textcolor{princeton_orange}{Molmo2-4B (FT)} & \underline{66.4} & \underline{59.8} & \underline{52.0} & 69.4 & \underline{52.1} & \underline{64.9} & \underline{44.4} & \underline{53.5} \\
        \bottomrule[1.1pt]
    \end{tabular}
    \label{tab:main_mcq_table}
    }
\end{tcolorbox}
\end{table}

\subsection{Multiple-choice evaluation}

Given a video from a domain and four actions from that domain, we ask a model to choose which of the four actions appears in the video. Example Q\&A pairs are provided in \Cref{fig:teaser}. Random chance is just above 25\%. As detailed in Section \ref{subsec:hard_negatives}, we construct the negative options to minimize the likelihood that multiple of the four provided actions appear in the same video. We experiment with prompt variations that encourage models to explicitly reason before outputting a final answer, but decide to use a simpler prompt after observing a negligible difference in model performance. 

\Cref{tab:main_mcq_table} shows that existing open models struggle with domain-specific action recognition, failing to reach 50\% accuracy on the multiple-choice configuration of \dataset. Meanwhile, the open model fine-tuned on our training data reaches 53.5\%, which is 8.5 percentage points better than the next-best open model. (See \S~\ref{sub_training_result} for details.)

Models appear to cluster by overall performance on \dataset. The existing open 8B models perform within a 1 percentage point range of one other, while the closed models all lie within a 2.5 percentage point range. This clustering makes it difficult to decipher if differences in overall performance among models in a cluster are emblematic of varying model capabilities or simply evaluation noise. 

Disparities in model performance among clusters are, in some cases, starker at the category-level. Molmo2-8B, for instance, is 11.2 and 8.6 percentage points better at ``Beauty'' than Qwen3-VL and InternVL-3.5 respectively. Meanwhile Qwen3-VL maintains a relative edge of 9.5\% and 10.9\% over Molmo2 and InternVL-3.5 in the ``Crafts'' category. These divergent results suggest that open models have varied expertise across domains, implying that systematic evaluations can help identify their specific weaknesses and inform which domains to prioritize in future training.

The ``Food'' category sees high performance across-the-board, suggesting that many of its actions may be identifiable without the need for advanced video understanding. We hypothesize that this is because some Food actions, such as ``Air-Frying'', can be recognized through object detection. It is difficult to assign truly \textit{hard} negatives for such actions (i.e., what other actions would involve an air-fryer?). As models progress, it may be prudent to create a hard subset of \dataset, akin to \cite{mmmu}.

We expect researchers to benchmark their models on the multiple-choice version of \dataset, presented above. We now turn to the binary setting to investigate the effects of changing visual \& textual inputs on model performance. Shifting to the binary setting is necessary because few-shot video examples in a multiple-choice setting would likely overload models. For example, providing 3 in-context examples for each of the 4 actions in a multiple-choice question would yield 12 videos; it is unlikely that models are trained on inputs with 12 in-context videos. Processing 12 videos may also overwhelm model context lengths; InternVL-3.5, for instance, can only handle 64 total input frames.

\subsection{Zero-shot evaluation}
\label{subsec:zero_shot}

Given a video along with an action name and domain, we prompt a model to determine whether or not the video contains the specified action. We use a balanced set of $2$ positive and $2$ negative clips per action. Models are prompted to explicitly reason or analyze the video before providing their answer. We use binary accuracy as our metric, where random chance gets 50\%. 

\begin{table*}[h!]
    \begin{tcolorbox}[
    enhanced,
    width=1.0\linewidth,
    colback=polaris-bg-elevated,
    colframe=polaris-border-subtle,
    arc=3mm,
    boxrule=0.4pt,
    left=8pt,
    right=8pt,
    top=8pt,
    bottom=8pt,
    ]
    \caption{\textbf{Binary 0-shot evaluation results.} Despite a significantly easier evaluation setup, closed models begin to plateau, inspiring the inclusion of in-context examples in Section \ref{subsec:few_shot_eval}. Fine-tuning on our \titleoneclip~dataset, \textcolor{princeton_orange}{Molmo2-4B (FT)} catches up to just 3.7 points shy of Gemini 3 Flash. Highest accuracy for each column is in \textbf{bold}; highest accuracy by a non-proprietary model is \underline{underlined}.}
    \vspace{5pt}
    \centering
    \resizebox{\linewidth}{!}{
    \begin{tabular}{@{}llcccccccc@{}}
        \toprule[1.5pt] 
         & \textbf{Model Name} & \textbf{Beauty} & \textbf{Crafts} & \textbf{Dance} & \textbf{Food} & \textbf{Hobbies} & \textbf{Medical} & \textbf{Sports} & \textbf{Overall} \\ \midrule
         & Random & 50.0 & 50.0 & 50.0 & 50.0 & 50.0 & 50.0 & 50.0 & 50.0 \\
         [0.8ex] \hdashline \\[-1.2ex]

        \multirow{4}{*}{\centering \textit{Closed}} 
         & Gemini 3.1 Pro & 71.6 & \textbf{77.9} & \textbf{68.6} & 91.9 & 66.7 & 78.2 & 68.3 & 72.0 \\
         & Gemini 3 Flash & \textbf{79.3} & 76.3 & 63.3 & \textbf{92.2} & 63.8 & \textbf{79.8} & 67.2 & 70.3 \\
         & GPT-5.4 & 75.9 & 77.2 & 66.4 & 91.7 & 66.1 & 75.0 & \textbf{71.5} & 72.4 \\
         & GPT-5 & 76.7 & 77.7 & 66.2 & 90.6 & \textbf{69.2} & 75.0 & 70.6 & \textbf{72.9} \\
        [0.8ex] \hdashline \\[-1.2ex]
         
        \multirow{2}{*}{\centering \textit{Open Weight}}
         & Qwen3-VL-8B & 62.1 & 60.0 & 54.7 & \underline{79.3} & 55.2 & 58.5 & 58.1 & 59.3 \\
         & InternVL3.5-8B & 58.6 & 54.2 & 52.0 & 75.3 & 55.1 & 58.0 & 57.4 & 57.4 \\
         [0.8ex] \hdashline \\[-1.2ex]
         
        \multirow{3}{*}{\centering \textit{Fully Open}}
         & Molmo2-8B & 56.0 & 55.8 & 53.1 & 78.5 & 53.2 & 61.7 & 51.2 & 55.8 \\
         & Molmo2-4B (base) & 56.0 & 54.0 & 53.1 & 73.1 & 52.3 & 61.2 & 53.0 & 55.3 \\
         & \textcolor{princeton_orange}{Molmo2-4B (FT)} & \underline{73.3} & \underline{69.0} & \underline{66.1} & 73.7 & \underline{67.8} & \underline{69.2} & \underline{61.5} & \underline{66.6} \\
        \bottomrule[1.1pt]
    \end{tabular}
    \label{tab:main_binary_table}
    \label{tab:main_table}
    }
\end{tcolorbox}
\end{table*}

We report our results in \Cref{tab:main_binary_table}. The clustering continues, with open models lagging significantly behind closed models. Once again, our fine-tuned 4B model outperforms all 8B models, achieving 66.6\% accuracy. Within the 8B category, Qwen3-VL beats InternVL3.5 and Molmo2, retaining its top position. Unlike the multiple-choice setting, the GPT models overtake the Gemini models in the binary setting, with GPT-5 achieving the best performance at 72.9\%.

Changing from the multiple-choice setting to the 0-shot binary setting doubles random chance, but only improves the performance of closed models by 3.3 percentage points on average. Similarly, non-expert human performance increases minimally from 68.5\% to 69.1\%. This suggests that other improvements (such as few-shot examples for humans) are necessary to prevent a plateau in performance. On the other hand, open 8B models see a 12.8 percentage point increase on average, suggesting that they have plenty of room for growth in their existing paradigms.

\begin{figure*}
\centering
\begin{tcolorbox}[
    enhanced,
    width=1.0\linewidth,
    colback=polaris-bg-elevated,
    colframe=polaris-border-subtle,
    arc=3mm,
    boxrule=0.4pt,
    left=8pt,
    right=8pt,
    top=8pt,
    bottom=8pt,
]
    \begin{subfigure}[t]{0.44\textwidth}
        \centering
        \includegraphics[width=\textwidth]{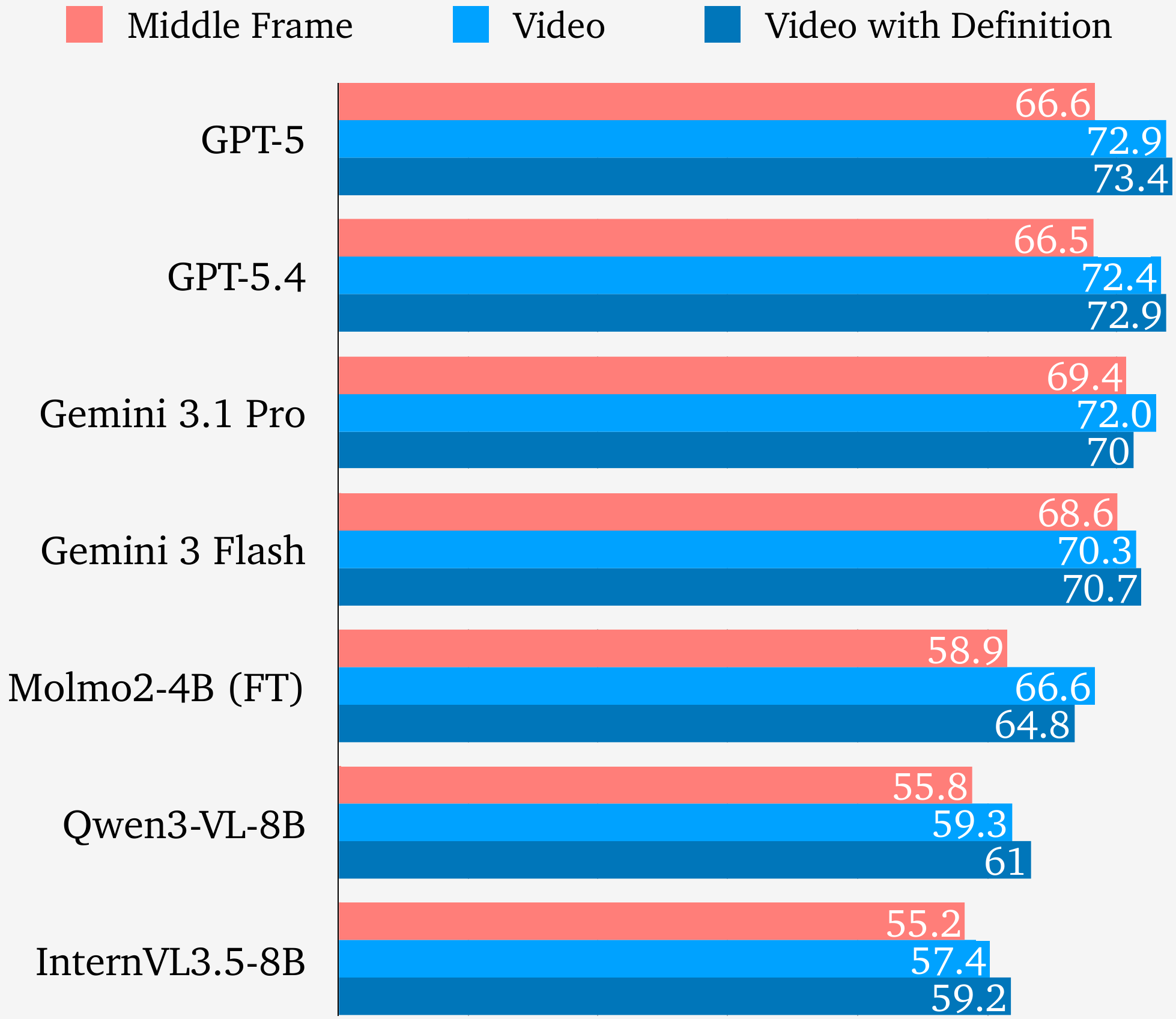}
        \caption{Accuracy of open and proprietary models with different video input settings. Molmo-2 4B (FT) denotes our fine-tuned model using the \titleoneclip~dataset.}
        \label{fig:zeroshot_input_ablations}
    \end{subfigure}
    \hspace{0.03\textwidth}
    \begin{subfigure}[t]{0.53\textwidth}
        \centering
        \includegraphics[width=\textwidth]{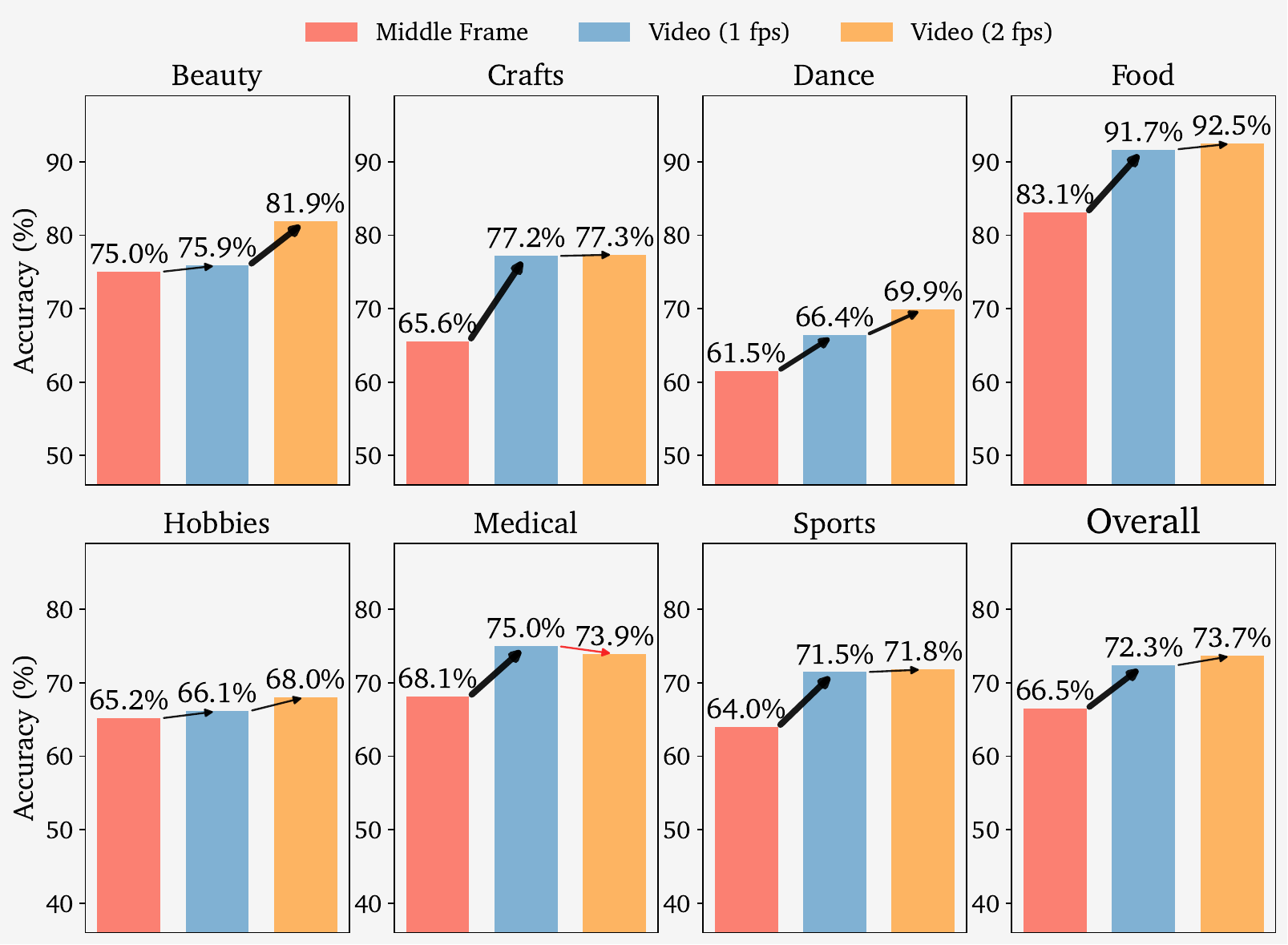}
        \caption{GPT-5.4 accuracy across categories for single frame vs. video at varying frame rates. Black arrows are gains from previous setting with width proportional to relative improvement; red arrows are decreases in performance.}
        \label{fig:gpt4.1_fps_ablations}
    \end{subfigure}

    \caption{\textbf{Ablations on video input configurations} in the binary 0-shot setting. Open models show limited gains from full-video input, indicating difficulty in effectively leveraging video context. (A notable exception is our model, which benefits significantly.)  GPT-5.4 shows only a slight improvement at higher fps, suggesting that test-time scaling via denser video sampling is insufficient for solving domain-specific action recognition.}
    \label{fig:ablations}

\end{tcolorbox}
\end{figure*}

To identify which improvements may help overcome the aforementioned plateau--especially in closed models and humans--we conduct ablation studies. Specifically, we examine whether performance issues arise from insufficient motion understanding or limited action knowledge. \Cref{fig:zeroshot_input_ablations} compares accuracy when provided with (1) a single middle frame, (2) the entire video (default setup), and (3) the video with an action definition (per \S~\ref{subsubsec:definitions}). \Cref{fig:gpt4.1_fps_ablations} shows GPT-5.4 accuracy across categories and varied frame rates. Category-level results for all ablations are available in Appendix \ref{appendix:zero_shot_ablations}. 

\noindent \textbf{Image bias vs. motion understanding}. Existing open-source models show only slight improvements when moving from a single middle frame to the entire video, implying that they struggle to effectively ground actions in detailed motion cues and instead rely heavily on static visual biases (Figure \ref{fig:zeroshot_input_ablations}). In contrast, our fine-tuned model and the GPT models benefit from full-video input, indicating their stronger capability to utilize video information for action recognition.
    
\noindent \textbf{Impact of action definitions}. Figure \ref{fig:zeroshot_input_ablations} shows that providing explicit action definitions yields minimal gains, especially in proprietary models. VLMs appear to already possess sufficient inherent knowledge about actions, likely comparable to expert community sources from the web, and their primary limitation is effectively mapping this knowledge to subtle motion details. 

\noindent \textbf{Higher FPS}. Across action categories, GPT-5.4 significantly improves from single-frame to full-video inputs (Figure \ref{fig:gpt4.1_fps_ablations}). However, increasing the fps further yields diminishing returns, even in motion-intensive categories like Sports, suggesting that models struggle to leverage higher temporal resolution for capturing subtle or rapid motions. 4fps results are available in Table \ref{tab:detailed_fps_0shot}.

\subsection{Few-shot evaluation}
\label{subsec:few_shot_eval}

\begin{figure*}[ht]
    \begin{tcolorbox}[
    enhanced,
    width=1.0\linewidth,
    colback=polaris-bg-elevated,
    colframe=polaris-border-subtle,
    arc=3mm,
    boxrule=0.4pt,
    left=8pt,
    right=8pt,
    top=8pt,
    bottom=8pt,
    ]
        \centering
        \includegraphics[width=\linewidth]{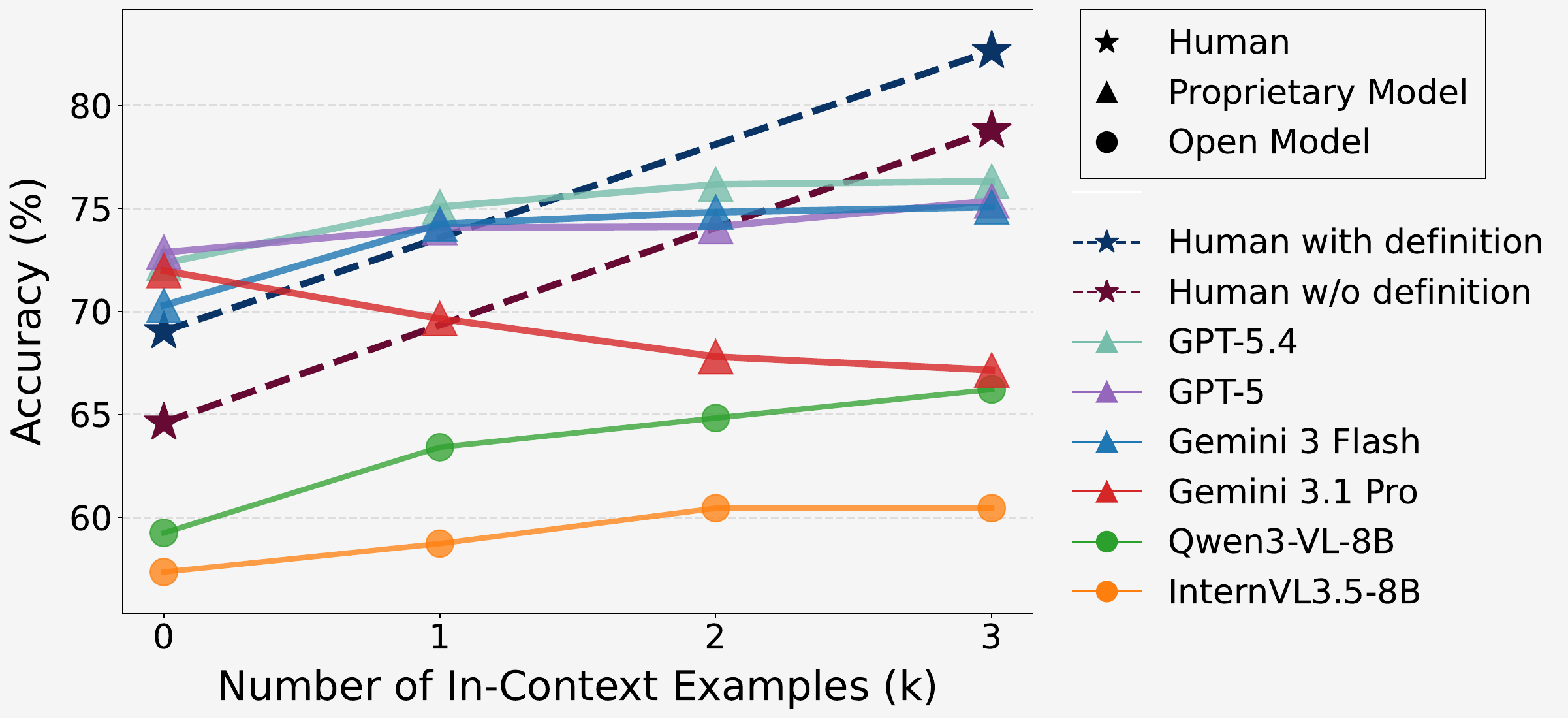}
        \captionof{figure}{\textbf{Binary few-shot accuracy} of VLMs and humans with $k$ in-context video demonstrations. Humans (dotted lines) benefit significantly more than models (solid lines). Among models, there is great variation in their ability to exploit few-shot examples. For example, Gemini 3.1 Pro (red) loses 4.8 percentage points of accuracy while Qwen3-VL (green) gains 7.0 points from $k=0$ to $k=3$ in-context examples.}
        \label{fig:few_shot_plot}
\end{tcolorbox}
\end{figure*}

LLMs excel at learning from textual few-shot examples~\cite{brown2020languagemodelsfewshotlearners, min-etal-2022-metaicl}. We ask whether VLMs similarly excel at learning from visual few-shot examples \cite{kim2024videoicl}. To investigate this, we provide models with $k\in\{1,2,3\}$ example clips of the action in question. These in-context clips are drawn separately from the test set clips used in the binary 0-shot evaluation. The binary test set clips are fixed regardless of how many in-context examples are provided.

\noindent\textbf{VLMs utilize in-context examples with low to medium success.} \Cref{fig:few_shot_plot} shows overall model accuracy as more in-context examples are provided (exact numbers in Appendix \ref{appendix:few_shot_category_level_vlm_results}). Model performance improves on average by 2.95 percentage points and typically by 3.98 points. In the best case, Qwen3-VL improves from 59.3\% to 66.2\% ($+6.9\%$);  in the worst case, Gemini 3.1 Pro declines from 72.0\% to 67.2\% ($-4.8\%$). Surprisingly, Gemini 3 Flash improves from 70.3\% to 75.0\% ($+4.7\%$), despite being older and more lightweight than Gemini 3.1 Pro. The inconsistency in behavior between models suggests that frontier models are still learning to learn from visual few-shot examples. 

In nearly all cases, the biggest jump occurs from $k=0$ to $k=1$. It is not clear if this is because 1 few-shot example provides enough visual context, or if models are unable to effectively utilize additional examples. Overall, the small few-shot gains suggest that video models struggle to fully exploit visual demonstrations. 

\noindent\textbf{Humans are far more effective few-shot learners.}
We sample 698 questions across our benchmark and solicit three human responses for each of these binary questions, taking the majority vote as the final answer (details in Appendix \ref{appendix:human_eval}). \Cref{fig:few_shot_plot} shows that in the zero-shot setup, non-expert humans without action definitions perform worse than proprietary models, likely due to limited domain knowledge. When given definitions, zero-shot human performance improves (+4.5\%) to 69.1\%, close to Gemini 3 Flash. We see the most striking difference in the 3-shot setting, where humans 3-shot with definitions improve significantly to 82.7\% (+13.6\% from 0-shot), while even those without definitions achieve 78.8\% with examples alone. This suggests that humans are highly efficient few-shot learners, quickly generalizing visual patterns from a few demonstrations. The large gap between human and model few-shot performance indicates current VLMs may lack the perceptual mechanisms underlying such human visual learning~\cite{Buccino2004TheMN}.

\noindent \textbf{Random vs. hard negatives.} While non-expert humans achieve high few-shot accuracy (82.7\%), their performance remains imperfect. This is expected, as domain-specific action understanding inherently calls for expert knowledge. In particular, the lack of domain-specific expertise appears to be most problematic when trying to distinguish difficult negatives; few-shot human annotators achieve 94.4\% on positive clips, but only 71.9\% on our hard negative clips.\footnote{A ``positive clip'' is a test clip with the ground truth of ``yes'', i.e., a test clip that \textit{contains} the action that the question inquires about. A ``negative clip'' is a test clip with the ground truth of ``no'', i.e., a test clip that does \textit{not} contain the action that the question inquires about.}

\begin{wraptable}{r}{0.48\linewidth}
    \centering
    \vspace{-0.7cm}
    \begin{tcolorbox}[
    enhanced,
    width=\linewidth,
    colback=polaris-bg-elevated,
    colframe=polaris-border-subtle,
    arc=3mm,
    boxrule=0.4pt,
    left=4pt,
    right=4pt,
    top=6pt,
    bottom=6pt,
]
        \centering
        \caption{\textbf{Accuracy with hard vs. random negatives.} As intended, our selection of hard negatives makes the benchmark more difficult, especially for humans.}
        \resizebox{\linewidth}{!}{
        \begin{tabular}{@{}l | c c | c c@{}}
            \toprule
             \multirow{2}{*}{\textbf{Models}} & \multicolumn{2}{c}{\textbf{Hard}} & \multicolumn{2}{c}{\textbf{Random}} \\
             & k=0 & k=3 & k=0 & k=3 \\
             \midrule
             Gemini 3.1 Pro & 72.0 & 67.2 & 75.4 & 69.2 \\
             Gemini 3 Flash & 70.3 & 75.1 & 78.3 & 80.3\\
             GPT-5.4 & 72.4 & 76.3 & 78.8 & 81.0 \\
             GPT-5 & 72.9 & 75.4 & 79.2 & 78.7 \\
             Qwen3-VL-8B & 59.3 & 66.2 & 62.5 & 71.2 \\
             InternVL3.5-8B & 57.4 & 60.5 & 61.7 & 64.4\\
             \midrule 
             Human w/ defn. & 69.1 & 82.7 & 81.5 & 93.5 \\
            \bottomrule
        \end{tabular}
        }
        \label{tab:random_vs_hard_negatives}
\end{tcolorbox}
\vspace{-1cm}
\end{wraptable}

We compare performance on our hard negatives (default, \S~\ref{subsec:hard_negatives}) vs. random negatives (actions randomly chosen within the same domain), and report results in \Cref{tab:random_vs_hard_negatives}. As expected, random negatives yield consistently higher performance for models and humans in both zero-shot and few-shot settings, with GPT-5.4 (3-shot) reaching 81.0\% and humans achieving 93.5\%. The switch in negatives leads to much higher gains in humans than in models. The accuracy drop from random to hard negatives -- especially for humans -- suggests that \textbf{\dataset~contains challenging, fine-grained visual distinctions that require expertise to solve}.

\subsection{Training Results}
\label{sub_training_result}We finetune a \model~\cite{clark2025molmo2} model on the datasets yielded by the filtering strategies detailed in Section \ref{subsec:training_data}. As reported in \Cref{tab:filter_table}, training with any of our subsets improves results over the base model, substantiating the claim that open models suffer from a lack of domain-specific training data. Using \titleoneclip~as the filtering strategy provides the most gains, improving by $11.5$ percentage points over the base model in the multiple-choice setting. Notably, all of our 4B models beat all existing 8B models in the multiple-choice setting. 

\begin{table*}[h]
    \begin{tcolorbox}[
        enhanced,
        width=1.0\linewidth,
        colback=polaris-bg-elevated,
        colframe=polaris-border-subtle,
        arc=3mm,
        boxrule=0.4pt,
        left=8pt,
        right=8pt,
        top=8pt,
        bottom=8pt,
    ]
    \caption{\textbf{Comparing data filtering strategies.} We compare Molmo2-4B~\cite{clark2025molmo2} models fine-tuned using different data filtering strategies. Observe that the smallest dataset gives considerably better performance, yielding an additional 5 percentage point gain in the multiple-choice setting. The \titleoneclip~binary 0-shot accuracy (66.6\%) even exceeds Qwen3-VL-8B's binary 3-shot accuracy (66.2\%), confirming the effectiveness of domain-specific training data vs. test-time few-shot examples for the domain-specific action recognition task. The delta in a fine-tuned model's accuracy over the base model is in \textcolor{princeton_orange}{(purple)}.}
    \centering
    \resizebox{0.8\textwidth}{!}{
    \setlength{\tabcolsep}{10pt}
    \begin{tabular}{@{}lcll@{}}
    \toprule[1.5pt]
    \multirow{2}{*}{\textbf{Filter}} & \multirow{2}{*}{\textbf{Size}} & \multirow{2}{*}{\textbf{\shortstack{Multiple-Choice\\Accuracy (\%)}}} & \multirow{2}{*}{\textbf{\shortstack{Binary 0-shot\\Accuracy (\%)}}} \\
    & & & \\
    \midrule
    \model~(base) & - & 42.0 & 55.3 \\
    [0.5ex] \hdashline \\[-1.5ex]
    \transcript & 496K & 48.2 \textcolor{princeton_orange}{$(6.2)$} & 65.0 \textcolor{princeton_orange}{$(9.7)$} \\
    \strict & 207K & 48.5 \textcolor{princeton_orange}{$(6.5)$} & 63.7 \textcolor{princeton_orange}{$(8.4)$} \\
    \titleoneclip & 162K & \textbf{53.5} \textcolor{princeton_orange}{\textbf{(}$\boldsymbol{11.5}$\textbf{)}} & \textbf{66.6} \textcolor{princeton_orange}{\textbf{(}$\boldsymbol{11.3}$\textbf{)}} \\
    \bottomrule[1.1pt]
    \end{tabular}
    }
    \label{tab:filter_table}
\end{tcolorbox}
\end{table*}

Granular results are available in Appendix \ref{appendix:data_filtering_strategies}. Tables \ref{tab:filtering_strategy_domain_performance_mcq} and \ref{tab:filtering_strategy_domain_performance_binary} provide per-domain accuracies in the multiple-choice and binary settings. Table \ref{tab:filtering_strategy_category_performance} provides per-category accuracies for both settings.

Our results indicate that the quality of data \textit{generally} influences the performance more than the quantity, since \titleoneclip~is the strictest filter -- in terms of samples selected -- and leads to the highest accuracy. However, for domains in the long-tail, \textit{coverage} becomes an important factor. For instance, \transcript~yields 1,582 clips for juggling, while \titleoneclip~only yields 348 clips. Indeed, the juggling accuracy for the model trained on the former filter surpasses that for the model trained on the latter filter (49.0\% vs. 45.2\%). It is unclear how scale impacts the relative importance of coverage versus quality; we leave this to future work.

\label{subsec:training_results}

\section{Conclusion}
\label{sec:conclusion}
We introduce \dataset, a benchmark to evaluate the domain-specific, fine-grained action understanding of large vision-language models. Our findings reveal that models still have room for improvement in recognizing such actions, both in a standard multiple-choice setting and a relaxed binary 0-shot setting. In order to improve models, we collect a training dataset of automatically-labeled clips of fine-grained, domain-specific actions. Post-training a 4B VLM on this data surpasses all 8B models. We also explore a few-shot evaluation setting where even the best-performing models struggle, implying that current VLMs are not as effective few-shot learners as their text-only counterparts.


\section*{Acknowledgements}

This project was partially funded by a grant from Apple.

\noindent We thank the Hyak and Beaker teams at UW and Ai2 for maintaining their respective compute clusters.

\noindent We thank Oncel Tuzel and Chun-Liang Li for their feedback and guidance.

\noindent We thank members of the UW RAIVN Lab and the Ai2 PRIOR team for insightful discussions and morale boosts, including but not limited to (in alphabetical order) Chris Dongjoo Kim, Etash Guha, Ethan Shen, George Stoica, Haoquan Fang, Jason Lee, Kevin Farhat, Kevin Zhang, Madeline Brumley, Matthew Wallingford, Peter Sushko, and Sarah Pratt. We likewise thank Hayoung Jung.

\noindent We thank Arhan Jain for sharing the template in which this document was typeset.

\clearpage
{
    \small
    \bibliographystyle{unsrt} 
    \bibliography{references.bib}
}

\clearpage

\beginappendix{
\appendix

\label{sec:appendix_table_of_contents}
\vspace{5pt}
This Appendix contains the following sections:
\vspace{5pt}
\begin{itemize}[leftmargin=1cm]
    \itemsep0.3em
    \item \S~\ref{appendix:benchmark_statistics} - \textbf{Benchmark statistics}; discusses \dataset's inter-domain breadth and intra-domain depth, the latter in comparison to existing works. 
    \item \S~\ref{appendix:benchmark_collection} - \textbf{Benchmark collection}; prints LLM prompts and UIs used during benchmark construction.
    \item \S~\ref{appendix:eval} - \textbf{Model evaluation}; details on how we evaluated existing models on the \dataset~benchmark (prompts, video sampling, model versions, etc.).
    \item \S~\ref{appendix:zero_shot_ablations} - \textbf{Zero-shot ablations}; detailed results for the ablations shown in \Cref{fig:ablations}. 
    \item \S~\ref{appendix:few_shot} - \textbf{Few-shot results}; detailed results for models in the few-shot setting. Additional results for CLIP models and optical flow models. Discussion of prompt-sensitivity in Gemini and the impact of few-shot examples on yes/no bias. 
    \item \S~\ref{appendix:human_eval} - \textbf{Human evaluation}; details on the human evaluation setup. In-depth human evaluation results.
    \item \S~\ref{appendix:additional_training_details} - \textbf{Additional training details}; construction of VQA pairs from labeled video clips. Listing of learning rates, image pooling, etc.
    \item \S~\ref{appendix:data_filtering_strategies} - \textbf{Data filtering strategies}; description of and motivation behind filtering strategies. Analysis of differences in downstream performance on \dataset~benchmark when different filters are applied. Per-domain results of our fine-tuned models.
\end{itemize}

\vspace{10pt}
\clearpage
\section{Benchmark Statistics}
\label{appendix:benchmark_statistics}

Given that previous domain-specific benchmarks (e.g., \cite{finegym, finediving, fine_figure_skate, finesports, finegrained_novel_basketball}, see \Cref{sec:related_work}) have chosen to sacrifice breadth for depth, it is natural to ask whether \dataset~inevitably sacrifices depth for breadth. As shown in \Cref{tab:benchmark_depth}, \dataset~achieves greater depth in many of the domains it covers when compared to previous one-domain works.

\begin{table*}[!h]
\begin{tcolorbox}[
    enhanced,
    width=\linewidth,
    colback=polaris-bg-elevated,
    colframe=polaris-border-subtle,
    arc=3mm,
    boxrule=0.4pt,
    left=8pt,
    right=8pt,
    top=8pt,
    bottom=8pt,
]
    \centering
    \caption{\textbf{Depth of \dataset.} The last two columns report, for a given domain, the \# of actions in other benchmarks and the \# of actions in \dataset~respectively. When compared to domain-specific benchmarks that focus on fewer domains, it is clear that \dataset~maintains sufficient depth in the domains it covers. (Many values in the second-to-last column sourced from Table 1 in \cite{finesports}.)}
    \label{tab:benchmark_depth}
    \resizebox{0.7\linewidth}{!}{
    \begin{tabular}{ccccc}
        \toprule[1.5pt]
        \textbf{Domain} & \textbf{Paper Name} & \textbf{Paper Venue} & \textbf{Theirs} & \textbf{Ours} \\
        \midrule 
        \multirow{5}{*}{Figure Skating}
        & MCFS \cite{mcfs} & AAAI 2021 & 130 & \multirow{5}{*}{40} \\
        & MMFS \cite{fine_figure_skate} & arXiv & 46 \\
        & FSBench \cite{fsbench} & CVPR 2025 & 20 \\
        & Fis-V \cite{score_figure_skate} & TCSVT 2020 & 13 \\
        & FSD-10 \cite{fsd_10} & Neurocomputing & 10 \\
        [1ex] \hdashline \\[-1ex]
        \multirow{4}{*}{Basketball}
        & FineSports \cite{finesports} & CVPR 2024 & 52 & \multirow{4}{*}{45} \\
        & Basket \cite{basket} & CVPR 2025 & 20 & \\
        & \textit{Basketball} \cite{finegrained_novel_basketball} & ICASSP 2020 & 27 \\
        & MultiSports \cite{multisports} & ICCV 2021 & 18\\
        [1ex] \hdashline \\[-1ex]
        \multirow{2}{*}{Soccer}
        & MultiSports \cite{multisports} & ICCV 2021 & 21 & \multirow{2}{*}{40} \\
        & SoccerNet \cite{soccernet} & CVPR 2018 & 3 \\
        \bottomrule[1.5pt]
    \end{tabular}
    }
\end{tcolorbox}
\end{table*}

For the \dataset~benchmark, we release 5,000 clips spanning 37 domains within 7 categories. \Cref{tab:benchmark_per_domain} provides a breakdown of each domain's category, number of actions, number of clips, and the length of these clips.

\begin{table}[h!]
    \begin{tcolorbox}[
    enhanced,
    width=\linewidth,
    colback=polaris-bg-elevated,
    colframe=polaris-border-subtle,
    arc=3mm,
    boxrule=0.4pt,
    left=8pt,
    right=8pt,
    top=8pt,
    bottom=8pt,
]
    \centering
    \caption{\textbf{Benchmark Video Duration (seconds).} The clips are well-trimmed, meaning that they contain the entirety of the action and minimal ``fluff" around it. 
    We removed clips longer than 5 minutes after observing open models struggle with context lengths for long videos, especially in the 3-shot setting.}
    \vspace{-2pt}
    \resizebox{0.6\linewidth}{!}{
    \begin{tabular}{cccccc}
        \toprule
        \textbf{Min} & \textbf{Max} & \textbf{Mean}  & \textbf{Median} & \textbf{Standard Deviation} \\
        \midrule
        0.26 & 260.0 & 12.2 & 5.0 & 21.5 \\
    \bottomrule
    \end{tabular}
    \label{tab:benchmark_clip_stats}
    }
\end{tcolorbox}
\end{table}

\begin{table*}[h!]
    \begin{tcolorbox}[
    enhanced,
    width=\linewidth,
    colback=polaris-bg-elevated,
    colframe=polaris-border-subtle,
    arc=3mm,
    boxrule=0.4pt,
    left=8pt,
    right=8pt,
    top=8pt,
    bottom=8pt,
]
    \centering
    \caption{\textbf{Actions and Clips per Domain.} We report the number of actions and number of clips in our benchmark for each domain, as well as statistics on the length of clips in each domain. Entries are ordered alphabetically, left-to-right.}
    \vspace{10pt}
    \label{tab:benchmark_per_domain}
    \resizebox{0.9\textwidth}{!}{
    \begin{tabular}{@{}llcccc@{}}
        \toprule[1.5pt]
        \multirow{2}{*}{\textbf{Category Name}} & \multirow{2}{*}{\textbf{Domain Name}} & \multirow{2}{*}{\textbf{\# Actions}} & \multirow{2}{*}{\textbf{\# Clips}} & \multicolumn{2}{c}{\textbf{Clip Duration (s)}} \\
        & & & & \textbf{Mean} & \textbf{Median} \\
        \midrule 
        \multirow{3}{*}{\centering Beauty \& Self Care} 
        & Hairstyling & 12 & 60 & 6.7 & 4.0 \\
        & Spa Massage & 11 & 55 & 20.7 & 11.0 \\
        & Tattooing & 6 & 30 & 10.1 & 6.1 \\
        \multirow{8}{*}{\centering Crafts \& Art} \\
        [-1.8ex] \hdashline \\[-1.2ex]
        & Calligraphy & 8 & 40 & 4.7 & 3.7 \\
        & Crochet & 15 & 75 & 36.1 & 26.5 \\
        & Hand Sewing / Embroidery & 38 & 190 & 42.4 & 28.0 \\
        & Knots & 55 & 275 & 42.9 & 36.0 \\
        & Painting & 8 & 40 & 23.7 & 10.0 \\
        & Pottery & 10 & 50 & 27.1 & 12.0 \\
        & Woodworking / Whittling & 4 & 20 & 8.3 & 6.4 \\
        \multirow{6}{*}{\centering Dance} \\
        [-1.8ex] \hdashline \\[-1.2ex]
        & Ballet & 39 & 195 & 4.9 & 3.5 \\
        & Bharatanatyam & 18 & 90 & 16.3 & 11.0 \\ 
        & Break Dance & 33 & 165 & 6.8 & 5.0\\ 
        & Salsa & 19 & 95 & 6.7 & 5.0 \\
        & Tap Dance & 28 & 140 & 6.4 & 4.0 \\
        \multirow{4}{*}{\centering Food \& Beverage} \\
        [-1.8ex] \hdashline \\[-1.2ex]
        & Bartending & 28 & 140 & 10.2 & 4.7 \\
        & Coffee & 16 & 80 & 10.5 & 8.0 \\
        & Cooking & 49 & 245 & 17.4 & 10.8 \\
        \multirow{9}{*}{\centering Hobbies} \\
        [-1.8ex] \hdashline \\ [-1.2ex]
        & Bouldering & 22 & 110 & 7.0 & 5.0 \\
        & Gardening & 20 & 100 & 22.0 & 12.0 \\
        & Gym & 22 & 110 & 5.4 & 4.0 \\
        & Juggling & 26 & 130 & 6.5 & 4.3 \\
        & Parkour & 40 & 200 & 4.4 & 3.0 \\
        & Pen Spinning & 32 & 160 & 3.6 & 3.0 \\
        & Skateboarding & 47 & 235 & 4.5 & 3.5 \\
        & Yo-yo & 50 & 250 & 6.8 & 5.0 \\
        \multirow{4}{*}{\centering Medical} \\
        [-1.8ex] \hdashline \\ [-1.2ex]
        & Neurological Abnormalities & 20 & 100 & 11.2 & 8.0  \\
        & Neurological Assessments & 15 & 75 & 9.4 & 7.0 \\
        & Suturing & 12 & 60 & 72.6 & 63.0 \\
        \multirow{9}{*}{\centering Sports} \\
        [-1.8ex] \hdashline \\ [-1.2ex]
        & American Football & 54 & 270 & 6.4 & 5.3 \\
        & Basketball & 45 & 225 & 4.3 & 3.6 \\
        & Cheerleading & 20 & 100 & 6.1 & 5.0 \\
        & Cricket & 41 & 205 & 4.5 & 4.0 \\
        & Figure Skating & 40 & 200 & 5.4 & 4.0 \\
        & Ice Hockey & 38 & 190 & 4.5 & 4.0 \\
        & Soccer & 40 & 200 & 3.8 & 3.2 \\
        & Tennis & 19 & 95 & 4.4 & 3.0 \\
        \multirow{2}{*}{} \\
        [-1.5ex] \hdashline \\ [-1.8ex]
        \textit{All} & \textit{All} & 1,000 & 5,000 & 12.2 & 5.0 \\
     \bottomrule[1.5pt]
    \end{tabular}
    }
\end{tcolorbox}
\end{table*}

Basic benchmark-wide statistics on video duration are provided in \Cref{tab:benchmark_clip_stats}. Here we emphasize the long-tail nature of video lengths in \dataset. This is caused by a handful of domains having much lengthier clips than most. For instance, the median length of a knots clip and a suturing clip are 36 seconds and 63 seconds respectively (see \Cref{tab:benchmark_per_domain}). Concretely, the kurtosis of video durations in \dataset~is 34.1, indicating a heavy tail.\footnote{We report the Pearson kurtosis, not the Fisher/excess kurtosis. For reference, the Pearson kurtosis of the normal distribution is 3.} The long tail is made evident by \Cref{fig:benchmark_duration_graphs}.

\begin{figure}[!htbp]
    \begin{tcolorbox}[
    enhanced,
    width=1.0\linewidth,
    colback=polaris-bg-elevated,
    colframe=polaris-border-subtle,
    arc=3mm,
    boxrule=0.4pt,
    left=8pt,
    right=8pt,
    top=8pt,
    bottom=8pt,
    ]
    \centering
    \begin{minipage}{0.49\linewidth}
        \centering
        \includegraphics[width=\linewidth]{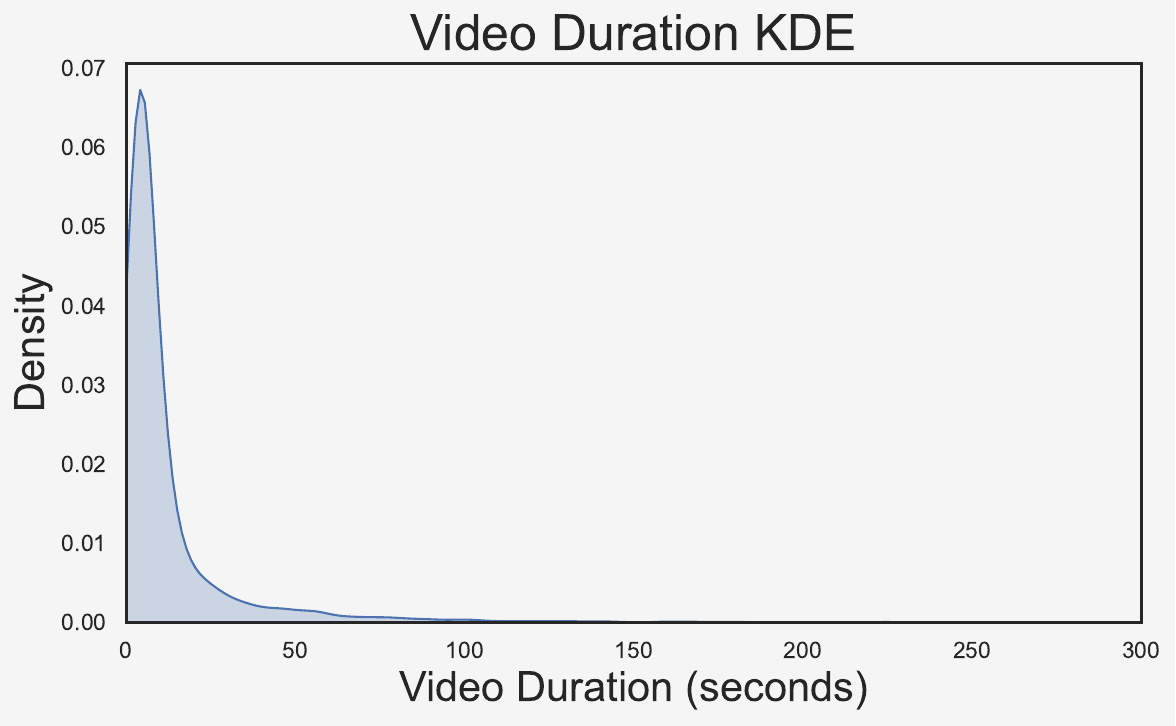}
    \end{minipage}
    \hfill
    \begin{minipage}{0.49\linewidth}
        \centering
        \includegraphics[width=\linewidth]{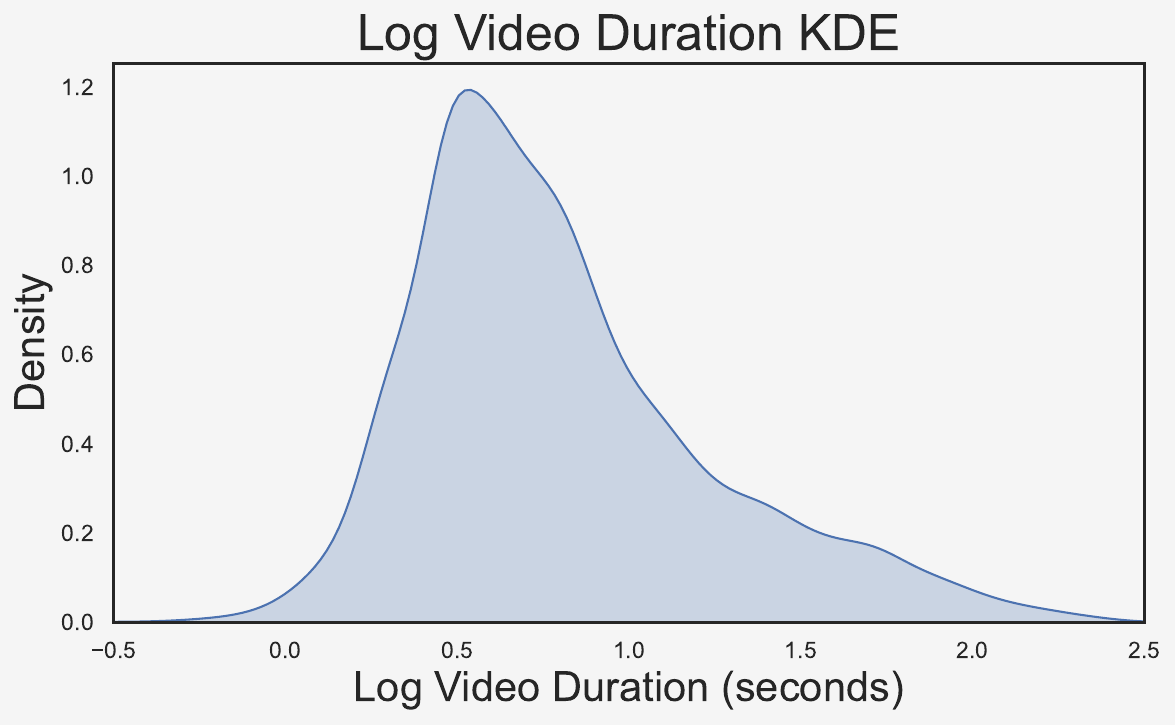}
    \end{minipage}
    \caption{\textbf{Kernel Density Estation for \dataset~Clip Durations.} The left graph clearly shows a long tail, but it is difficult to analyze due to the right tail's sheer length and the concentration of near-0 durations. For closer inspection, the right graph uses log duration. The smoothing bandwidth is determined by Scott's Rule.}
    \label{fig:benchmark_duration_graphs}
\end{tcolorbox}
\end{figure}
\clearpage
\clearpage
\section{Benchmark Collection}
\label{appendix:benchmark_collection}
\subsection{LLM Augmentation of Action Lists}
\label{appendix:action_list_llm}
    After collecting initial action lists from expert online sources, we expand them with Claude as specified in \Cref{fig:action_list_augmentation_prompt}.

\subsection{LLM Deduplication of Action Lists}
\label{appendix:action_deduplication_llm}
    We then de-duplicate the action lists. Note that the LLM's response is only taken as a suggestion -- the authors manually review duplicate actions identified by the LLM to decide if they are true duplicates or not. To preserve the integrity of our negatives and improve the fine-grained nature of our benchmark, if the action list has a general action (e.g., dunk) and many varieties of that action (e.g., tomahwak dunk, windmill dunk, alley-oop dunk), we remove the former and keep the latter. Refer to \Cref{fig:action_deduplication_prompt} for the prompt.
    
\subsection{LLM Generation of Action Definitions with Web-Search}
    \label{appendix:action_def_llm}
    We walk through our action definition generation pipeline as discussed earlier in \S~\ref{subsubsec:definitions}.
    
    Initially, our pilot annotation study revealed that annotators had trouble correctly identifying actions when provided only with action labels, mainly due to their lack of domain-specific knowledge; based on their feedback, they struggled to ground the performed action in video and distinguish the accurate actions from incorrect ones. This initial setup resulted in numerous inaccurately labeled video clips. 
    
    To address this knowledge gap, we provide explicit action definitions describing the visual characteristics of each action using layman's terms. We design these definitions to be a stand-alone resource, thereby removing the need for annotators to locate external references. 
    We use an LLM, Claude-3.7, with web-search capabilities to generate accurate action definitions informed by expert online communities. For each domain, we provide all actions at once and ensure the definitions satisfy the following conditions: they avoid overlap and do not reference other actions' definitions; they clearly elaborate on basic, atomic actions to minimize jargon, particularly for actions involving combinations of simpler actions; and they mention key differences from similar actions in the same list to prevent confusion.
    
    We observe that providing action definitions during the annotation stage significantly helps non-expert humans in understanding the action. These improvements are further supported by the human evaluation results presented in Figure~\ref{fig:few_shot_plot}. We provide our exact prompt in \Cref{fig:action_defn_prompt}.

\subsection{LLM Generated Hard Negatives}
\label{appendix:action_hard_negatives_llm}
        Figures~\ref{fig:hard-neg-prompt-system}-\ref{fig:hard-neg-prompt-stage4} present the prompts and LLM generation parameters used to create the hard negatives described in \S~\ref{subsec:hard_negatives}. In the first stage, we use \texttt{gpt-4.5-preview} to create an initial balanced set of hard negative candidates (Figure~\ref{fig:hard-neg-prompt-stage1}). In later stages, we use \texttt{o3-2025-04-16} to iteratively refine the negatives by 1) correcting false negatives that may co-occur with the positive actions, 2) diversifying the selection patterns by incorporating negatives with varying types of visual similarity, and 3) ensuring each action appears as a hard negative with balanced frequency (Figures~\ref{fig:hard-neg-prompt-stage2}-\ref{fig:hard-neg-prompt-stage4}).

\subsection{Human Annotator UIs}
    \label{appendix:benchmark_annotator_uis}
    Figures \ref{fig:collection_ui}, \ref{fig:verification_ui}, and \ref{fig:trimming_ui} contain the user interfaces shown to human annotators during the collection, verification, and trimming stages respectively. For full reproducibility, the HTML/CSS will be made available on our GitHub repository. Annotators were paid \$15-\$17 per hour for their efforts.

\subsection{Sourcing Human Annotators}

We begin with two pools of approximately 1000 and 50 human annotators. The annotators in these pools have done ``good'' and ``exemplary'' jobs, respectively, in previous Prolific studies hosted by the authors.\footnote{Prolific is a crowd-sourcing platform.}

(It may be helpful to review the annotation stages shown in \Cref{fig:collection}.) All annotators from the first pool are invited to complete Stage 1 (clip collection) on a small subset of domains (we later re-collected the data for this subset after we had filtered a set of ``great'' annotators). We then asked the second pool, in whom we had high confidence, to complete Stage 2 (clip verification). We kept the top one-fifth of annotators, as determined by the percentage of “yes” votes the clips they collected in Stage 1 received during the verification process in Stage 2. This newly-derived pool of approximately 200 annotators was used to collect clips for the \dataset~benchmark. 

\vspace{20pt}

\begin{figure*}[!htbp]
\begin{tcolorbox}[
    enhanced,
    width=1.0\linewidth,
    colback=polaris-bg-elevated,
    colframe=polaris-border-subtle,
    arc=3mm,
    boxrule=0.4pt,
    left=4pt,
    right=4pt,
    top=6pt,
    bottom=6pt,
]
\begin{verbatim}
I have the following list of <DOMAIN> actions:

<INITIAL ACTION LIST>

Provide me with suggestions of <DOMAIN> actions that are well-defined and
highly-discernible. Your suggestions should not overlap with each other, nor
should they overlap with any of the <DOMAIN> actions on the list I provided.\end{verbatim}
\caption{\textbf{Action list augmentation prompt.}}
\label{fig:action_list_augmentation_prompt}
\end{tcolorbox}
\end{figure*}

\begin{figure*}[!htbp]
\begin{tcolorbox}[
    enhanced,
    width=1.0\linewidth,
    colback=polaris-bg-elevated,
    colframe=polaris-border-subtle,
    arc=3mm,
    boxrule=0.4pt,
    left=4pt,
    right=4pt,
    top=6pt,
    bottom=6pt,
]
    \begin{verbatim}
Are there any duplicates or near-duplicates in this list of <DOMAIN> actions?
    \end{verbatim}
  \caption{\textbf{Action deduplication prompt.}}
  \label{fig:action_deduplication_prompt}
\end{tcolorbox}
\end{figure*}

\begin{figure*}[!htbp]
\begin{tcolorbox}[
    enhanced,
    width=1.0\linewidth,
    colback=polaris-bg-elevated,
    colframe=polaris-border-subtle,
    arc=3mm,
    boxrule=0.4pt,
    left=4pt,
    right=4pt,
    top=6pt,
    bottom=6pt,
]
    \begin{verbatim}
Generate detailed definitions for the following <DOMAIN> actions from 
<CATEGORY> category. Each definition should:

Be completely self-contained and understandable without referencing other 
actions. Explain any specialized terminology within the definition (using 
phrases like 
"which is..." or "meaning...")
Include visual identification cues (what to look for to recognize the action)
Describe how this action differs from similar actions when applicable.
Be written for a general audience with no prior knowledge of the domain.

Format each definition as:
[ACTION NAME]: [Complete definition with all elements above]
Use web search to gather accurate information about these actions, but DO NOT 
include source links or citations in your final output. The goal is to create 
clean, comprehensive definitions that can be easily copied into a spreadsheet 
or database.

Here are the actions to define:
<ACTION LIST>
Remember that each definition must stand alone since readers may only see 
one definition at a time.
    \end{verbatim}
  \caption{\textbf{Action definition prompt.}}
  \label{fig:action_defn_prompt}
\end{tcolorbox}
\end{figure*}
\begin{figure*}[!htbp]
\begin{tcolorbox}[
    enhanced,
    width=1.0\linewidth,
    colback=polaris-bg-elevated,
    colframe=polaris-border-subtle,
    arc=3mm,
    boxrule=0.4pt,
    left=4pt,
    right=4pt,
    top=6pt,
    bottom=6pt,
]
\underline{\textbf{System Prompt}}
\vspace{5pt}
\begin{verbatim}
  You are creating challenging "hard negative" options for multimodal 
  action classification datasets across various domains 
  (sports, arts, crafts, cooking, etc.).
  
  Each action requires 3 hard negative options that are genuinely difficult 
  for a machine learning model to distinguish from the positive action.

  A truly "hard" negative:
  - Shares visual/motion similarities with the positive action that would 
    be difficult to distinguish in brief clips
  - Is fundamentally different in purpose or technique despite visual 
    similarities
  - Cannot reasonably co-occur with the positive action in the same short 
    video
  - Avoids obvious selection patterns that would make classification too 
    easy

  Note:
    Negatives should only come from the action list provided (not 
    definitions or other sources)
      - Check that EXACT positive and negative action names are used 
        in the actions list when generating csv.
  \end{verbatim}
  
  \caption{\textbf{System prompt for hard negative generation}}
  \label{fig:hard-neg-prompt-system}
\end{tcolorbox}
\end{figure*}

\begin{figure*}[!htbp]
\begin{tcolorbox}[
    enhanced,
    width=1.0\linewidth,
    colback=polaris-bg-elevated,
    colframe=polaris-border-subtle,
    arc=3mm,
    boxrule=0.4pt,
    left=4pt,
    right=4pt,
    top=6pt,
    bottom=6pt,
]
    \underline{\textbf{User Prompt}}
    \vspace{5pt}
  \begin{small}
  \begin{verbatim}
  Below is my list of "<ACTION>" actions, along with their definitions:
  <ACTION DEFINITION>
  ===
  Your task is to create genuinely challenging "hard negative" options for each action
  that would confuse a computer vision model. Format your output as a clean CSV:
  
  action,negative_1,negative_2,negative_3
  (action_1),(hard negative_1),(hard negative_2),(hard negative_3)
  ...
  CRITICAL REQUIREMENTS FOR TRULY HARD NEGATIVES:

  1. MAXIMIZE VISUAL CONFUSION WITHOUT OBVIOUS PATTERNS:
     - Select actions that share visual features, body positions, or motion qualities with the
       positive action
     - Avoid predictable selection patterns (e.g., don't always choose the
       "next level up/down" or "same family" actions)
     - Mix selection criteria unpredictably to prevent the model from learning simple heuristics
     
  2. STRATEGIC AMBIGUITY:
     - Include some negatives that differ in subtle ways (small variations in technique/position)
     - Include some negatives that differ in more significant ways but still maintain visual
       similarity
     - Vary the type of similarity (sometimes motion-based, sometimes position-based, sometimes
       tool/environment-based)
       
  3. AVOID FUNCTIONALLY RELATED ACTIONS FOR NEGATIVES:
     - Never select actions that typically occur together with the positive action
     - Avoid actions that are commonly performed in sequence or as part of the same technique
     - Don't pair actions that would naturally appear in the same short video clip
     - Don't pair action categories that are too similar or the same as the positive action
       
  4. REASONABLE DISTRIBUTION:
     - Each action should appear as a negative approximately 2-5 times across the dataset
     - Avoid extreme over-representation or under-representation
     - The overall pattern of selections should appear random and unpredictable

  Please provide your hard negative choices for these actions in the 
  same order as provided:
  <ACTION LIST>
  Negatives should only EXACTLY come from the action list provided 
  (not definitions or made-up sources)
  - Check that EXACT positive and negative action names are used in 
    the actions list when generating csv.
\end{verbatim}
\end{small}
\vspace{10pt}
\underline{\textbf{API Details}}
\vspace{5pt}
\begin{small}
\begin{verbatim}
  model: gpt-4.5-preview-2025-02-27
  temperature: 0.5
  max_tokens: 4096
  \end{verbatim}
\end{small}
  \caption{\textbf{First user prompt for hard-negative generation} $(1/4)$}
  \label{fig:hard-neg-prompt-stage1}
\end{tcolorbox}
\end{figure*}

\begin{figure*}[!htbp]
\begin{tcolorbox}[
    enhanced,
    width=1.0\linewidth,
    colback=polaris-bg-elevated,
    colframe=polaris-border-subtle,
    arc=3mm,
    boxrule=0.4pt,
    left=4pt,
    right=4pt,
    top=6pt,
    bottom=6pt,
]
    \underline{\textbf{User Prompt}}
    \vspace{5pt}
    \begin{verbatim}
  Please provide your analysis of negative selections for 
  their effectiveness as genuinely "hard" negatives:

  First, check for selection patterns that could make classification too 
  easy:
  - Are there predictable patterns in how negatives were selected?
  - Is there too much consistency in how negatives relate to positives?
  - Would these patterns potentially provide shortcuts for a 
  classification model?

  Second, examine the visual confusion potential:
  - How visually similar are the negatives to their positive actions?
  - Is there sufficient variety in the types of visual similarity?
  - Are the differences appropriately subtle to create genuine challenges?

  Third, check for functional relationships:
  - Are there any positive-negative pairs that typically occur together?
  - Are there pairs that represent sequential or component actions?
  - Would any pairs likely appear together in a short video clip?

  Finally, review the overall distribution:
  - Is any action severely over-represented or under-represented as a 
    negative?
  - Does the selection appear sufficiently unpredictable and varied?
  - Are there imbalances that should be addressed?

  For any issues identified, suggest specific improvements to create more 
  genuinely challenging hard negatives.
  Provide a summary of the analysis and suggestions for improvement.
\end{verbatim}
\vspace{10pt}
\underline{\textbf{API Details}}
\vspace{5pt}
\begin{verbatim}
  model: o3-2025-04-16
  reasoning effort: high
\end{verbatim}
  \caption{\textbf{Second user prompt for hard-negative generation} $(2/4)$}
  \label{fig:hard-neg-prompt-stage2}
\end{tcolorbox}
\end{figure*}

\begin{figure*}[!htbp]
\begin{tcolorbox}[
    enhanced,
    width=1.0\linewidth,
    colback=polaris-bg-elevated,
    colframe=polaris-border-subtle,
    arc=3mm,
    boxrule=0.4pt,
    left=4pt,
    right=4pt,
    top=6pt,
    bottom=6pt,
]
\underline{\textbf{User Prompt}}
\vspace{5pt}
\begin{verbatim}
    Based on the analysis, provide a revised CSV with improved hard negatives. 
    
    Focus on fixing:
      1. The most problematic selection patterns identified
      2. Any actions with co-occurring negatives
      3. Distribution imbalances

    Briefly explain the changes made to each action's negatives, 
    ensuring that the new selections are genuinely challenging 
    and visually confusing.
    Then, provide the revised CSV with fixed negative selections,
    without detailed explanations for each change.
\end{verbatim}
\vspace{10pt}
\underline{\textbf{API Details}}
\vspace{5pt}
\begin{verbatim}
    model: o3-2025-04-16
    reasoning effort: high
\end{verbatim}
    \caption{\textbf{Third user prompt for hard-negative generation} $(3/4)$}
    \label{fig:hard-neg-prompt-stage3}
\end{tcolorbox}
\end{figure*}

\begin{figure*}[!htbp]
\begin{tcolorbox}[
    enhanced,
    width=1.0\linewidth,
    colback=polaris-bg-elevated,
    colframe=polaris-border-subtle,
    arc=3mm,
    boxrule=0.4pt,
    left=4pt,
    right=4pt,
    top=6pt,
    bottom=6pt,
]
    \underline{\textbf{User Prompt}}
    \vspace{5pt}
    \begin{verbatim}
    Based on the comprehensive analysis and specific suggestions, 
    synthesize a final CSV with truly challenging hard negatives
    for each action.
    
    Incorporate all the suggested improvements while ensuring:
      1. The final list follows the exact same order as the original 
         action list
      2. Each action has 3 negatives that create genuine visual confusion
      3. The selection patterns remain unpredictable and varied
      4. No functionally related actions are paired
      5. The distribution is reasonably balanced 
         (each action appears 2-5 times as a negative)
    
    Provide the final clean CSV with optimized hard negatives:
    \end{verbatim}
    \vspace{10pt}
    \underline{\textbf{API Details}}
    \vspace{5pt}
    \begin{verbatim}
  model: gpt-4.5-preview-2025-02-27
  temperature: 0.5
  max_tokens: 4096
    \end{verbatim}
    \caption{\textbf{Final user prompt for hard-negative generation} $(4/4)$}
    \label{fig:hard-neg-prompt-stage4}
\end{tcolorbox}
\end{figure*}

\begin{figure*}[tbp]
    \centering
    \includegraphics[width=0.6\linewidth]{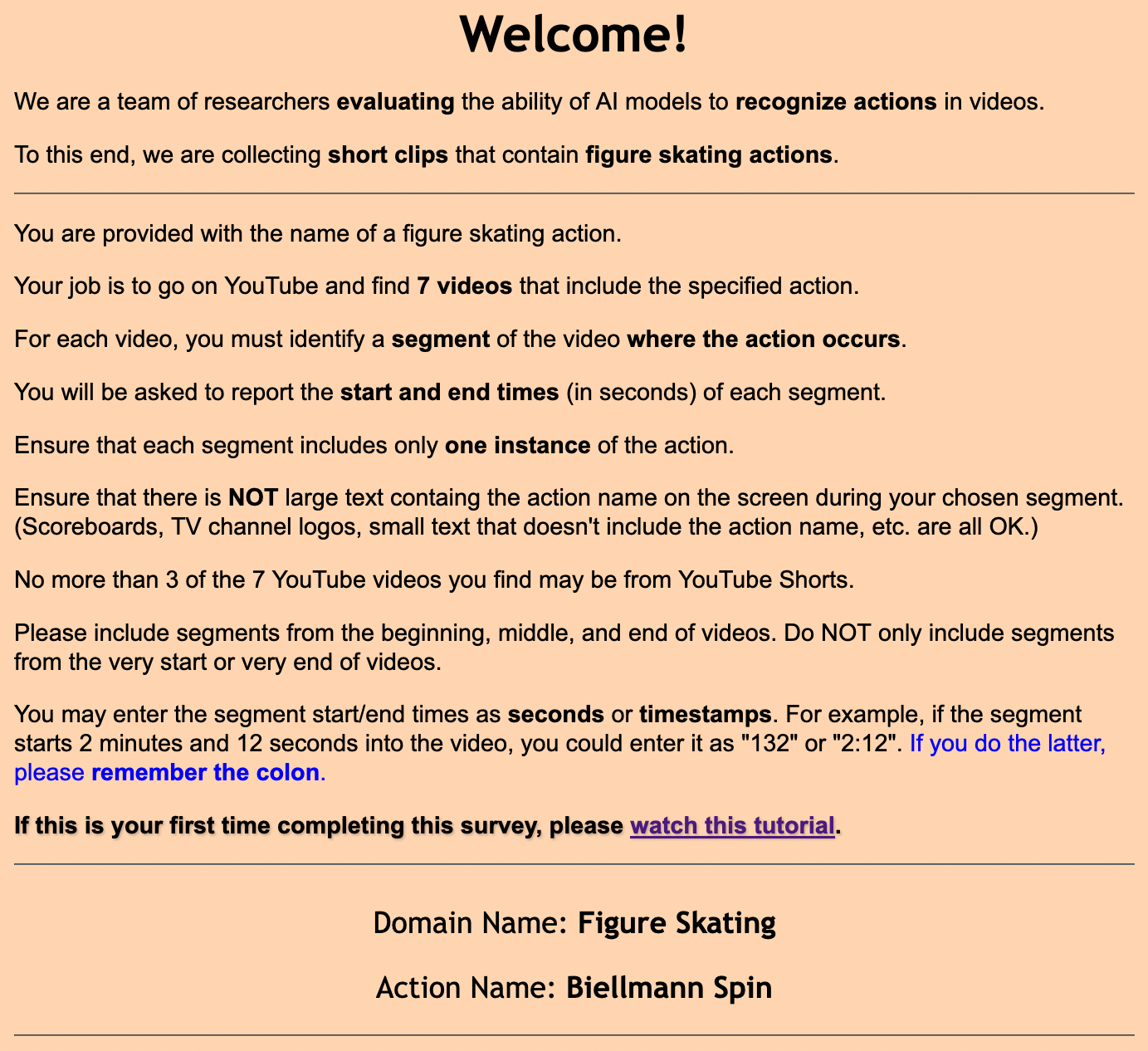}
    \includegraphics[width=0.6\linewidth]{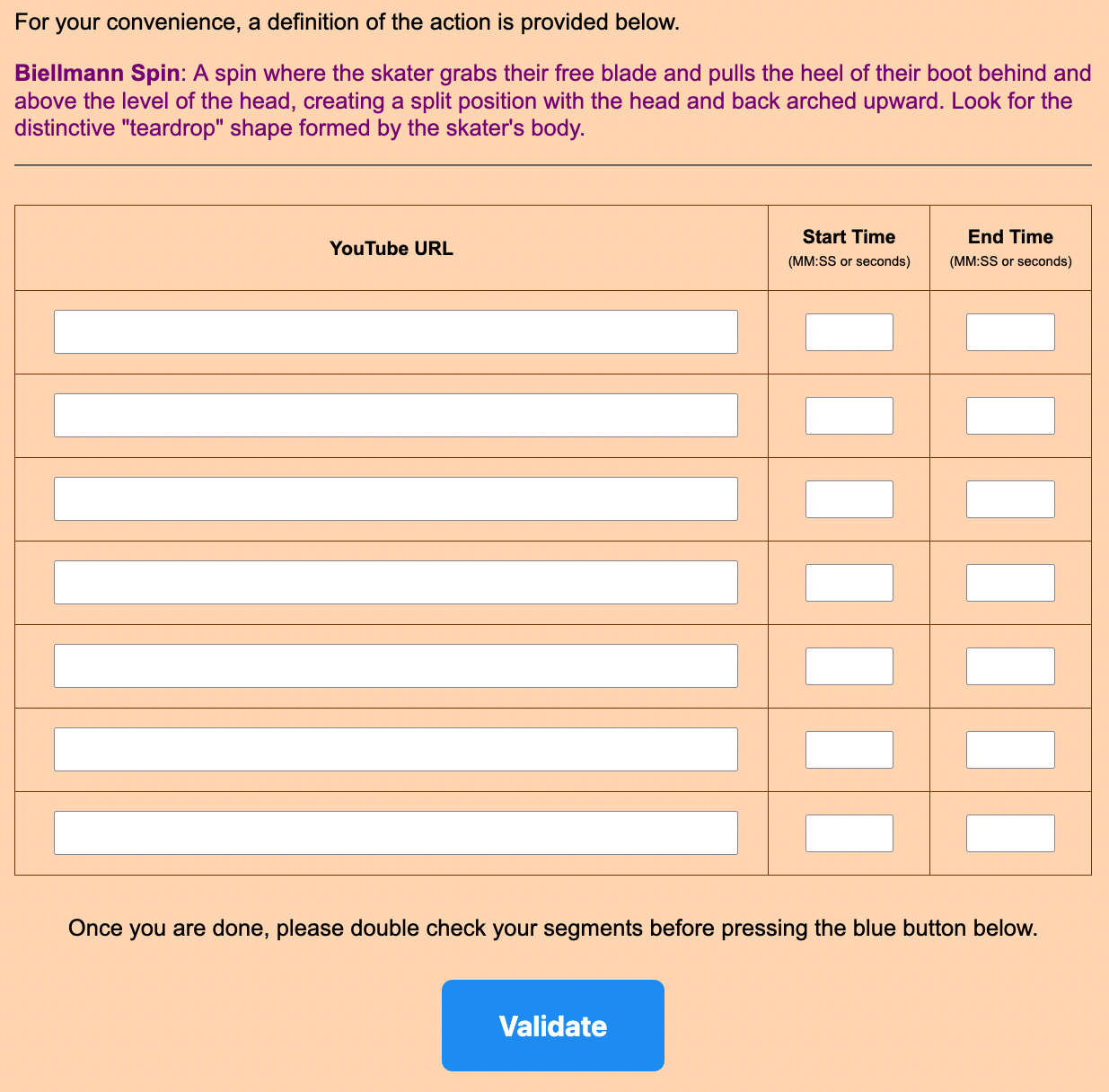}
    \caption{\textbf{Benchmark Clip Collection UI}. All of our UIs were refined based on annotator feedback. The annotators found this interface to be easy-to-use and appreciated the video tutorial.}
    \label{fig:collection_ui}
\end{figure*}

\begin{figure*}[!htbp]
    \centering
    \includegraphics[width=0.6\linewidth]{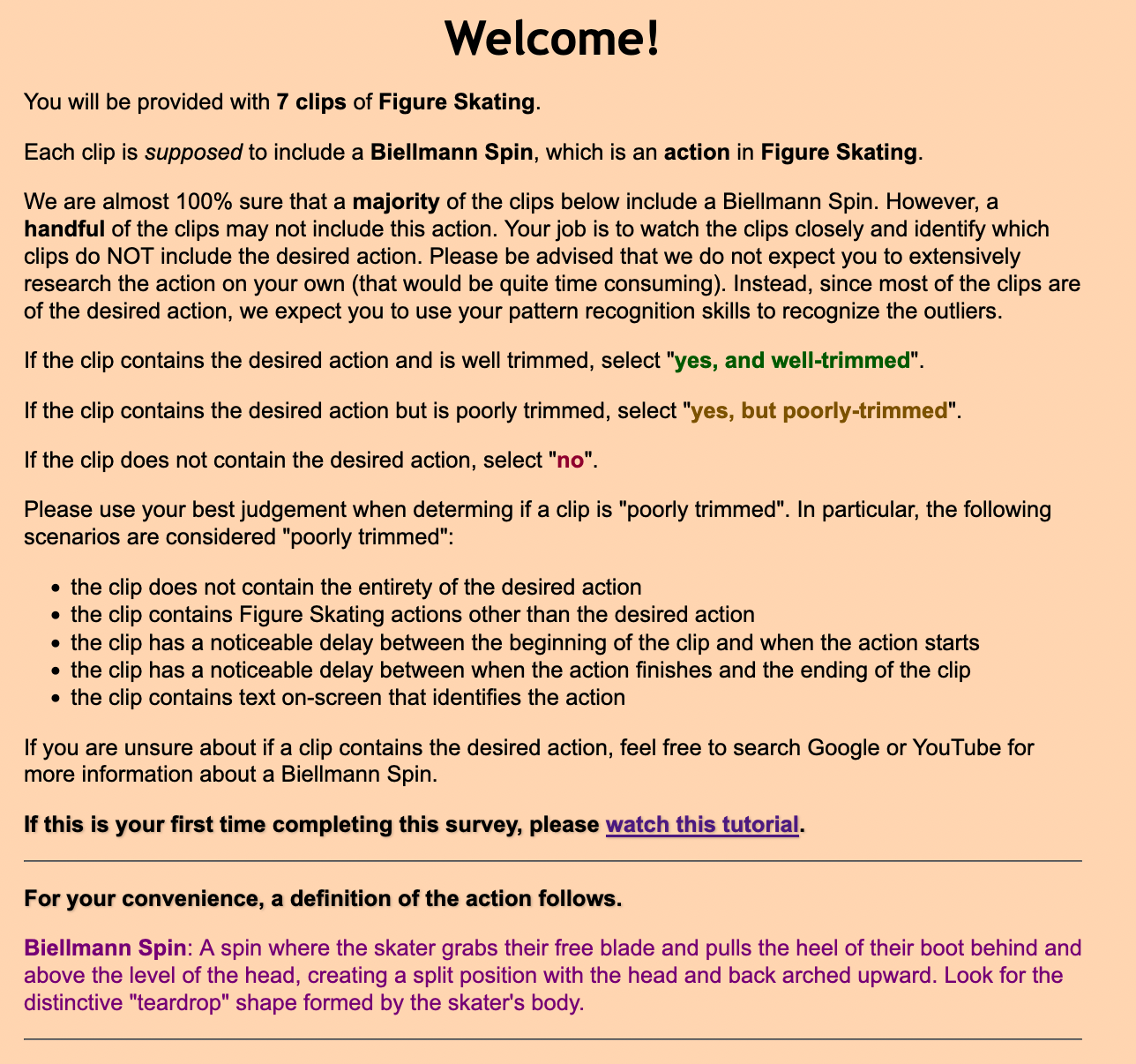}
    \includegraphics[width=0.6\linewidth]{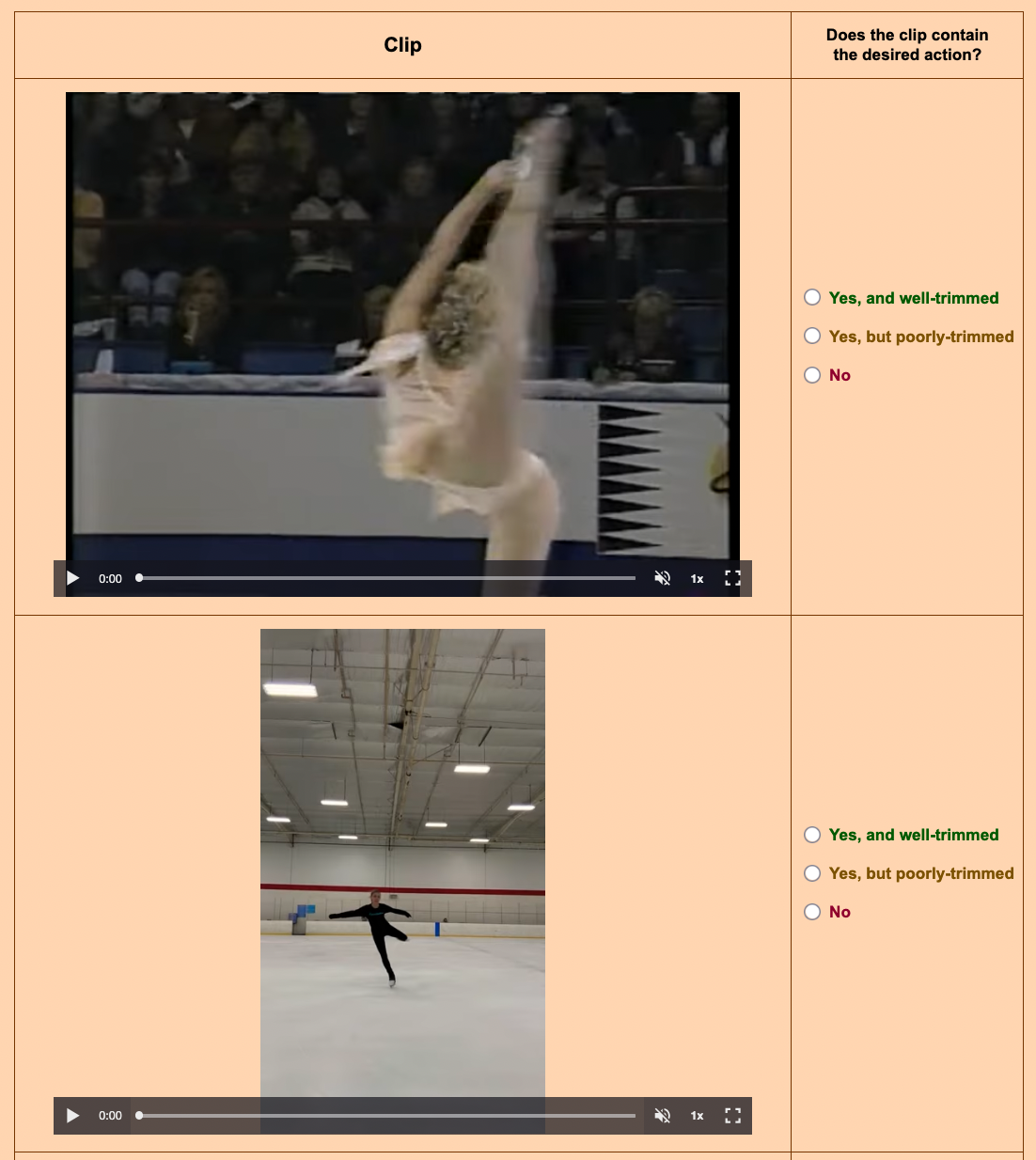}
    \caption{\textbf{Benchmark Clip Verification UI.} For brevity, only two of seven clips are displayed in the screenshot above. Likewise, a green submit button follows these clips, but is omitted above.}
    \label{fig:verification_ui}
\end{figure*}

\begin{figure*}[!htbp]
    \centering
    \includegraphics[width=0.65\linewidth]{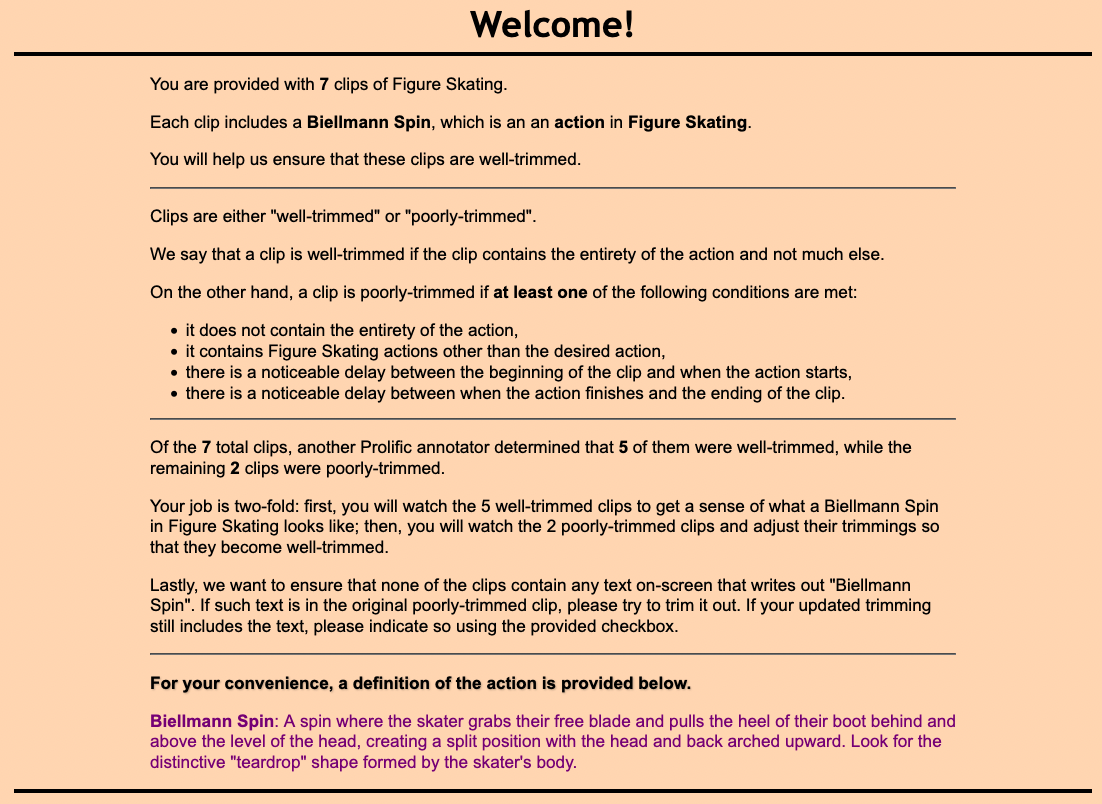}
    \includegraphics[width=0.65\linewidth]{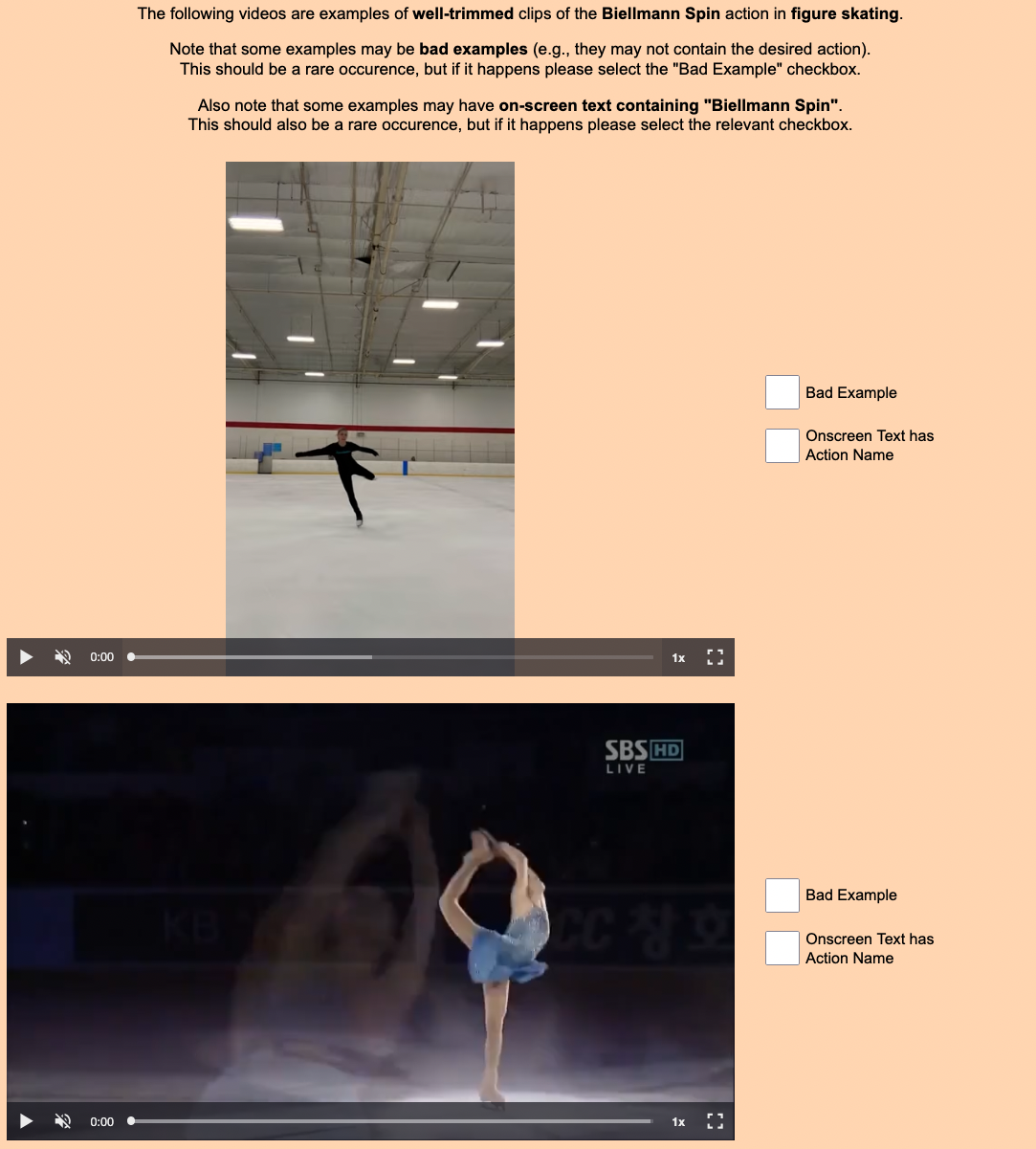}
\end{figure*}
\begin{figure*}[!htbp]
    \centering
    \includegraphics[width=0.8\linewidth]{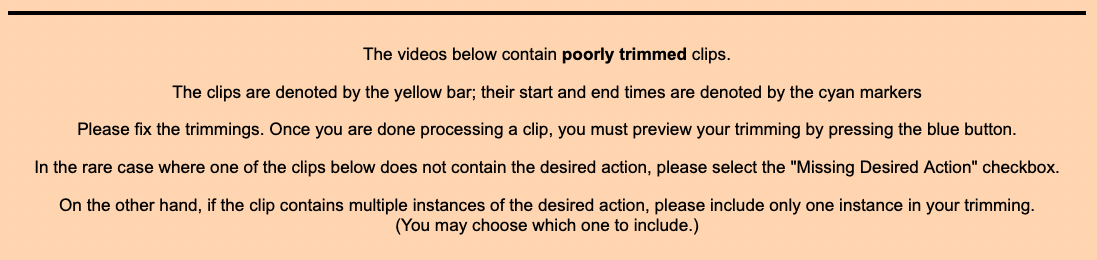}
    \includegraphics[width=0.8\linewidth]{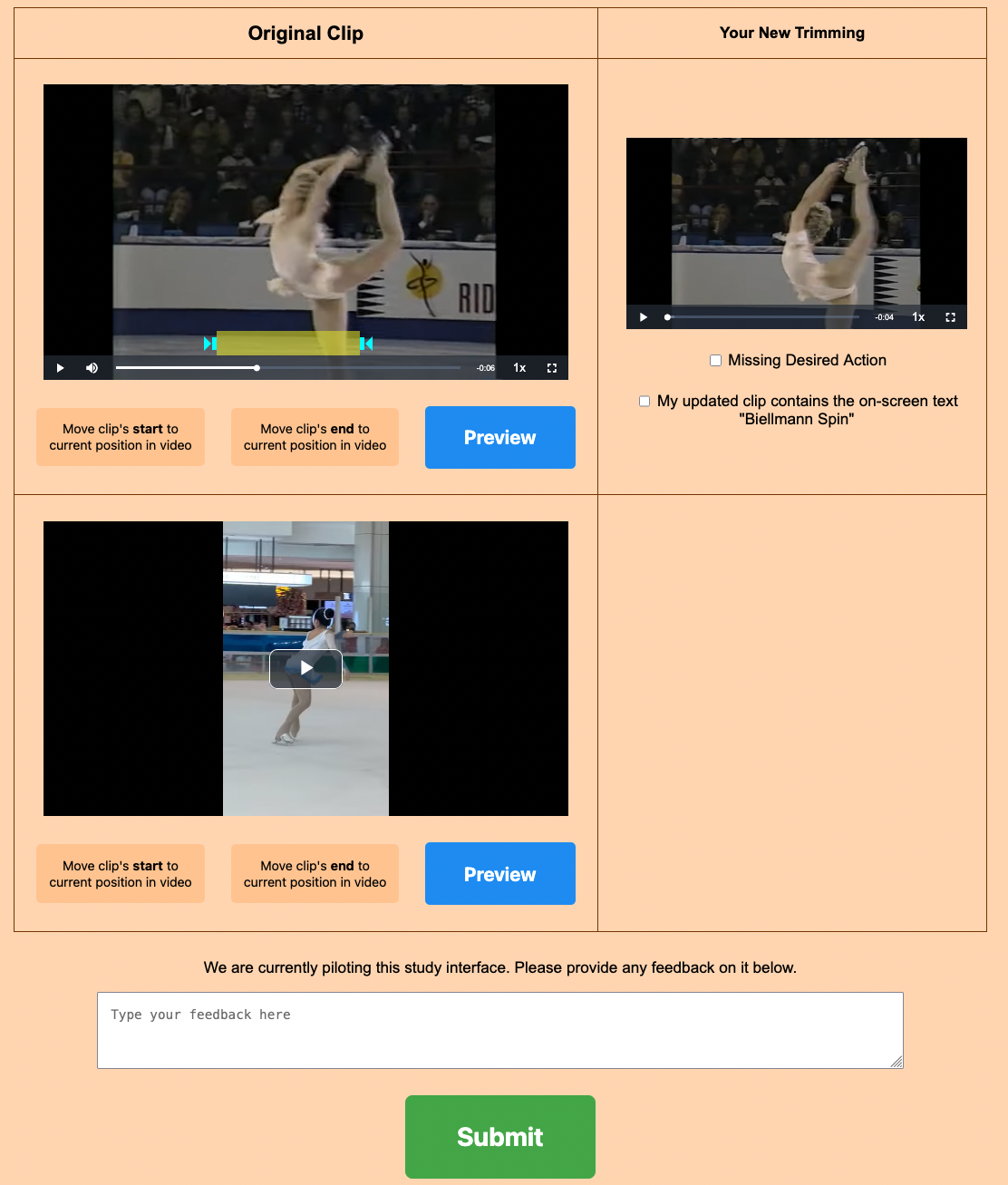}
    \caption{\textbf{Benchmark Clip Trimming UI.} The number of well-trimmed examples varies; for the action above, the true number is 5, but only 2 are shown for brevity. Similarly, the number of poorly-trimmed clips also varies.}
    \label{fig:trimming_ui}
\end{figure*}
\clearpage
\clearpage
\section{Model Evaluation}
\label{appendix:eval}

Evaluation code is \href{https://github.com/RAIVNLab/VideoNet}{available on GitHub}.

\subsection{Evaluation Prompts}

While we often tailor prompts to fit the expected input for each model, they closely resemble the following prompts. Minor adjustments are made to make the sentences read more smoothly.

\vspace{5pt}
\noindent ---
\vspace{5pt}

\noindent\textbf{Multiple-choice Prompt}

\noindent\texttt{Which of the following <DOMAIN> actions is shown in the video? \\\\
A. <FIRST OPTION> \\\\
B. <SECOND OPTION> \\\\
C. <THIRD OPTION> \\\\
D. <FOURTH OPTION> \\\\
Please respond with only the letter of the correct answer. \\\\
<VIDEO>
}

\vspace{5pt}
\noindent ---
\vspace{5pt}

\noindent\textbf{0-shot Prompt}
\vspace{10pt}

\noindent\texttt{Recall that <a OR an> <ACTION> is <a OR an> <SUBDOMAIN> in <DOMAIN>. Does the following video show <a OR an> <ACTION>? Please reason through your answer. It is critical that you output `yes' or `no' 
on the final line of your answer.
\\\\
<VIDEO>
}

\vspace{5pt}
\noindent ---
\vspace{5pt}

\noindent\textbf{3-shot Prompt}
\vspace{10pt}

\texttt{The following 3 videos show <a OR an> <ACTION>, which is <a OR an> <SUBDOMAIN> in <DOMAIN>.
\\\\
<VIDEO EXAMPLES>
\\\\
Now consider the following video. Is it also <a OR an> <ACTION>? 
Please reason through your answer. It is critical that you output 
`yes' or `no' on the final line of your answer.
\\\\
<VIDEO>}

\vspace{5pt}
\noindent ---
\vspace{5pt}

The \texttt{<SUBDOMAIN>} field defaults to the string \texttt{"action"}, but we sometimes provide a more descriptive word in its place (e.g., some American Football actions are classified under the subdomain of \texttt{"run"}).

The \texttt{<a OR an>} field is either the string \texttt{"a"} or the string \texttt{"an"} depending on if the word it precedes begins with a vowel.

The 1-shot and 2-shot prompts are nearly identical to the 3-shot prompt above and can be found on our GitHub repository. They are omitted here for brevity.

\subsection{Video Sampling}

We generally use the video sampling techniques recommended by the authors of each model. In certain cases, we place an upper bound on frame sampling due to compute constraints.
\vspace{5pt}
\begin{itemize}[leftmargin=0.5cm]
    \item InternVL3.5 \cite{internvl3_5}: uniformly sample, max 48 frames.
    \item Qwen3-VL \cite{Qwen3-VL}: two frames per second (fps)
    \item Molmo2 \cite{clark2025molmo2}: four fps, max 64 frames.
    \item Gemini 3.1 Pro \& Gemini 3 Flash \cite{gemini1.5}: one fps.
    \item GPT-5 \cite{gpt-5}: one fps, max 56 frames.
    \item GPT-5.4 \cite{gpt-5.4}: one fps, max 256 frames.
\end{itemize}

\subsection{Context Lengths}
For the open models, these numbers reflect a shared maximum on the number of tokens in both the input and output. For closed models, we have separate maximums for input tokens and output tokens.
\begin{itemize}[leftmargin=0.5cm]
    \item InternVL3.5: 12,000 tokens total
    \item Qwen3-VL: 128,000 tokens total
    \item Gemini 3.1 Pro \& Gemini 3 Flash: 1,048,576 input tokens; 65,536 output tokens
    \item GPT-5: 400,000 input tokens; 128,000 output tokens
    \item GPT-5.4: 1,050,000 input tokens; 128,000 output tokens
\end{itemize}

\subsection{Proprietary Model Versions}

We used the following versions of proprietary models.
\begin{itemize}[leftmargin=0.5cm]
    \item \texttt{gemini-3.1-pro-preview}
    \item \texttt{gemini-3-flash-preview}
    \item \texttt{gpt-5-2025-08-07}
    \item \texttt{gpt-5.4-2026-03-05}
\end{itemize}
We use the recommended reasoning levels for proprietary models, i.e., \texttt{medium} for GPT models and \texttt{high} for Gemini models.
\clearpage
\section{Zero-shot Ablations}
\label{appendix:zero_shot_ablations}

    Tables \ref{tab:detailed_fps_0shot} contains category-level results for Qwen3-VL-8B-Instruct and GPT-5.4 in the binary 0-shot setting with 1 frame per second (fps) sampling and 2 fps sampling.
    
    Qwen sees a slight performance improvement upon increasing the sampled frames per second from 1 to 2, although the model's default frame sampling rate is 2fps, so this gain may be attributable to shifting the video inputs to be more in-distribution. GPT-5.4, which has a recommended sampling rate of 1fps, sees a similar performance improvement, providing stronger evidence that test-time scaling (in terms of additional visual tokens) helps marginally on the domain-specific action recognition task. For comparison, increasing GPT-5.4's reasoning mode from \texttt{medium} to \texttt{xhigh} -- i.e., test-time scaling via reasoning tokens instead of visual tokens -- yields a similar improvement (73.3\% via \texttt{xhigh} vs 73.6\% via 2fps). Quadrupling the number of frames via 4fps sampling provides diminishing returns (0.7 percentage points from 2fps to 4fps vs. 1.3 points from 1fps to fps), and still fails to reach the accuracy attained by providing 1 in-context example (75.1\%).

    \begin{table*}[!h]
    \begin{tcolorbox}[
    enhanced,
    width=\linewidth,
    colback=polaris-bg-elevated,
    colframe=polaris-border-subtle,
    arc=3mm,
    boxrule=0.4pt,
    left=8pt,
    right=8pt,
    top=8pt,
    bottom=8pt,
    ]
    \caption{\textbf{Impact of higher FPS sampling in 0-shot.} Performance gain from the previous setup is highlighted in \textcolor{blue}{blue} and loss is highlighted in \textcolor{red}{red}. The Qwen-3-VL series defaults to 2fps sampling, whereas GPT-5.4 recommends 1fps. While 2fps sampling does provide gains in both models, these gains are limited; in fact, they are smaller in magnitude than the gains yielded by providing an in-context example.}
    \centering
    \resizebox{\textwidth}{!}{
    \begin{tabular}{lccccccccl}
        \toprule[1.5pt] 
        \textbf{Model} & \textbf{FPS} & \textbf{Beauty} & \textbf{Crafts} & \textbf{Dance} & \textbf{Food} & \textbf{Hobbies} & \textbf{Medical} & \textbf{Sports} & \textbf{\;\;\;Overall} \\ 
         
        \midrule
        \multirow{2}{*}{\centering Qwen3-VL-8B} 
        & 1 & 66.38 & 56.70 & 52.55 & 78.49 & 55.89 & 61.70 & 54.71 & 57.87\\
        & 2 & 62.07 & 59.96 & 54.74 & 79.30 & 55.21 & 58.51 & 58.08 & 59.25  \textcolor{blue}{\,$(+1.4)$} \\
        
        \midrule
        \multirow{2}{*}{\centering GPT-5.4} 
        & 1 & 75.86 & 77.17 & 66.42 & 91.67 & 66.12 & 75.00 & 71.46 & 72.35 \\
        & 2 & 81.90 & 77.35 & 69.89 & 92.47 & 68.05 & 73.94 & 71.80 & 73.65 \textcolor{blue}{\,$(+1.3)$} \\
        & 4 & 80.00 & 77.15 & 69.23 & 92.72 & 68.99 & 74.03 & 73.99 & 74.38\textcolor{blue}{\,$(+0.7)$}\\
        \bottomrule[1.2pt]
    \end{tabular}
    }
    \label{tab:detailed_fps_0shot}
\end{tcolorbox}
\end{table*}

        
         

    Table \ref{tab:detailed_zero_shot_input_ablations} contains category-level results for all models in the typical zero-shot setup of providing an input video, as well as two ablations: one where only the frame located at the (temporal) middle of the video is provided, and one where a definition of the action (as described in \S~\ref{subsubsec:definitions}) is given alongside the video. In general, performance is best when a definition is provided, and worst when only the middle frame is provided. The change in overall accuracy is visualized in \Cref{fig:zeroshot_input_ablations}.

    \begin{table*}[htbp!]
    \begin{tcolorbox}[
    enhanced,
    width=\linewidth,
    colback=polaris-bg-elevated,
    colframe=polaris-border-subtle,
    arc=3mm,
    boxrule=0.4pt,
    left=8pt,
    right=8pt,
    top=8pt,
    bottom=8pt,
    ]
    \caption{\textbf{Zero-shot results while varying video inputs.} Performance gain from the previous setup is highlighted in \textcolor{blue}{blue} and loss in \textcolor{red}{red}. Models are sorted by strongest to weakest overall performance in the default ``video'' input configuration. As expected, none of the models perform better when only given the center frame vs. the default video input. Notably, our fine-tuned model sees the biggest boost in accuracy from center frame input to the default, indicating its ability to leverage video inputs. The proprietary models benefit by less than half a percentage point when the action definition is added, suggesting that their language backbones possess sufficient world knowledge about the domain-specific actions in \dataset. Somewhat similarly, the open models experience no more than a 2 percentage point boost from action definitions.}
    \vspace{5pt}
    \centering
    \resizebox{\textwidth}{!}{
    \begin{tabular}{@{}llcccccccl@{}}
        \toprule[1.5pt] 
         \textbf{Model} & \textbf{Input} & \textbf{Beauty} & \textbf{Crafts} & \textbf{Dance} & \textbf{Food} & \textbf{Hobbies} & \textbf{Medical} & \textbf{Sports} & \textbf{\;\;\;Overall} \\ 
         
         \midrule

         \multirow{3}{*}{GPT-5} 
         & Middle Frame  & 73.28 & 67.39 & 60.58 & 85.75 & 63.51 & 71.81 & 64.06 & 66.55 \\
         & Video         & 76.72 & 77.72 & 66.24 & 90.59 & 69.21 & 75.00 & 70.62 & 72.88 \textcolor{blue}{\,($+6.3$)} \\
         & Video w/ Def. & 78.45 & 78.99 & 66.06 & 91.67 & 68.63 & 77.66 & 71.46 & 73.40 \textcolor{blue}{\,($+0.5$)} \\

         \midrule
         
         \multirow{3}{*}{GPT-5.4} 
         & Middle Frame  & 75.00 & 65.58 & 61.50 & 83.06 & 65.15 & 68.09 & 63.97 & 66.45 \\
         & Video         & 75.86 & 77.17 & 66.42 & 91.67 & 66.12 & 75.00 & 71.46 & 72.35 \textcolor{blue}{\,($+5.9$)} \\
         & Video w/ Def. & 81.03 & 76.81 & 68.25 & 90.86 & 68.15 & 74.47 & 70.62 & 72.88 \textcolor{blue}{\,($+0.5$)} \\

         \midrule

         \multirow{3}{*}{\centering Gemini 3.1 Pro}
         & Middle Frame  & 81.03 & 71.92 & 66.97 & 87.37 & 67.73 & 71.28 & 63.80 & 69.42 \\
         & Video         & 71.55 & 77.86 & 68.61 & 91.94 & 66.70 & 78.19 & 68.27 & 71.99 \textcolor{blue}{\,($+2.6$)} \\
         & Video w/ Def. & 68.97 & 77.90 & 65.69 & 88.44 & 64.09 & 78.72 & 66.25 & 69.95 \textcolor{red}{\,($-2.0$)} \\

         \midrule
         \multirow{3}{*}{Gemini 3 Flash}
         & Middle Frame  & 73.28 & 68.48 & 63.25 & 87.10 & 65.73 & 77.66 & 65.99 & 68.62 \\
         & Video         & 79.31 & 76.27 & 63.32 & 92.20 & 63.80 & 79.79 & 67.17 & 70.30 \textcolor{blue}{\,($+1.7$)} \\
         & Video w/ Def. & 76.72 & 75.91 & 63.69 & 91.62 & 65.51 & 80.85 & 67.40 & 70.72 \textcolor{blue}{\,($+0.4$)} \\

         \midrule
         \multirow{3}{*}{Molmo2-4B (FT)}
         & Middle Frame  & 56.90 & 59.78 & 54.74 & 68.89 & 58.01 & 62.77 & 57.41 & 58.93 \\
         & Video         & 73.28 & 69.02 & 66.06 & 73.66 & 67.76 & 69.15 & 61.45 & 66.60 \textcolor{blue}{\,\textbf{(}$\boldsymbol{+}$\textbf{7.7)}} \\
         & Video w/ Def. & 70.69 & 68.66 & 60.40 & 77.96 & 64.67 & 71.28 & 59.43 & 64.80 \textcolor{red}{\,($-1.8$)} \\
         
         \midrule
         \multirow{3}{*}{Qwen3-VL-8B}
         & Middle Frame  & 56.90 & 53.08 & 50.18 & 73.66 & 55.50 & 55.85 & 54.29 & 55.83 \\
         & Video         & 62.07 & 59.96 & 54.74 & 79.30 & 55.21 & 58.51 & 58.08 & 59.25 \textcolor{blue}{\,($+3.4$)} \\
         & Video w/ Def. & 65.52 & 61.96 & 55.84 & 83.06 & 59.65 & 65.43 & 56.06 & 61.00 \textcolor{blue}{\,($+1.8$)} \\
         
         \midrule
         \multirow{3}{*}{InternVL3.5-8B}
         & Middle Frame  & 57.76 & 51.45 & 52.55 & 67.20 & 55.12 & 56.91 & 53.79 & 55.15 \\
         & Video         & 58.62 & 54.17 & 52.01 & 75.27 & 55.12 & 57.98 & 57.41 & 57.35 \textcolor{blue}{\,($+2.2$)} \\
         & Video w/ Def. & 59.48 & 57.43 & 51.46 & 79.84 & 56.56 & 57.98 & 59.60 & 59.20 \textcolor{blue}{\,($+1.9$)} \\
        
        \bottomrule[1.2pt]
    \end{tabular}
    \label{tab:detailed_zero_shot_input_ablations}
    }
\end{tcolorbox}
\end{table*}
\clearpage
\section{Few-shot Results}
\label{appendix:few_shot}

This section includes category-level results for VLMs, results for traditional computer vision models in a modified evaluation setting, and a discussion of prompt sensitivity \& yes/no bias in Gemini 2.5 Pro.

\subsection{Category-level Results for VLMs}
\label{appendix:few_shot_category_level_vlm_results}

\Cref{tab:detailed_few_shot} contains category-level results for all models from \Cref{fig:few_shot_plot} in the 0-shot, 1-shot, 2-shot, and 3-shot binary setups.

\begin{table}[htbp!]
    \begin{tcolorbox}[
    enhanced,
    width=\linewidth,
    colback=polaris-bg-elevated,
    colframe=polaris-border-subtle,
    arc=3mm,
    boxrule=0.4pt,
    left=8pt,
    right=8pt,
    top=8pt,
    bottom=8pt,
    ]
    \caption{\textbf{Few-shot results.} Performance gain from the previous setup is highlighted in \textcolor{blue}{blue} and loss in \textcolor{red}{red}. Qwen-3-VL and Gemini 3.1 Pro experience the most noteworthy changes in performance.}
    \vspace{5pt}
    \centering
    \resizebox{\textwidth}{!}{
    \begin{tabular}{l c ccccccc l}
        \toprule[1.5pt] 
         \textbf{Model} & \textbf{$k$-shot} & \textbf{Beauty} & \textbf{Crafts} & \textbf{Dance} & \textbf{Food} & \textbf{Hobbies} & \textbf{Medical} & \textbf{Sports} & \textbf{\;\;\;Overall} \\ 
        \midrule
        \multirow{4}{*}{GPT-5} 
        & 0 & 76.72 & 77.72 & 66.24 & 90.59 & 69.21 & 75.00 & 70.62 & 72.88 \\
        & 1 & 77.59 & 79.17 & 71.17 & 91.13 & 69.79 & 71.28 & 71.55 & 74.08 \textcolor{blue}{\,($+1.2$)}\\
        & 2 & 76.72 & 79.18 & 70.80 & 90.86 & 71.04 & 70.21 & 71.38 & 74.18 \textcolor{blue}{\,($+0.1$)}\\
        & 3 & 76.72 & 78.99 & 74.09 & 91.67 & 72.10 & 72.87 & 72.31 & 75.38 \textcolor{blue}{\,($+1.2$)}\\

        \midrule

        \multirow{4}{*}{GPT-5.4}
        & 0 & 75.86 & 77.17 & 66.42 & 91.67 & 66.12 & 75.00 & 71.46 & 72.35 \\
        & 1 & 79.31 & 79.85 & 74.27 & 90.86 & 70.37 & 73.94 & 72.22 & 75.09 \textcolor{blue}{\,($+2.7$)}\\
        & 2 & 78.45 & 84.34 & 74.82 & 91.37 & 72.68 & 73.40 & 71.55 & 76.18 \textcolor{blue}{\,($+1.1$)}\\
        & 3 & 78.45 & 81.41 & 76.56 & 91.40 & 73.79 & 74.19 & 71.51 & 76.33 \textcolor{blue}{\,($+0.2$)}\\

        \midrule
        
        \multirow{4}{*}{Gemini 3.1 Pro}
        & 0 & 71.55 & 77.86 & 68.61 & 91.94 & 66.70 & 78.19 & 68.27 & 71.99 \\
        & 1 & 67.24 & 72.10 & 66.91 & 89.78 & 66.31 & 79.26 & 65.12 & 69.66 \textcolor{red}{\,($-2.3$)}\\
        & 2 & 69.57 & 72.00 & 65.07 & 88.92 & 63.79 & 69.15 & 63.62 & 67.81 \textcolor{red}{\,($-1.9$)}\\
        & 3 & 68.10 & 70.83 & 64.96 & 88.17 & 62.96 & 74.47 & 62.29 & 67.16 \textcolor{red}{\,($-0.7$)}\\

        \midrule

        \multirow{4}{*}{Gemini 3 Flash}
        & 0 & 79.31 & 76.27 & 63.32 & 92.20 & 63.80 & 79.79 & 67.17 & 70.30 \\
        & 1 & 77.59 & 81.16 & 72.99 & 92.74 & 69.79 & 81.38 & 68.27 & 74.25 \textcolor{blue}{\,($+4.0$)} \\
        & 2 & 81.03 & 79.53 & 73.18 & 93.55 & 71.81 & 79.26 & 68.86 & 74.82 \textcolor{blue}{\,($+0.6$)} \\
        & 3 & 82.76 & 79.71 & 73.72 & 94.89 & 72.10 & 78.72 & 68.60 & 75.08 \textcolor{blue}{\,($+0.3$)} \\

        \midrule
         
        \multirow{4}{*}{Qwen3-VL-8B} 
        & 0 & 62.07 & 59.96 & 54.74 & 79.30 & 55.21 & 58.51 & 58.08 & 59.25 \\
        & 1 & 68.10 & 67.45 & 61.13 & 85.22 & 62.26 & 61.17 & 56.65 & 63.41 \textcolor{blue}{\,\textbf{(}$\boldsymbol{+4.2}$\textbf{)}}\\
        & 2 & 69.83 & 64.61 & 60.22 & 86.29 & 62.51 & 65.93 & 61.70 & 64.83 \textcolor{blue}{\,($+1.4$)}\\
        & 3 & 74.56 & 65.41 & 62.04 & 87.98 & 60.58 & 69.46 & 65.40 & 66.22 \textcolor{blue}{\,($+1.4$)}\\

        \midrule

        \multirow{4}{*}{InternVL3.5-8B}
        & 0 & 58.62 & 54.17 & 52.01 & 75.27 & 55.12 & 57.98 & 57.41 & 57.35 \\
        & 1 & 62.07 & 55.25 & 55.66 & 76.88 & 56.95 & 57.98 & 57.41 & 58.73 \textcolor{blue}{\,($+1.4$)}\\
        & 2 & 62.07 & 57.97 & 55.66 & 80.38 & 59.07 & 65.96 & 57.74 & 60.45 \textcolor{blue}{\,($+1.7$)}\\
        & 3 & 68.97 & 56.88 & 53.28 & 78.23 & 59.65 & 66.49 & 58.75 & 60.45 \textcolor{blue}{\,($+0.0$)}\\

        \bottomrule[1.2pt]
    \end{tabular}
    \label{tab:detailed_few_shot}
    }
\end{tcolorbox}
\end{table}

\clearpage

\subsection{Results for Traditional Models}
\label{appendix:traditional_models}

We also evaluate traditional models (i.e., models that are not VLMs) on \dataset. In particular, we evaluate 4 recent CLIP models \cite{XCLIP,internvid,longclip,videoclipxl} and the 3 convolutional neural networks (CNNs) from \cite{quovadis_kinetics}. All of the CLIP models except \cite{longclip} were designed for video inputs; following \cite{videoclipxl}, we uniformly sample 8 frames from the video and average their features when evaluating \cite{longclip}.

These models do not natively support visual question answering with natural language. They also cannot be provided multiple in-context videos. Hence, we adapt our few-shot evaluation setup for these models. We have two separate adaptations: one for the CLIP models, one for the CNNs. 

\begin{table}[h!]
    \begin{tcolorbox}[
    enhanced,
    width=\linewidth,
    colback=polaris-bg-elevated,
    colframe=polaris-border-subtle,
    arc=3mm,
    boxrule=0.4pt,
    left=8pt,
    right=8pt,
    top=8pt,
    bottom=8pt,
]
    \centering
    \caption{\textbf{CLIP results}. Even the best CLIP model fails to match the worst VLM we evaluated. Given that random chance is 50\%, these results indicate that CLIP models, as trained, struggle on the domain-specific action recognition task. ``Acc.'' is short for accuracy.}
    \label{tab:clip_results}
    \resizebox{0.65\linewidth}{!}{
    \begin{tabular}{@{}cccc@{}}
        \toprule[1.5pt] 
         \textbf{Model} & \textbf{Test Acc. (\%)} & \textbf{Val Acc. (\%)} & \textbf{Threshold}\\
         \midrule
         ViCLIP & 51.45 & 52.21 &  0.2002 \\
         LongCLIP-L & \textbf{52.14} & \textbf{52.44} & 0.6938 \\
         VideoCLIP-XL-v2 & 50.67 & 52.21 & 0.2087 \\
         X-CLIP-L/14 & 50.18 & 51.75 & 0.1224 \\
        \bottomrule[1.5pt]
    \end{tabular}
    }
\end{tcolorbox}
\end{table}
\vspace{-10pt}
\begin{table}[h!]
    \begin{tcolorbox}[
    enhanced,
    width=\linewidth,
    colback=polaris-bg-elevated,
    colframe=polaris-border-subtle,
    arc=3mm,
    boxrule=0.4pt,
    left=8pt,
    right=8pt,
    top=8pt,
    bottom=8pt,
    ]
    \centering
    \caption{\textbf{CLIP results when ``cheating''}. The optimal threshold is now computed over the test set rather than the validation set. The poor results persist. This suggests that the CLIP architecture or training regime, rather than the size of our validation set, is at fault for the CLIP models' lackluster performance on \dataset. }
    \label{tab:clip_cheating_results}
    \resizebox{0.5\linewidth}{!}{
    \begin{tabular}{@{}ccc@{}}
        \toprule[1.5pt] 
         \textbf{Model} & \textbf{Test Acc.} (\%) & \textbf{Threshold}\\
         \midrule
         ViCLIP & 53.47 & 0.2121 \\
         LongCLIP-L & 53.29 & 0.7400 \\
         VideoCLIP-XL-v2 & \textbf{52.83} & 0.1969 \\
         X-CLIP-L/14 & 51.10 & 0.1272 \\
        \bottomrule[1.5pt]
    \end{tabular}
    }
\end{tcolorbox}
\end{table}

We begin by computing CLIP scores for all clips in \dataset~with their corresponding all-lowercase text labels formatted as "\texttt{<<DOMAIN>> <<ACTION>>}" (e.g., ``figure skating biellmann spin''). We then search for the \textit{optimal threshold} on a balanced\footnote{Here, ``balanced'' denotes that, if the validation set is thought of as containing binary questions, then precisely half the validation set contains binary positive questions (i.e., binary questions where the answer is ``yes'').} validation set constructed from clips in \dataset~which do NOT appear in the test set. To do so, we compute the validation accuracy for all candidate thresholds where the validation accuracy can change.\footnote{Our validation set contains 2,174 questions. Hence, there are at most 2,175 critical points at which the validation accuracy can change.} Concretely, if the CLIP score exceeds or equals the threshold, the CLIP model's answer to the test set question is considered ``yes''; otherwise, the answer is considered ``no''. At last, after finding the optimal threshold on the validation set, we present the model with the test set, which contains the same pairs of clips and actions that VLMs see in the normal \dataset~evaluation setup. The results for this setup are in \Cref{tab:clip_results}. The CLIP models struggle immensely, falling short of every VLM we tested. To alleviate concerns that the validation set may have been too small to find a decent threshold, we also search for the optimal threshold directly on the test set in \Cref{tab:clip_cheating_results}. Still, the CLIP models struggle, suggesting that they are ill suited for this task.

Next, we consider CNNs, which extract video features but do not provide a way to align these video features to text. Accordingly, we opt for a k-nearest neighbors classifier (kNN) approach in evaluating the CNNs. In particular, we extract video features from the 3 in-context examples provided in \dataset~for each action and use these features as the support set for a kNN. The kNN then classifies the test samples based on Euclidean distance. We try all $k\in\{1,2,3\}$. It is worth noting that no two \dataset~clips for any given action are taken from the same source video, minimizing concerns about a kNN ``hacking'' correct answers via factors like the video background. The kNN is deemed to answer “does the following video show X” with a “yes” if it classifies the test sample as action X, and “no” if it classifies the test sample as another action. For comparison, we also evaluate the two best CLIP models using this approach by feeding their video features to a kNN. As shown in \Cref{tab:cnn_results}, the best CNN, Two-Stream I3D, rivals the best CLIP models but still falls short of all VLMs. Similar to CLIP, the I3D models, as trained, seem poorly suited for the domain-specific action recognition task.

\begin{table}[h!]
    \begin{tcolorbox}[
    enhanced,
    width=\linewidth,
    colback=polaris-bg-elevated,
    colframe=polaris-border-subtle,
    arc=3mm,
    boxrule=0.4pt,
    left=4pt,
    right=4pt,
    top=6pt,
    bottom=6pt,
]
    \centering
    \caption{\textbf{CNN results}. Rows sorted by top-5 accuracy on Kinetics \cite{kinetics}. With some exceptions, higher \dataset~accuracy tends to be correlated with better performance on Kinetics.}
    \label{tab:cnn_results}
    \resizebox{0.65\linewidth}{!}{
    \begin{tabular}{@{}c|ccc|c@{}}
        \toprule[1.5pt] 
         \multirow{2}{*}{\textbf{Model}} & \multicolumn{3}{c}
         {\textbf{\dataset~Accuracy (\%)}} & \multirow{2}{3cm}{\centering \textbf{Kinetics Top-5 Accuracy (\%)}} \\
         & $\boldsymbol{k=1}$ & $\boldsymbol{k=2}$ & $\boldsymbol{k=3}$ \\
         \midrule
         ViCLIP & 54.58 & 53.73 & 53.36 & 98.2 \\
         Two-Stream I3D & 54.55 & 53.26 & 53.03 & 91.3 \\
         RGB-I3D & 53.52 & 52.56 & 52.09 & 89.3 \\
         Flow-I3D & 53.03 & 52.46 & 52.45 & 84.9 \\
         LongCLIP-L & 54.12 & 52.87 & 52.55 & - \\
        \bottomrule[1.5pt]
    \end{tabular}
    }
\end{tcolorbox}
\end{table}

\subsection{Prompt Sensitivity \& Yes/No Bias}

\begin{figure*}[h!]
    \centering
    \includegraphics[width=\linewidth]{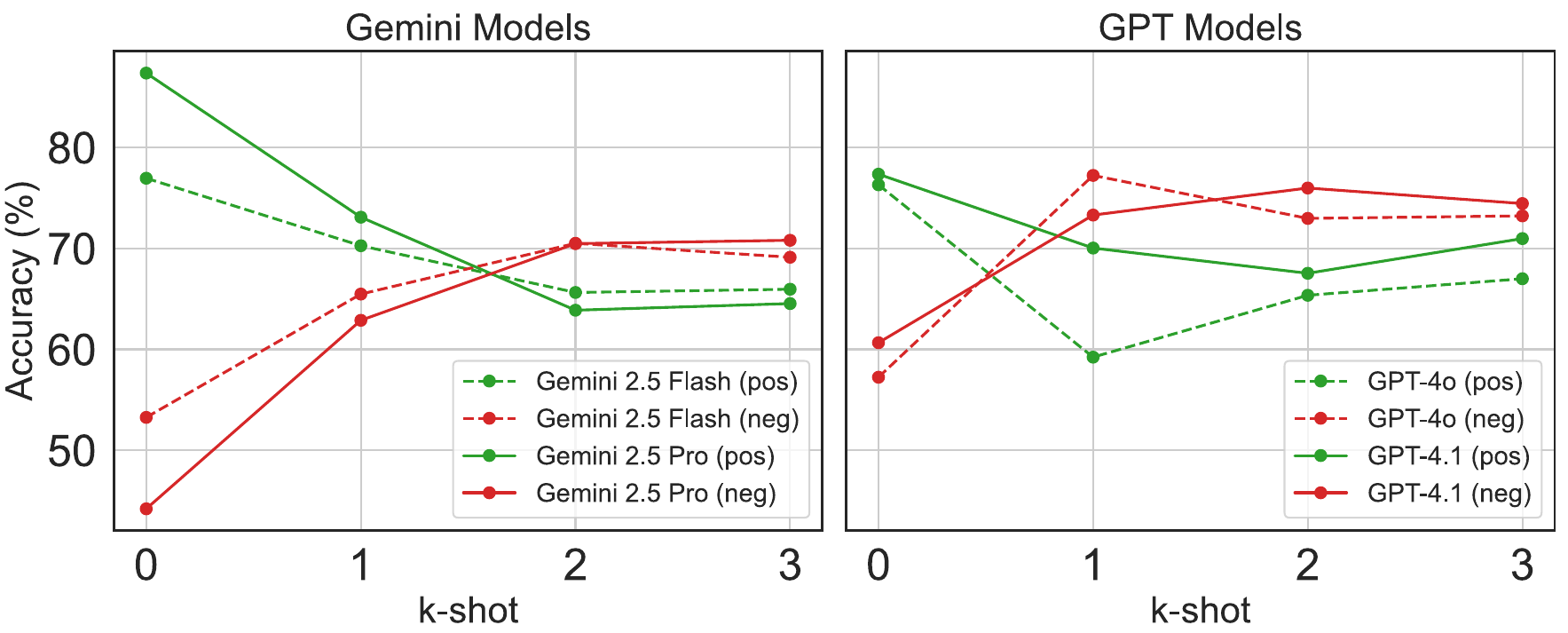}
    \caption{\textbf{Positive \& negative accuracy with in-context examples.} Accuracy on positive clips is in green; accuracy on negative clips is in red. In both plots, the weaker model is shown with dashed lines, while the stronger reasoning model is shown with solid lines. Note that the GPT models (right), which attain a higher accuracy on \dataset~than the Gemini models (left), see smaller changes in their yes/no bias as additional few-shot examples are provided.}
    \label{fig:fewshot_yesno_bias}
\end{figure*}

We observe that model performance on positive clips and negative clips changes significantly when in-context examples are provided (see Table \ref{tab:aar_posneg_trend}). Given the poor performance of open models on our benchmark, we focus on analyzing the behavior of Gemini and GPT models (see Figure \ref{fig:fewshot_yesno_bias}). 

Gemini 2.5 Pro exhibits a stark pattern, performing better on negative clips and worse on positive clips as additional in-context examples are provided. GPT-4.1 exhibits a similar pattern, but to a much lesser (and thus, ``more acceptable'') extent. We believe there are two main hypotheses to explain this phenomenon. One is that Gemini 2.5 Pro over-emphasizes insignificant details from the the in-context examples (e.g., background composition, camera angle, etc.) as opposed to the fine-grained details of the action at-hand. The other is this behavior can be attributed to our prompt. 

We test the latter hypothesis by constructing two prompts (see \Cref{fig:biased_prompts}): a ``lenient'' prompt which should bias models towards saying ``yes'', and a ``balanced'' prompt which attempts to eliminate any unintended bias introduced by few-shot examples. (As discussed previously, our ``default'' prompt seems to bias the model twards saying ``no''.) We tailor these prompts based on how they impact performance in the weaker models (Qwen, Intern, Gemini), before evaluating their impact on two proprietary models (GPT-4o and GPT-4.1). \Cref{tab:fewshot_prompt_sensitivity} confirms that even large proprietary \textbf{models are NOT robust to slight changes in the prompt}. Surprisingly, the \textit{overall accuracy is relatively unaffected} by these changes.

Given that small differences in the prompt cause dramatic shifts in yes/no accuracies, we hypothesize that such ``prompt sensitivity'' is an indicator that these models are not confident in their answers. This is reminiscent of early generations of LLMs, which were often not confident in their answers and hence would easily change their answers based on the smallest of pushback from the user \cite{calibratebeforeuse}.


\begin{table}[htbp!]
\centering
\begin{tcolorbox}[
enhanced,
width=\linewidth,
colback=polaris-bg-elevated,
colframe=polaris-border-subtle,
arc=3mm,
boxrule=0.4pt,
left=8pt,
right=8pt,
top=8pt,
bottom=8pt,
]

\caption{\textbf{Prompt sensitivity in GPT models.} The first table shows numbers for GPT-4o; the second shows numbers for GPT-4.1. Reading the first two numeric columns top-to-bottom, we see that accuracies on positive and negative clips change drastically with the choice of prompt. However, reading down the last column, the accuracy across all clips exhibits minimal change. The trend remains equally present in both tables despite GPT-4.1's stronger performance (69.02\% vs 66.76\%) on \dataset.}
\label{tab:fewshot_prompt_sensitivity}

\centering
\noindent
\makebox[\linewidth]{%
\begin{minipage}[t]{0.48\linewidth}
\centering
\resizebox{\linewidth}{!}{
\begin{tabular}{@{} c | ccc @{}}
\toprule
\multirow{2}{*}{\textbf{Prompt}} & \multicolumn{3}{c}{\textbf{Binary 0-shot Accuracy for GPT-4o (\%)}} \\
& \textbf{Positive Clips} & \textbf{Negative Clips} & \textbf{All Clips} \\
\midrule
\textbf{Default} & \multirow{2}{*}{67.00} & \multirow{2}{*}{73.23} & \multirow{2}{*}{70.12} \\
\textit{\small biases ``no''} \\
\midrule
\textbf{Balanced} & \multirow{2}{*}{77.55} & \multirow{2}{*}{62.46} & \multirow{2}{*}{70.11} \\
\textit{\small minimal bias} \\
\midrule
\textbf{Lenient} & \multirow{2}{*}{87.20} & \multirow{2}{*}{55.21} & \multirow{2}{*}{71.22} \\
\textit{\small biases ``yes''} \\
\bottomrule
\end{tabular}
}
\end{minipage}\hfill
\begin{minipage}[t]{0.48\linewidth}
\centering
\resizebox{\linewidth}{!}{
\begin{tabular}{@{} c | ccc @{}}
\toprule
\multirow{2}{*}{\textbf{Prompt}} & \multicolumn{3}{c}{\textbf{Binary 0-shot Accuracy for GPT-4.1 (\%)}} \\
& \textbf{Positive Clips} & \textbf{Negative Clips} & \textbf{All Clips} \\
\midrule
\textbf{Default} & \multirow{2}{*}{70.98} & \multirow{2}{*}{74.45} & \multirow{2}{*}{72.71} \\
\textit{\small biases ``no''} \\
\midrule
\textbf{Balanced} & \multirow{2}{*}{84.73} & \multirow{2}{*}{59.44} & \multirow{2}{*}{72.12} \\
\textit{\small minimal bias} \\
\midrule
\textbf{Lenient} & \multirow{2}{*}{90.13} & \multirow{2}{*}{53.94} & \multirow{2}{*}{72.06} \\
\textit{\small biases ``yes''} \\
\bottomrule
\end{tabular}
}
\end{minipage}%
}

\end{tcolorbox}
\end{table}
\begin{figure*}
    \begin{tcolorbox}[
    enhanced,
    width=1.0\linewidth,
    colback=polaris-bg-elevated,
    colframe=polaris-border-subtle,
    arc=3mm,
    boxrule=0.4pt,
    left=4pt,
    right=4pt,
    top=6pt,
    bottom=6pt,
]
    \centering
    \underline{\textbf{Default 3-shot Prompt}}
    \vspace{5pt}
    \begin{verbatim}
The following 3 videos show examples of <a OR an> <ACTION>, which is <a OR an> 
<SUBDOMAIN> in <DOMAIN>.

<VIDEO EXAMPLES>

Now consider the following video. Is it also <a OR an> <ACTION>?

Please reason through your answer. It is critical that you output 
`yes' or `no' on the final line of your answer.
    \end{verbatim}
    \vspace{10pt}
    
    \underline{\textbf{Balanced 3-shot Prompt}}
    \vspace{5pt}
\begin{verbatim}
The following 3 videos show examples of <a OR an> <ACTION>, which is <a OR an> 
<SUBDOMAIN> in <DOMAIN>.

<VIDEO EXAMPLES>

Now consider the following video. Is it also <a OR an> <ACTION>?

An appropriate instance of <ACTION> must include all essential defining
elements, but minor variations or slight differences in style or execution
are acceptable. Analyze carefully, explicitly noting the presence or absence
of essential elements, while considering natural variations. Clearly explain
your reasoning and justify your final decision. It is critical that you output 
`yes' or `no' on the final line of your answer.\end{verbatim}
    \vspace{10pt}
    
    \underline{\textbf{Lenient 3-shot Prompt}}
    \vspace{5pt}
    \begin{verbatim}
The following 3 videos show examples of <a OR an> <ACTION>, which is <a OR an> 
<SUBDOMAIN> in <DOMAIN>.

<VIDEO EXAMPLES>

Now consider the following video. Is it also <a OR an> <ACTION>?

Focus on identifying the core defining elements rather than expecting an
exact match to the examples. The action may have natural variations in
execution while still being the same action. Please reason through your
answer. It is critical that you output `yes' or `no' on the final line 
of your answer.\end{verbatim}
    \vspace{10pt}
    
    \caption{\textbf{Default, Balanced, and Lenient Prompts.} Observe that there are only small differences between each prompt.}
    \label{fig:biased_prompts}
\end{tcolorbox}
\end{figure*}
\begin{table*}[htbp!]
    \begin{tcolorbox}[
    enhanced,
    width=\linewidth,
    colback=polaris-bg-elevated,
    colframe=polaris-border-subtle,
    arc=3mm,
    boxrule=0.4pt,
    left=8pt,
    right=8pt,
    top=8pt,
    bottom=8pt,
]   
    \caption{\textbf{Performance on positive vs. negative clips with in-context examples.} Positive clips are clips where the ground truth answer is ``yes'' (i.e., the clip contains the action that the question is asking about); negative clips are ones where the ground truth is ``no''. Gemini 3.1 Pro--the only model which continually declines in accuracy as $k$ is increased--displays significant bias towards answering ``no'' when in-context examples are provided. Meanwhile, Gemini 3 Flash, which improves by 4 to 5 percentage points with in-context examples, shifts from displaying significant positive bias when $k=0$ to being relatively balanced when $k>0$. NB: although the benchmark contains the same number of positive and negative clips, the entries in the last column may not be exact averages of the entries in the prior two columns due to one-off model errors on benchmark questions.}
    \centering
    \resizebox{0.7\linewidth}{!}{
    \begin{tabular}{lcccc}
        \toprule[1.5pt]
        \textbf{Model Name} & \textbf{$k$-shot} & \textbf{Positive Clips} & \textbf{Negative Clips} & \textbf{Overall} \\
        \midrule
        
        \multirow{4}{*}{GPT-5}
        & 0 & 76.00 & 69.75 & 72.88 \\
        & 1 & 65.75 & 82.40 & 74.08 \\
        & 2 & 63.34 & 85.01 & 74.13 \\
        & 3 & 66.20 & 84.55 & 75.38 \\
        
        \midrule
        \multirow{4}{*}{GPT-5.4}
        & 0 & 74.55 & 70.15 & 72.35 \\
        & 1 & 72.54 & 77.65 & 75.09 \\
        & 2 & 71.74 & 80.60 & 76.18 \\
        & 3 & 73.31 & 79.34 & 76.33 \\

        \midrule
        \multirow{4}{*}{Gemini 3.1 Pro}
        & 0 & 65.08 & 78.90 & 71.99 \\
        & 1 & 46.52 & 92.80 & 69.66 \\
        & 2 & 41.88 & 93.68 & 67.81 \\
        & 3 & 41.02 & 93.30 & 67.16 \\
        
        \midrule
        \multirow{4}{*}{Gemini 3 Flash}
        & 0 & 91.40 & 49.20 & 70.30 \\
        & 1 & 73.80 & 74.70 & 74.25 \\
        & 2 & 73.75 & 75.90 & 74.83 \\
        & 3 & 72.65 & 77.50 & 75.08 \\

        \midrule
        \multirow{4}{*}{Qwen3-VL-8B} 
        & 0 & 46.90 & 71.60 & 59.25 \\
        & 1 & 43.84 & 82.95 & 63.41 \\
        & 2 & 63.96 & 65.70 & 64.83 \\
        & 3 & 72.33 & 60.12 & 66.22 \\
        
        \midrule
        \multirow{4}{*}{InternVL3.5-8B}
        & 0 & 73.80 & 40.90 & 57.35 \\
        & 1 & 69.20 & 48.25 & 58.73 \\
        & 2 & 79.45 & 41.45 & 60.45 \\
        & 3 & 82.05 & 38.85 & 60.45 \\
        
        \bottomrule[1.2pt]
    \end{tabular}
    }
    \label{tab:aar_posneg_trend}
\end{tcolorbox}
\end{table*}
\clearpage
\clearpage
\section{Human Evaluation}
\label{appendix:human_eval}

We have four versions of the human evaluation UI, depending on if the human is shown few-shot examples and whether they are shown the action definition. Figure \ref{fig:human_eval_ui} displays one of these setups. 

Both humans and models are shown silenced videos. 

In Table~\ref{tab:human_eval}, we report human performance with different binary configurations, namely 0-shot vs. 3-shot and with vs. without definition. We also report performance with random negatives.

\begin{table*}[h!]
    \begin{tcolorbox}[
    enhanced,
    width=1.0\linewidth,
    colback=polaris-bg-elevated,
    colframe=polaris-border-subtle,
    arc=3mm,
    boxrule=0.4pt,
    left=8pt,
    right=8pt,
    top=8pt,
    bottom=8pt,
    ]
    \caption{\textbf{Human performance on~\dataset}. We report accuracy on positive clips, negative clips, and all clips, across different sets of negative (hard vs. random). For reference, the best $3$-shot model, GPT-5.4, achieves an overall accuracy of 76.33\% and 81.03\% on hard and random negatives, respectively.}
    \label{tab:human_eval}
    \centering
    \resizebox{0.7\linewidth}{!}{
    \begin{tabular}{l | c c c}
        \toprule
        \textbf{Human Evaluation} & \textbf{Positive Clips} & \textbf{Negative Clips} & \textbf{Overall} \\
        \midrule
        \textit{Hard Negatives} \\
        \quad 0-shot without definition & 85.96 & 43.27 & 64.61 \\
        \quad 0-shot with definition & 86.53 & 51.58 & 69.05 \\
        \quad 3-shot without definition & 91.98 & 65.61 & 78.80 \\
        \quad 3-shot with definition & \textbf{93.41} & \textbf{71.92} & \textbf{82.66} \\
        \midrule
        \textit{Random Negatives} \\
        \quad 0-shot with definition & \textbf{93.41} & 69.63 & 81.52 \\
        \quad 3-shot with definition & 91.69 & \textbf{95.42} & \textbf{93.55} \\
        \bottomrule
    \end{tabular}
    }
\end{tcolorbox}
\end{table*}

Across the board, we see humans excel at identifying positive clips, achieving high accuracy (above 85\%) even without definitions or examples. They even attain accuracies above 91\% when provided with examples (in the 3-shot setting). However, humans struggle with identifying negative clips, especially in the hard negative setup. Despite being given 3 example videos and a definition, humans get only 71.92\%, while the 0-shot with-definition configuration attains a mere 51.58\%. 

Promisingly, we see a steady improvement in negative clip accuracy as more in-context examples and the action definition are provided. In fact, 3-shot humans armed with action definitions achieve notably high accuracy on random negatives (95.42\%), nearly solving the task. 

Overall, these findings suggest that while providing definitions and in-context examples significantly helps humans distinguish general in-domain actions, additional domain expertise or perceptual skills might be needed to reliably differentiate highly similar actions.

As mentioned in \Cref{subsec:few_shot_eval}, we sample 698 questions. Each question is answered by three annotators.\footnote{We use pools of approximately 200 annotators per human evaluation setup.} Based on additional experiments (not reported here), we find that this process effectively estimates the accuracy that non-expert humans would attain on the entire benchmark.

\begin{figure*}[htbp!]
    \centering
    \includegraphics[width=0.75\linewidth]{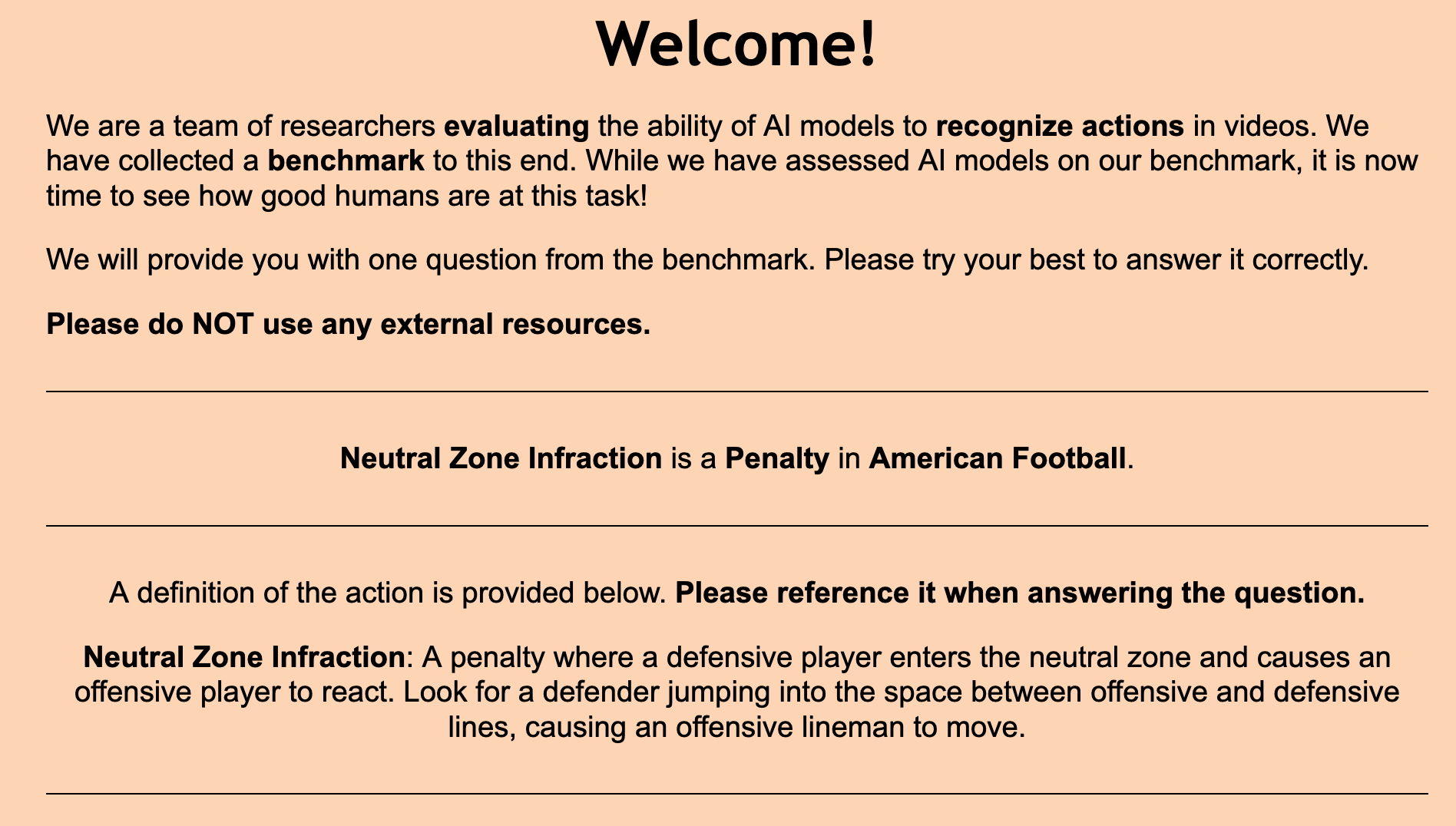}
    \includegraphics[width=0.75\linewidth]{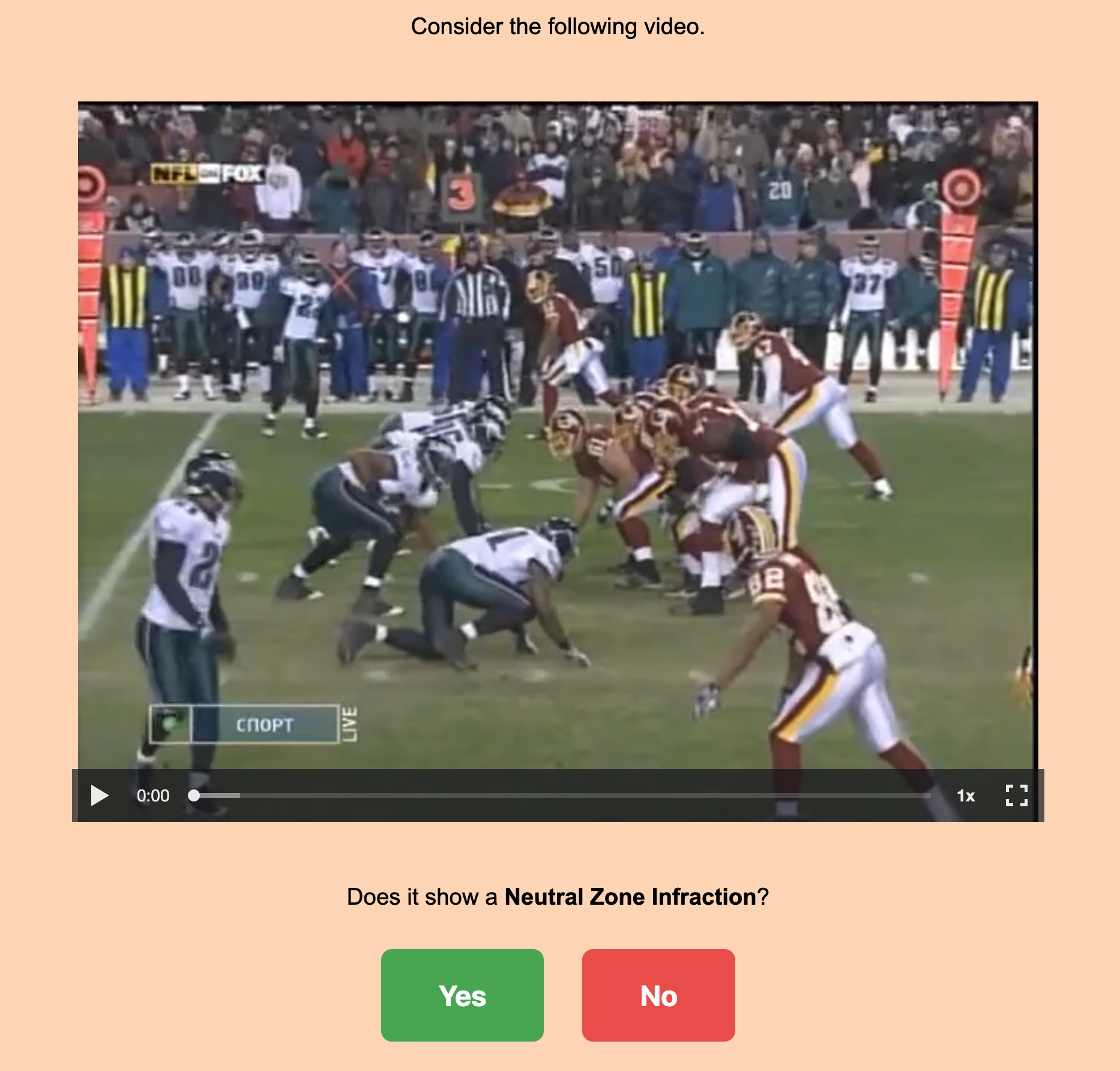}
    \caption{\textbf{Human evaluation UI.} In this configuration, the human is provided with a definition but is given no in-context examples.}
    \label{fig:human_eval_ui}
\end{figure*}
\clearpage
\section{Additional Training Details}
\label{appendix:additional_training_details}

This appendix elaborates on \Cref{sec:model_training}.

\subsection{Dataset Construction}
\label{appendix:dataset_construction}

In \Cref{subsec:training_data} we explained how we derive sets of clips with one action label each. Here we walk through the construction of VQA pairs from those labeled clips.

During training, we construct three questions from each clip: one binary question where the answer is ``yes'' (i.e., binary positive), one binary question where the answer is ``no" (i.e., binary negative), and one multiple-choice question (i.e., MCQ). For the binary negative question, we randomly select one action that is not the ground truth from the action list for that domain. For the MCQ, we randomly choose three negative options that are not the ground truth from the action list for the relevant domain. Although the \dataset~benchmark only consists of binary questions, initial experiments showed that including MCQs in the training mixture improves binary accuracy. We also experimented with 10-way MCQs (i.e., a MCQ with 9 negative distractors), but decided against it because it induced a much higher \textit{binary bias} (which we define as the absolute difference between binary positive accuracy and binary negative accuracy).

\subsection{Training Setup}
\label{appendix:training_setup}

In \Cref{subsec:post_training} we detailed the model architecture and our frame sampling approach. Here we include additional information on our training procedure. During training, we train the ViT, the connector, and the LLM using learning rates $5\times10^{-6}$, $5\times10^{-6}$ and $1\times10^{-5}$ respectively. We employ a cosine learning rate decay to $0.1$ of the initial learning rate. Following~\cite{molmov1}, the connector uses features from the third-to-last and ninth-from-last ViT layers. For each frame, $3\times3$ patch windows are pooled into a single vector using a multi-headed attention layer, where the mean of the patches serves as the query and the pooled features are projected using an MLP to the LLM's token space. For each training video sample, we pack multiple question-answer (QA) pairs. The LLM attention mask is customized such that text from one QA pair does not attend to the text from another pair. (As mentioned above in \S\ref{appendix:dataset_construction}, each video clip is accompanied by three QA pairs.) For additional inquires about the model, please refer to \cite{clark2025molmo2}.
\clearpage
\section{Data Filtering Strategies}
\label{appendix:data_filtering_strategies}

The data filtering strategies we employ are briefly described in \Cref{subsec:training_data}. Here we explain our intuition behind each strategy, the per-domain yields of each strategy, the category-level results of post-training a Molmo2-4B model on each strategy, and a brief analysis of the training results.

We began with the hypothesis that having as many independent signals align as possible would yield the highest-quality labels. There were two signals that were easily extracted at scale: the presence of an action in the video's title (``title match''), and the presence of an action in the video's transcript (``transcript match''). Adhering to our philosophy of having an \textit{extremely strict filter}, we chose to require the action to be said within one second of the clip for the ``transcript match'' to count. This resulted in the \strict~filter. While our hypothesis of such a strict filter yielding high-quality data was largely confirmed by initial experiments on domains like skateboarding, this filter's yield was too low on domains like whittling and fencing (see \Cref{tab:filtering_strategy_yields}). A natural solution to increasing the number of clips yielded by a filter is to relax the filter's strictness. Hence, we dropped the title match requirement, thereby keeping all clips with a transcript match; this is the \transcript~filter. In many cases, \transcript~yielded more clips than \strict, largely solving our problem of low yields. Once we had derived a filter (\transcript) by relaxing the title match requirement of \strict, it seemed fitting to derive a filter by relaxing the transcript match requirement. After some experimentation, we landed on \titleoneclip. The intuition here is that if there is a title match, then the video is likely to contain at least one clip of that action; if our localizer only finds one clip of that domain, then that clip must be of the title action. To make an analogy to the classic pigeonhole problem, if there is one pigeon (i.e., action from the title) and only one hole (i.e., clip found by localizer), then the pigeon must be assigned to that hole (i.e., the title action must be assigned to the one and only clip). Thus we arrived at our filtering strategies.

\begin{table*}[!h]
    \begin{tcolorbox}[
    enhanced,
    width=1.0\linewidth,
    colback=polaris-bg-elevated,
    colframe=polaris-border-subtle,
    arc=3mm,
    boxrule=0.4pt,
    left=8pt,
    right=8pt,
    top=8pt,
    bottom=8pt,
]
    \caption{\textbf{Per-category performance of different filtering strategies.} The first table reports multiple-choice accuracy; the second table reports binary 0-shot accuracy. In each table, the highest value in each column is in \textbf{bold}. The winning filtering strategy, \titleoneclip, has the same weaknesses -- Beauty and Food -- across evaluation settings, suggesting that its data for these categories is poor.}
    \centering
    \resizebox{\linewidth}{!}{
    \begin{tabular}{@{}lccccccccc@{}}
        \toprule[1.5pt] 
         \textbf{Filter} & \textbf{Size} & \textbf{Beauty} & \textbf{Crafts} & \textbf{Dance} & \textbf{Food} & \textbf{Hobbies} & \textbf{Medical} & \textbf{Sports} & \textbf{Overall} \\ \midrule
         \model~(base) & - & 44.8 & 35.7 & 30.3 & \textbf{74.7} & 40.2 & 59.0 & 38.7 & 42.0 \\
         [0.8ex] \hdashline \\[-1.2ex]
         \transcript & 496K & \textbf{69.8} & 52.7 & 46.0 & 63.2 & 49.6 & 54.8 & 38.1 & 48.2 \\
         \strict & 207K & 57.8 & 50.0 & 48.2 & 65.1 & 46.6 & 54.8 & 42.4 & 48.5 \\
         \titleoneclip & 162K & 66.4 & \textbf{59.8} & \textbf{52.0} & 69.4 & \textbf{52.1} & \textbf{64.9} & \textbf{44.4} & \textbf{53.5} \\
        \bottomrule[1.1pt]
    \end{tabular}
    }
    
    \vspace{10pt}
    
    \resizebox{\linewidth}{!}{
    \begin{tabular}{@{}lccccccccc@{}}
        \toprule[1.1pt] 
         \textbf{Filter} & \textbf{Size} & \textbf{Beauty} & \textbf{Crafts} & \textbf{Dance} & \textbf{Food} & \textbf{Hobbies} & \textbf{Medical} & \textbf{Sports} & \textbf{Overall} \\ \midrule
         \model~(base) & - & 56.0 & 54.0 & 53.1 & 73.1 & 52.3 & 61.2 & 53.0 & 55.3  \\
         [0.8ex] \hdashline \\[-1.2ex]
         \transcript & 496K & \textbf{75.9} & 63.4 & 62.4 & \textbf{76.3} & 66.7 & 67.6 & 60.4 & 65.0 \\
         \strict & 207K & 68.1 & 60.9 & 61.7 & 75.0 & 64.8 & 63.3 & 61.1 & 63.7 \\
         \titleoneclip & 162K & 73.3 & \textbf{69.0} & \textbf{66.1} & 73.7 & \textbf{67.8} & \textbf{69.2} & \textbf{61.5} & \textbf{66.6} \\
        \bottomrule[1.5pt]
    \end{tabular}
    }
    
    \label{tab:filtering_strategy_category_performance}
\end{tcolorbox}
\end{table*}

We train three models, one each for the datasets yielded by each filtering strategy. The overall accuracies of these models are reported in \Cref{tab:filter_table}. Category-level results are in \Cref{tab:filtering_strategy_category_performance}. Domain-level results are in \Cref{tab:filtering_strategy_domain_performance_binary} and \Cref{tab:filtering_strategy_domain_performance_mcq}. Even though \titleoneclip~attains the best overall performance (among filtering strategies) on \dataset~in both the binary and multiple-choice settings, it only achieves the highest accuracy on 22 of 37 and 17 of 37 domains respectively,\footnote{Including 3 and 2 ties, respectively.} affirming the domain-to-domain variation in filtering strategy effectiveness. 

Perusing these tables, the question naturally arises: why do certain filtering strategies fare better than others in terms of downstream performance on \dataset? Unlike other tasks \cite{openthoughts} where dataset size has a profound impact on downstream performance, the filter with the best \dataset~performance is actually the smallest in size. Hence, scale itself cannot explain the differences in downstream performance. Rather, we hypothesize that downstream performance is primarily impacted by \textit{clip quality} and \textit{intra-domain uniformity}. Concretely, clip quality refers to the accuracy with which action labels are assigned to clips by a filtering strategy, and intra-domain uniformity refers to the extent to which the counts of clips labeled by each action (within a domain) follows the uniform distribution. The intuition for the former is trivial; for the latter, since the test set presents a uniform \# of questions for each action in a domain, we believe that a training dataset which contains equal numbers of clips for each action within a domain is poised to perform best.\footnote{NB: certain filtering strategies yield skewed distributions for certain domains. For instance, the \transcript~gym data contains nearly 30k clips of \texttt{squats}; we believe that seeing such a disproportionate number of squat clips during training makes the model worse at discerning other gym actions such as pushups or deadlifts.}  We leave rigorous testing of this hypothesis to future work.

\begin{table*}[!h]
    \begin{tcolorbox}[
    enhanced,
    width=1.0\linewidth,
    colback=polaris-bg-elevated,
    colframe=polaris-border-subtle,
    arc=3mm,
    boxrule=0.4pt,
    left=8pt,
    right=8pt,
    top=8pt,
    bottom=8pt,
    ]
    \caption{\textbf{Filtering strategy yields.} The last three columns list clip yields for the filtering strategies in decreasing order of total yield: \transcript, \strict, and \titleoneclip. For a given row, compare the relative ranking of values in the last three columns of this table to the relative ranking of values in the last three columns of Tables \ref{tab:filtering_strategy_domain_performance_mcq} and \ref{tab:filtering_strategy_domain_performance_binary}; such a comparison proves that the yield of a filtering strategy is a poor indicator of downstream performance on the \dataset~benchmark. Note that the \# of actions in the training data exceeds the \# of actions in the benchmark; the latter action list is further refined (hence used for evaluation), but we are releasing training data with the former since it is more expansive. }
    \label{tab:filtering_strategy_yields}
    \centering
    \resizebox{\linewidth}{!}{
    \begin{tabular}{@{}cccccc@{}}
        \toprule[1.5pt]
        \multirow{2}{*}{\textbf{Category Name}} & \multirow{2}{*}{\centering \textbf{Domain Name}} & \multirow{2}{*}{\centering \textbf{\# Actions}} & \multirow{2}{2cm}{\centering \textbf{Transcript Localized}} & \multirow{2}{3.5cm}{\centering \textbf{Transcript Localized Title Match}} & \multirow{2}{2cm}{\centering \textbf{Single Action}} \\
        \\
        \midrule 
        \multirow{3}{*}{\centering Beauty \& Self Care} 
        & Hairstyling & 13 & \textbf{5,775} & 2,029 & 1,407 \\
        & Spa Massage & 13 & \textbf{3,258} & 1,350 & 763 \\
        & Tattooing & 7 & \textbf{782} & 145 & 222 \\
        \multirow{8}{*}{\centering Crafts \& Art} \\
        [-1.8ex] \hdashline \\[-1.2ex]
        & Calligraphy & 8 & \textbf{5,508} & 231 & 101 \\
        & Crochet & 41 & 5,081 & 4,570 & \textbf{10,988}\\
        & Hand Sewing / Embroidery & 36 & 687 & 459 & \textbf{6,544}\\
        & Knots & 55 & 4,092 & 3,882 & \textbf{20,357} \\
        & Painting & 8 & \textbf{2,949 }& 1,472 & 491 \\
        & Pottery & 11 & \textbf{5,889 }& 3,077 & 817 \\
        & Woodworking / Whittling & 4 & \textbf{1,140} & 29 & 11 \\
        \multirow{6}{*}{\centering Dance} \\
        [-1.8ex] \hdashline \\[-1.2ex]
        & Ballet & 39 & \textbf{17,358 }& 5,563 & 3,476 \\
        & Bharatanatyam & 31 & 856 & 311 & \textbf{3,198} \\ 
        & Break Dance & 33 & \textbf{2,573} & 1,507 & 395 \\ 
        & Salsa & 21 & \textbf{8,424} & 2,029 & 2,425 \\
        & Tap Dance & 31 & \textbf{13,089} & 3,563 & 1,398 \\
        \multirow{4}{*}{\centering Food \& Beverage} \\
        [-1.8ex] \hdashline \\[-1.2ex]
        & Bartending & 27 & \textbf{2,017} & 1,273 & 389\\
        & Coffee & 15 & \textbf{3,361 }& 2,432 & 611 \\
        & Cooking & 48 &\textbf{82,795} & 35,519 & 2,878 \\
        \multirow{9}{*}{\centering Hobbies} \\
        [-1.8ex] \hdashline \\ [-1.2ex]
        & Bouldering & 22 & 2,256 & 933 & \textbf{5,387} \\
        & Gardening & 26 & \textbf{4,063 }& 2,297 & 1,468\\
        & Gym & 26 &\textbf{ 79,793} & 68,869 & 21,338\\
        & Juggling & 28 & \textbf{1,582 }& 941 & 347\\
        & Parkour & 40 & 6,424 & 4,034 & \textbf{7,109} \\
        & Pen Spinning & 34 &\textbf{ 6,766 }& 2,777 & 2,316 \\
        & Skateboarding & 49 & \textbf{52,823} & 10,062 & 16,911 \\
        & Yo-yo & 53 & \textbf{6,742} & 3,526 & 2,259 \\
        \multirow{4}{*}{\centering Medical} \\
        [-1.8ex] \hdashline \\ [-1.2ex]
        & Neurological Abnormalities & 24 & \textbf{2,912} & 983 & 433 \\
        & Neurological Assessments & 15 & \textbf{820} & 381 & 313 \\
        & Suturing & 15 & 748 & 415 & \textbf{1,439} \\
        \multirow{10}{*}{\centering Sports} \\
        [-1.8ex] \hdashline \\ [-1.2ex]
        & American Football & 64 & \textbf{36,317} & 13,608 & 11,658 \\
        & Basketball & 46 & \textbf{82,847} & 13,205 & 11,217 \\
        & Cheerleading & 24 & 1,027 & 622 & \textbf{3,502} \\
        & Cricket & 46 & \textbf{8,423} & 3,602 & 7,033 \\
        & Figure Skating & 40 & \textbf{17,507}& 2,693 & 4,584 \\
        & Ice Hockey & 38 & \textbf{4,247} & 1,955 & 2,248 \\
        & Soccer & 42 & \textbf{12,108} & 4,247 & 4,382 \\
        & Tennis & 19 & \textbf{3,404} & 2,128 & 1,186 \\
        \multirow{2}{*}{} \\
        [-1.5ex] \hdashline \\ [-1.8ex]
        \textit{All} & \textit{All} & 1,100 & \textbf{496,443} & 206,719 & 161,601 \\
     \bottomrule[1.5pt]
    \end{tabular}
    }
\end{tcolorbox}
\end{table*}
\begin{table*}[!h]
    \begin{tcolorbox}[
    enhanced,
    width=1.0\linewidth,
    colback=polaris-bg-elevated,
    colframe=polaris-border-subtle,
    arc=3mm,
    boxrule=0.4pt,
    left=8pt,
    right=8pt,
    top=8pt,
    bottom=8pt,
]
    \caption{\textbf{Per-domain multiple-choice performance of different filtering strategies.} The last three columns contain accuracy percentages for the three filtering strategies in decreasing order of total yield: \transcript, \strict, and \titleoneclip. Please keep the number of questions for each domain, listed in the third column, in mind when considering the significance of a change in accuracy. The highest accuracy for each domain is in \textbf{bold}. Note that the base model achieves the best performance for coffee, cooking, gardening, neurological abnormalities, and football, suggesting that our training data for these domains may be poorly-labeled.}
    \label{tab:filtering_strategy_domain_performance_mcq}
    \centering
    \resizebox{\linewidth}{!}{
    \begin{tabular}{@{}ccccccc@{}}
        \toprule[1.5pt]
        \multirow{2}{*}{\textbf{Category Name}} & \multirow{2}{*}{\centering \textbf{Domain Name}} & \multirow{2}{*}{\centering \textbf{\# Questions}} & \multirow{2}{*}{\centering \textbf{Base Model}} & \multirow{2}{2cm}{\centering \textbf{Transcript Localized}} & \multirow{2}{3.5cm}{\centering \textbf{Transcript Localized Title Match}} & \multirow{2}{2cm}{\centering \textbf{Single Action}} \\
        \\
        \midrule 
        \multirow{3}{*}{\centering Beauty \& Self Care} 
        & Hairstyling & 48 & 52.08 & \textbf{68.75} & 64.58 & \textbf{68.75} \\
        & Spa Massage & 44 & 40.91 & \textbf{79.55} & 63.64 & 77.27 \\
        & Tattooing   & 24 & 37.50 & \textbf{54.17} & 33.33 & 41.67 \\
        \multirow{8}{*}{\centering Crafts \& Art} \\
        [-1.8ex] \hdashline \\[-1.2ex]
        & Calligraphy               & 32  & 40.63 & \textbf{59.38} & 43.75 & 40.63 \\
        & Crochet                   & 60  & 38.33 & 48.33 & 51.67 & \textbf{56.67} \\
        & Hand Sewing / Embroidery  & 152 & 33.55 & 43.42 & 36.18 & \textbf{60.53} \\
        & Knots                     & 220 & 32.27 & 57.73 & 60.00 & \textbf{68.18} \\
        & Painting                  & 32  & 40.63 & \textbf{65.63} & 53.13 & 50.00 \\
        & Pottery                   & 40  & 55.00 & \textbf{57.50} & \textbf{57.50} & 52.50 \\
        & Woodworking / Whittling   & 16  & 25.00 & \textbf{37.50} & 25.00 & 25.00 \\
        \multirow{6}{*}{\centering Dance} \\
        [-1.8ex] \hdashline \\[-1.2ex]
        & Ballet        & 156 & 36.54 & 47.44 & \textbf{53.21} & 52.56 \\
        & Bharatanatyam & 72  & 15.28 & 20.83 & 38.89 & \textbf{75.00} \\
        & Break Dance   & 132 & 29.55 & \textbf{60.61} & 51.52 & 50.76 \\
        & Salsa         & 76  & 36.84 & 48.68 & \textbf{56.58} & \textbf{56.58} \\
        & Tap Dance     & 112 & 27.68 & \textbf{41.07} & 37.50 & 34.82 \\
        \multirow{4}{*}{\centering Food \& Beverage} \\
        [-1.8ex] \hdashline \\[-1.2ex]
        & Bartending & 112 & 71.43 & \textbf{80.36} & 72.32 & 72.32 \\
        & Coffee     & 64  & \textbf{78.13} & 65.63 & 62.50 & 67.19 \\
        & Cooking    & 196 & \textbf{75.51} & 52.55 & 61.73 & 68.37  \\
        \multirow{9}{*}{\centering Hobbies} \\
        [-1.8ex] \hdashline \\ [-1.2ex]
        & Bouldering    & 88  & 30.68 & 42.05 & 40.91 & \textbf{44.32} \\
        & Gardening     & 80  & \textbf{91.25} & 81.25 & 71.25 & 78.75 \\
        & Gym           & 88  & 68.18 & 62.50 & 64.77 & \textbf{69.32} \\
        & Juggling      & 104 & 28.85 & \textbf{49.04} & 32.69 & 45.19 \\
        & Parkour       & 160 & 46.88 & 55.00 & 49.38 & \textbf{56.88} \\
        & Pen Spinning  & 128 & 27.34 & 53.13 & 47.66 & \textbf{58.59} \\
        & Skateboarding & 188 & 32.98 & 26.60 & 29.79 & \textbf{33.51} \\
        & Yo-yo         & 200 & 27.00 & 50.00 & \textbf{51.50} & 50.50 \\
        \multirow{4}{*}{\centering Medical} \\
        [-1.8ex] \hdashline \\ [-1.2ex]
        & Neurological Abnormalities & 80 & \textbf{60.00} & 51.25 & 50.00 & 58.75 \\
        & Neurological Assessments   & 60 & 71.67 & 70.00 & 63.33 & \textbf{83.33} \\
        & Suturing                   & 48 & 41.67 & 41.67 & \textbf{52.08} & \textbf{52.08} \\
        \multirow{10}{*}{\centering Sports} \\
        [-1.8ex] \hdashline \\ [-1.2ex]
        & American Football & 216 & \textbf{38.89} & 31.02 & 36.11 & 37.04 \\
        & Basketball        & 180 & 42.78 & 35.00 & 42.22 & \textbf{44.44} \\
        & Cheerleading      & 80 & 60.00 & 62.50 & 65.00 & \textbf{82.50} \\
        & Cricket           & 164 & 28.66 & 35.37 & \textbf{35.98} & 34.76 \\
        & Figure Skating    & 160 & 38.75 & 36.88 & 39.38 & \textbf{43.13} \\
        & Ice Hockey        & 152 & 39.47 & 41.45 & 44.74 & \textbf{53.29} \\
        & Soccer            & 160 & 39.38 & 35.62 & \textbf{49.38} & 41.87 \\
        & Tennis            & 76  & 25.00 & \textbf{46.05} & 38.16 & 36.84 \\
        \multirow{2}{*}{} \\
        [-1.5ex] \hdashline \\ [-1.8ex]
        \textit{All} & \textit{All} & 4,000 & 42.00 & 48.20 & 48.47 & \textbf{53.50} \\
     \bottomrule[1.5pt]
    \end{tabular}
    }
\end{tcolorbox}
\end{table*}
\begin{table*}[!h]
    \begin{tcolorbox}[
    enhanced,
    width=1.0\linewidth,
    colback=polaris-bg-elevated,
    colframe=polaris-border-subtle,
    arc=3mm,
    boxrule=0.4pt,
    left=8pt,
    right=8pt,
    top=8pt,
    bottom=8pt,
]
    \caption{\textbf{Per-domain binary 0-shot performance of different filtering strategies.} The last three columns contain accuracy percentages for the three filtering strategies in decreasing order of total yield: \transcript, \strict, and \titleoneclip. Please keep the number of questions for each domain, listed in the third column, in mind when considering the significance of a change in accuracy. The highest accuracy for each domain is in \textbf{bold}. Note that there are a whole 12 domains where the base model performs below random chance.}
    \label{tab:filtering_strategy_domain_performance_binary}
    \centering
    \resizebox{\linewidth}{!}{
    \begin{tabular}{@{}ccccccc@{}}
        \toprule[1.5pt]
        \multirow{2}{*}{\textbf{Category Name}} & \multirow{2}{*}{\centering \textbf{Domain Name}} & \multirow{2}{*}{\centering \textbf{\# Questions}} & \multirow{2}{*}{\centering \textbf{Base Model}} & \multirow{2}{2cm}{\centering \textbf{Transcript Localized}} & \multirow{2}{3.5cm}{\centering \textbf{Transcript Localized Title Match}} & \multirow{2}{2cm}{\centering \textbf{Single Action}} \\
        \\
        \midrule 
        \multirow{3}{*}{\centering Beauty \& Self Care} 
        & Hairstyling & 48 & 66.67 & \textbf{75.00} & 72.92 & 66.67 \\
        & Spa Massage & 44 & 47.73 & 81.82 & 65.91 & \textbf{86.36} \\
        & Tattooing   & 24 & 50.00 & \textbf{66.67 }& 62.50 & 62.50 \\
        \multirow{8}{*}{\centering Crafts \& Art} \\
        [-1.8ex] \hdashline \\[-1.2ex]
        & Calligraphy               & 32  & 56.25 & \textbf{71.88} & 56.25 & 56.25 \\
        & Crochet                   & 60  & 50.00 & 56.67 & 61.67 & \textbf{65.00} \\
        & Hand Sewing / Embroidery  & 152 & 50.00 & 57.24 & 51.97 & \textbf{64.47} \\
        & Knots                     & 220 & 54.09 & 67.27 & 65.45 & \textbf{75.00}\\
        & Painting                  & 32  & 62.50 & 65.63 & 65.63 & \textbf{68.75} \\
        & Pottery                   & 40  & 70.00 & \textbf{75.00} & 70.00 & \textbf{75.00} \\
        & Woodworking / Whittling   & 16  & 43.75 & 43.75 & \textbf{56.25} & \textbf{56.25} \\
        \multirow{6}{*}{\centering Dance} \\
        [-1.8ex] \hdashline \\[-1.2ex]
        & Ballet        & 156 & 59.62 & 66.67 & 63.46 & \textbf{74.36} \\
        & Bharatanatyam & 72  & 48.61 & 55.56 & 52.78 & \textbf{65.28} \\
        & Break Dance   & 132 & 53.79 & 64.39 & 62.88 & \textbf{65.15} \\
        & Salsa         & 76  & 48.68 & 65.79 & \textbf{69.74} & \textbf{69.74} \\
        & Tap Dance     & 112 & 49.11 & 56.25 & \textbf{58.04} & 53.57 \\
        \multirow{4}{*}{\centering Food \& Beverage} \\
        [-1.8ex] \hdashline \\[-1.2ex]
        & Bartending & 112 & 70.54 & \textbf{85.71} & 75.89 & 75.89 \\
        & Coffee     & 64  & \textbf{82.81} & 71.88 & 73.44 & 71.88 \\
        & Cooking    & 196 & 71.43 & 72.45 & \textbf{75.00} & 72.96 \\
        \multirow{9}{*}{\centering Hobbies} \\
        [-1.8ex] \hdashline \\ [-1.2ex]
        & Bouldering    & 88  & 47.73 & 57.95 & 48.86 & \textbf{67.05} \\
        & Gardening     & 80  & 65.00 & \textbf{87.50} & 80.00 & 82.50 \\
        & Gym           & 88  & 68.18 & 75.00 & 71.59 & \textbf{79.55} \\
        & Juggling      & 104 & 45.19 & \textbf{61.54} & 58.65 & 59.62 \\
        & Parkour       & 160 & 56.88 & 72.50 & 71.25 & \textbf{74.37} \\
        & Pen Spinning  & 128 & 45.31 & 72.66 & 66.41 & \textbf{73.44} \\
        & Skateboarding & 188 & 47.34 & 54.26 & \textbf{59.57} & 57.45 \\
        & Yo-yo         & 200 & 51.50 & \textbf{64.50} & \textbf{64.50} & 62.00 \\
        \multirow{4}{*}{\centering Medical} \\
        [-1.8ex] \hdashline \\ [-1.2ex]
        & Neurological Abnormalities & 80 & 65.00 & \textbf{67.50} & 62.50 & \textbf{67.50} \\
        & Neurological Assessments   & 60 & 63.33 & 73.33 & 63.33 & \textbf{75.00} \\
        & Suturing                   & 48 & 52.08 & 60.42 & \textbf{64.58} & \textbf{64.58} \\
        \multirow{10}{*}{\centering Sports} \\
        [-1.8ex] \hdashline \\ [-1.2ex]
        & American Football & 216 & 48.61 & 53.24 & \textbf{56.94} & 54.17 \\
        & Basketball        & 180 & 48.89 & 55.56 & \textbf{56.11} & 55.00 \\
        & Cheerleading      & 80 & 66.25 & 77.50 & 75.00 & \textbf{86.25} \\
        & Cricket           & 164 & 49.39 & 59.76 & 58.54 & \textbf{60.37} \\
        & Figure Skating    & 160 & 51.25 & \textbf{60.00} & 58.13 & \textbf{60.00} \\
        & Ice Hockey        & 152 & 59.21 & 68.42 & 65.13 & \textbf{71.05} \\
        & Soccer            & 160 & 55.63 & 60.62 & \textbf{65.00} & 64.38 \\
        & Tennis            & 76 & 55.26 & 60.53 & \textbf{65.79} & 51.32 \\
        \multirow{2}{*}{} \\
        [-1.5ex] \hdashline \\ [-1.8ex]
        \textit{All} & \textit{All} & 4,000 & 55.33 & 65.00 & 63.70 & \textbf{66.60} \\
     \bottomrule[1.5pt]
    \end{tabular}
    }
\end{tcolorbox}
\end{table*}
}

\end{document}